\documentclass[AMS,STIX1COL]{WileyNJD-v2}

\usepackage{graphicx}
\usepackage{hyperref}
\hypersetup{
    colorlinks=true,
    linkcolor=blue,
    filecolor=magenta,      
    urlcolor=cyan,
}
\usepackage{subcaption}

\articletype{Article Type}%

\received{xx}
\revised{xx}
\accepted{xx}

\newcommand{\mbold}[1]{\mbox{\boldmath ${#1}$}}
\graphicspath{{./figures/}}
\begin{document}

\title{(Decision and regression) tree ensemble based kernels for regression and classification
 \protect\thanks{This is an example for title footnote.}}

\author[1]{Dai Feng*}

\author[2]{Richard Baumgartner}

\authormark{Dai Feng \& Richard Baumgartner}

\address[1]{\orgdiv{Data and Statistical Sciences}, \orgname{AbbVie Inc.}, \orgaddress{\state{North Chicago, Illinois}, \country{United States of America}}}

\address[2]{\orgdiv{Merck \& Co., Inc.}, \orgname{Kenilworth, NJ}, \country{United States of America}}

\corres{*Dai Feng, Data and Statistical Sciences, AbbVie Inc. \email{dai.feng@abbvie.com}}

\presentaddress{This is sample for present address text this is sample for present address text}

\abstract[Summary] {Tree based ensembles such as Breiman’s random forest (RF) and Gradient Boosted Trees (GBT) can be interpreted as implicit kernel generators, where the ensuing proximity matrix represents the data-driven tree ensemble kernel. Kernel perspective on the RF has been used to develop a principled framework for theoretical investigation of its statistical properties. Recently, it has been shown that the kernel interpretation is germane to other tree-based ensembles e.g. GBTs. However, practical utility of the links between kernels and the tree ensembles has not been widely explored and systematically evaluated. 

Focus of our work is investigation of the interplay between kernel methods and the tree based ensembles including the RF and GBT. We elucidate the performance and properties of the RF and GBT based kernels in a comprehensive simulation study comprising of continuous and binary targets. We show that for continuous targets, the RF/GBT kernels are competitive to their respective ensembles in higher dimensional scenarios, particularly in cases with larger number of noisy features. For the binary target, the RF/GBT kernels and their respective ensembles exhibit comparable performance. We provide the results from real life data sets for regression and classification to show how these insights may be leveraged in practice. Overall, our results support the tree ensemble based kernels as a valuable addition to the practitioner's toolbox.

Finally, we discuss extensions of the tree ensemble based kernels for survival targets, interpretable prototype and landmarking classification and regression. We outline future line of research for kernels furnished by Bayesian counterparts of the frequentist tree ensembles.
}


\jnlcitation{\cname{%
\author{Dai Feng}, 
\author{Richard Baumgartner}(\cyear{2020})}, 
\ctitle{Random Forest (RF) Kernel for Regression, Classification and Survival} 
}

\maketitle


\section{Introduction}\label{sec1}
Random forest (RF) has been a successful and time-proven statistical machine learning method \cite{biau2016}. At first, RF was developed for classification and regression \cite{breiman2000}. Later it has been extended and adopted for additional types of targets such as time-to-event or ordered outcomes \cite{ishwaran2019}. RF belongs to the tree ensemble methods, where “base” tree learners are grown on bootstrapped samples of the training data set and then their predictions are aggregated to yield a final prediction. RF was conceived originally under the frequentist framework. However, Bayesian counterparts e.g. Mondrian random forest were also proposed \cite{balog2016}. 

In Ref \cite{breiman2000}, Breiman pointed out an alternative interpretation of the RF as a kernel generator.  The $n \times n$ proximity matrix (where $n$ is the number of samples) naturally ensuing from the construction of the RF plays here a key role. Each entry of the RF proximity matrix is an estimate of the probability that two points end up in the same terminal node \cite{breiman2000}. It is a symmetric positive-semidefinite matrix and it can be interpreted as a kernel akin to those previously proposed for the kernel methods \cite{breiman2000},\cite{scornet2016}.  To note, in Bayesian framework, Mondrian kernel denotes also the empirical frequency with which two points end up in the 
 same partition cell of a Mondrian sample \cite{balog2016}. Moreover the examples of frequently used (analytical) kernels include linear kernel, radial basis function (RBF), polynomial kernels, etc. \cite{herbich2001},\cite{schoelkopf2001}.  Asymptotically the RF kernel converges to the Laplace kernel \cite{breiman2000}. Similar convergence results were obtained also for Mondrian forests and Bayesian additive regression trees (BART) \cite{balog2016, linero2017}. In addition, other tree based ensembles such as gradient boosted trees (GBT) can be considered as kernel generators as recently shown in \cite{Chen2018}. 

Supervised kernel methods as usually applied, fit linear models in non-linear feature spaces that are induced by the kernels.  Popular choices in this class of algorithms include support vector machines (SVMs) and kernel ridge regression \cite{herbich2001},\cite{schoelkopf2001} or their refinements e.g. generalized "kernel" elastic net \cite{sokolov2016}. Bayesian approach to the kernel methods is represented by the Gaussian processes \cite{davies2014},\cite{rasmussen2006}. 

Relevant to our work is also similarity/dissimilarity based learning \cite{chen2009},\cite{pekalska2001},\cite{balcan2008} that was proposed for classification and regression. In similarity/dissimilarity learning the kernel entries are explicitly interpreted as pairwise similarities/dissimilarities between points (samples). RF proximity matrix or RF kernel fits readily into this paradigm.

The kernel interpretation of the RF was further explored and expounded theoretically to investigate its statistical 
 properties such as asymptotic convergence of RF and RF kernel estimates \cite{scornet2016}. Recent work on the GBT trees \cite{Chen2018} also provided insights into their kernel interpretation. On the other hand, there has been interest in the use of algorithms based on the RF kernel \cite{davies2014} in practice.  In the Ref \cite{davies2014} performance of RF kernels was found competitive for regression tasks on various data sets from the UCI repository. 

We are primarily building on the  \cite{davies2014} and \cite{Chen2018}. Our focus is investigation of the RF and GBT kernel based algorithms in regression and classification  and elucidation of their performance characteristics. The 
remainder of the manuscript is organized as follows: Section 2 introduces the theoretical framework of the tree ensemble based kernels (RF and GBT) for targets of interest, Section 3 provides a motivational example using the well known Fisher Iris data, Section 4 details a simulation study that systematically evaluates performance of the RF and GBT 
kernels in various scenarios, Section 5 summarizes the results on real life data sets and Section 6 provides discussion, conclusions and future research directions.

\section{Methodology}\label{sec2}

\subsection{Terminology}\label{sec2sub2}
Following Breiman \cite{breiman2000} and Refs. \cite{ishwaran2019},\cite{Chen2018} and \cite{scornet2016} we consider a supervised learning problem, where training set $D_n=\{(\mbold{X_1},Y_1), (\mbold{X_2},Y_2),\ldots,({\mbold{X_1},Y_n})\}$ is provided. $\mbold{X_i} \in R^p$ and $Y_i$ can be continuous or binary target. For continuous and binary targets the $Y_i \in R$ and $Y_i\in \{0,1\}$, respectively. 

\subsection{Kernels for Regression and Classification}\label{sec2sub3}
Kernel methods in the machine learning literature are a class of methods that are formulated in terms a similarity (Gram) matrix $\mbold{K}$. The similarity matrix $K_{i,j}=k(\mbold{X_i},\mbold{X_j})$ represents the similarity between two points $\mbold{X_i}$ and $\mbold{X_j}$.
Kernel methods have been well developed and there is a large body of references covering their different aspects \cite{herbich2001},\cite{schoelkopf2001},\cite{friedmanHastieTibshirani2009}. 
In our work we used a common kernel algorithm, namely kernel Ridge Regression (KRR) for regression and classification. For the two class classification, we developed the KRR model with targets of -1 and 1 denoting the two classes. The predicted class label was obtained by thresholding around 0.

KRR is a kernelized version of the traditional linear ridge regression with the L2-norm penalty. Given the kernel matrix $\mbold{K}$ estimated from the training set, first the coefficients $\mbold{\alpha}$ of the (linear) KRR predictor in the non-linear feature space induced by the kernel $k(.,.)$ are 
obtained:
\begin{eqnarray}
\mbold{\alpha}&=&(\mbold{K}+\lambda \mbold{I_n})^{-1}\mbold{Y}
\end{eqnarray}
where $\lambda$ is the regularization parameter.

The KRR predictor $h_{\text{KRR}}(\mbold{X})$ is given as:  

\begin{eqnarray}
h_{KRR}(\mbold{X})&=&\sum_{i=1}^{n} \alpha_i
\mbold{k(X_i,X)}=\mbold{Y^T}(\mbold{K}+\lambda\mbold{I_n})^{-1}\mbold{K_i}
\label{Eq:KRRPredictor}
\end{eqnarray}
where $\mbold{K_i}=(k(X_1,X), \ldots, k(X_n,X))$.

\subsection{Random Forest (RF) and the RF Kernel}\label{sec2sub4}
 Random Forest (RF) is defined  as an ensemble of tree predictors grown on bootstrapped samples of a training set\cite{breiman2000}. When considering an ensemble of tree predictors $\{h(.,\Theta_m,D_n), m=1,2,\ldots,M\}$, with $\{h(.,\Theta_m,D_n)\}$ representing a single tree. The $\Theta_1, \Theta_2,\ldots\Theta_M$ are iid random variables that encode the randomization necessary for the tree construction \cite{scornet2016},\cite{ishwaran2019}.

The RF predictor is obtained as:
\begin{eqnarray}
h_{\text{RF}}(\mbold{X},\Theta_1,\ldots,\Theta_m,D_n)&=& \frac{1}{M}\sum_{m=1}^M h(\mbold{X},\Theta_m,D_n)
 \end{eqnarray}
RF kernel ensuing from the RF is defined as a probability that  $\mbold{X_i}$ and $\mbold{X_j}$ are in the same terminal node $R_k(\Theta_m$) \cite{breiman2000},\cite{scornet2016}.
\begin{eqnarray}
k_{RF}(\mbold{X_i},\mbold{X_j})=\frac{1}{M}
\sum_{m=1}^M \sum_{k=1}^T I(\mbold{X_i},\mbold{X_j} \in R_k(\Theta_m))
\label{EqKernelRF}
\end{eqnarray}
where $I(\cdot)$ denotes the indicator function.

\subsection{Gradient Boosted Trees (GBT) and the GBT Kernel}\label{sec2sub5}
The GBT are (similarly to RF) ensemble of tree predictors. In contrast to the RF, the GBT ensemble predictor is obtained as a sum of weighted individual tree predictors $h_m(\mbold{X},D_n)$ through iterative optimization of an objective (cost) function \cite{friedman2001},\cite{Chen2016} :

\begin{eqnarray}
h_{\text{GBT}}(\mbold{X},D_n)&=& \sum_{m=1}^M h_m(\mbold{X},D_n)
\end{eqnarray}

The objective function of GBT comprises of a loss function and for the extreme gradient boosting a regularization term is added to control the model complexity. In our work we used the extreme gradient boosting (XGB) 
implementation of the GBTs \cite{Chen2016}. The objective function  of the XGB 
algorithm $L_{XGB}$ is given in the Appendix. We use GBT and XGB interchangeably hereafter.

As for the RF kernel, the GBT kernel is defined as a probability that  $\mbold{X_i}$ and $\mbold{X_j}$ are in the same terminal node $R_k$($h_m$) \cite{Chen2018}.

\begin{eqnarray}
k_{XGB}(\mbold{X_i},\mbold{X_j})=\frac{1}{M}
\sum_{m=1}^M \sum_{k=1}^T I(\mbold{X_i},\mbold{X_j} \in R_k(h_m))
\label{EqKernelXGB}
\end{eqnarray}

In general, the tree-based ensemble kernels can be characterized by a feature map $\phi$ that maps the input feature space ($R^{p}$) to the space generated by feature space partitioning ensuing from the tree ensemble \cite{balog2016},\cite{ren2015}, \cite{gogic2021}. The feature space induced by $\phi$ through random partitioning is also referred to as a random feature space \cite{balog2016},\cite{fan2020}. The mapping $\phi$ is defined as $\phi: R^{p}\rightarrow \{0,1\}^{P}$, where $P=MT$ is the number of terminal nodes across the tree ensemble. Thus, $\phi({X_i})$ can be represented by a $P$-dimensional vector where each entry corresponds to a particular terminal node in the tree ensemble. For each sample $X_i$, an entry in $\phi({X_i})$ then encodes belonging of $X_i$ to a particular terminal node $R_k$ and $\phi({X_i})_{k,m}=I(\mbold{X_i} \in R_k(\Theta_m))$. A tree ensemble based kernel that represents the random input feature space partitions can then obtained via $\phi$ (represented as $P$ dimensional column vector)  as:
\begin{eqnarray}
k(\mbold{X_i},\mbold{X_j})=\frac{1}{M}\phi({X_i})^T \phi(X_j)
\label{EqKernelGen}
\end{eqnarray}

For example, when the RF and XGB kernels are considered as special cases of tree based ensemble kernels, the Eqs. \ref{EqKernelGen} and \ref{EqKernelRF} coincide with \ref{EqKernelXGB}, respectively.

\subsection{RF and GBT Kernel Predictors for Regression and Classification}\label{sec2sub5}
  
RF/XGB kernel predictors for regression was obtained by substituting for the RF and XGB kernel in Eq.\ref{Eq:KRRPredictor}, respectively, as:
\begin{eqnarray}
h_{\text{RF/XGB-KRR}}(\mbold{X})&=&\sum_{i=1}^{n} \alpha_i
\mbold{k_{RF/XGB}(X_i,X)}=\mbold{Y^T}(\mbold{K_{RF/XGB}}+\lambda\mbold{I_n})^{-1}\mbold{K_{RF/XGB_i}}
\label{Eq:KRRrfpredictor}
\end{eqnarray}
The RF and XGB kernel predictors for classification was also obtained as that for regression using Eq.\ref{Eq:KRRrfpredictor}, by building a regression model with target classes denoted as $\{-1,1\}$ and a class prediction threshold of $0$.

The code for the simulation and real life data analysis was developed in the R programming language \cite{Rcran}. For the continuous and binary targets the ranger \cite{ranger} implementation of RF and xgboost \cite{xgboost} of XGB was used, respectively.
The regularization parameter $\lambda$ was chosen at minimum value, such that the matrix, $\mbold{K}+\lambda \mbold{I_n}$ was invertible. 

In the simulations, all algorithms were applied using their default parameters. In order to further elucidate the impact of tree depth on the simulation results we carried out a sensitivity analysis, where we doubled the minimum size of the tree terminal node to generate more shallow trees for RF. The doubled minimum tree node size equaled to 10 and 2 for regression and classification, respectively. For the XGB, the maximum depth of the tree equaled to 2 in the sensitivity analysis.

\section{Motivating Example}\label{sec3Mot}
\begin{figure}[ht]
\centerline{
\includegraphics[width=0.8\textwidth,height=0.5\textheight]{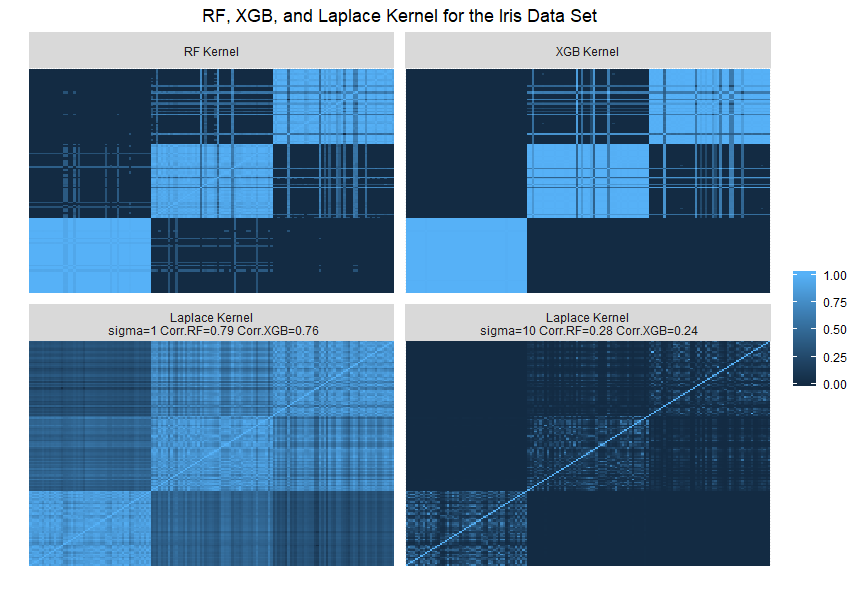}}
\caption{RF Kernel, XGB Kernel, and the Laplace Kernel for the Fisher Iris data set. Corr denotes the matrix correlation given by the Mantel statistic and sigma is the $\sigma$ parameter of the Laplace kernel.\label{fig:iris}}
\end{figure}

As a motivating example we show kernel matrices obtained from the the Fisher's Iris data. The Iris data consists of recordings of three Iris subspecies: Setosa, Versicolor and Virginica (50 samples each recorded
 on 4 numerical features). We compare the RF/GBT kernels and the Laplace kernel for this data set.
The Laplace kernel is defined as $k(\mbold{X_1}, \mbold{X_2})=\exp(\frac{-||X_1-X_2||_1}{\sigma}$).
In Figure \ref{fig:iris}, RF and GBT kernels 
are compared with the Laplace kernel for a couple of different values of the parameter $\sigma$.

The similarity between RF/GBT kernels obtained as a proximity matrix and the Laplace kernel is assessed by the Mantel statistic, i.e. matrix correlation in this case between two similarity matrices \cite{legendre2012}, respectively. In the Figure \ref{fig:iris}, the RF/GBT kernels capture the underlying structure of the data well and the three classes can be clearly distinguished. Similarly, the Laplace kernels also reflect (with different success) the partitioning of the data in three classes. The Laplace kernel that has the higher Mantel statistics with respect to the RF/GBT kernels appears to be the best in terms of the target alignment. 

\begin{figure}
\centerline{\includegraphics[width=0.8\textwidth,height=0.5\textheight]{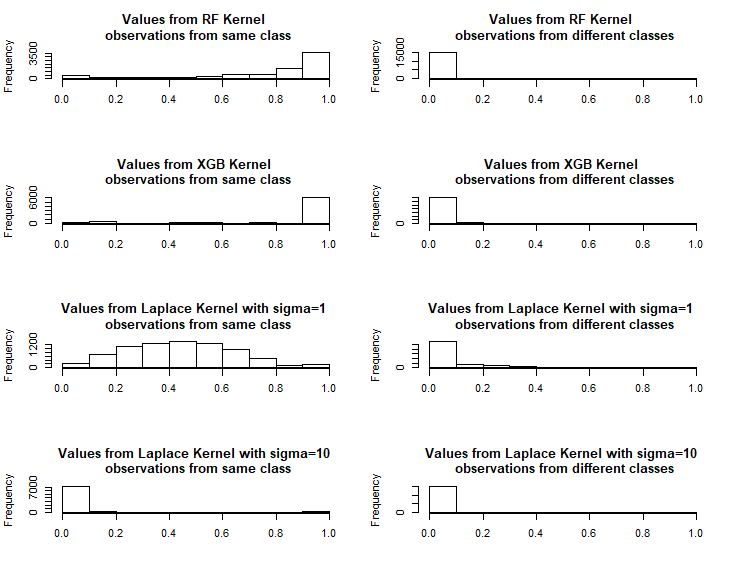}}
\caption{Distributions of values from RF Kernel, XGB Kernel, and the Laplace Kernel with $\sigma=1$ and $10$ for the Fisher Iris data set. 
} 
\label{fig:iris-kerVal}
\end{figure}

Furthermore, the RF/GBT kernels could characterize a more precise similarity function \cite{balcan2008} leading to more accurate classification results. When RF/GBT kernels are compared to the Laplace kernel, the observations from the same class are more similar (closer) to each other than those from different classes. This is demonstrated by the histograms shown in Figure \ref{fig:iris-kerVal}, with the RF/GBT kernel histograms peaking at 1 and 0 for the observations from the same class and those from different classes, respectively.

In this motivating example we illustrated potential utility of the tree ensemble 
based kernels on an example of a RF/GBT kernels. We also compared them with the Laplace kernel. In more complicated simulated and real life scenarios, the tree ensemble based kernels such as RF/GBT kernels usually outperform the Laplace kernel. The Laplace kernel is isotropic \cite{balog2016}. As the tree ensemble based kernels are anisotropic \cite{vens2011} they adapt to local data features well \cite{Chen2018}. Preliminary experiments with the Laplace kernel confirmed this notion. Therefore, in our manuscript we focus on the investigation of the utility of the RF/GBT kernels in building predictive models for regression and classification.

\section{Simulation}\label{sec3}
Simulation scenarios for performance evaluation of RF/GBT kernels for regression  were set up according to previously reported simulation benchmarks for continuous targets including
Friedman \cite{friedman}, Meier 1, Meier 2 \cite{meier},  van der Laan \cite{van2007super} and Checkerboard \cite{zhu2015}. These were also adapted for classification.

\subsection{Simulation Setup}\label{sec3sub1}

For each simulation scenario, the predictors were simulated from Uniform (Friedman, Meier 1, Meier 2, van der Laan) or Normal distributions (Checkerboard), respectively. 

Continuous targets were generated as $Y_i=f(\mbold{X_i})+\epsilon_i$. For the definitions of $f(\mbold{X_i})$ for each simulation case see below.

To generate a binary outcome $Y_i$ the continuous outcome was first centered by the median $M$ of its marginal distribution to obtain a balanced two class problem. Then the binary target was generated as a Bernoulli variable with $p_i=prob(Y_i=1|\mbold{X_i})$, where $p_i$ was calculated as follows:
\begin{eqnarray}
p_i &=& \frac{\exp{(f(\mbold{X_i})-M)}}{1+\exp{(f(\mbold{X_i})-M)}}\nonumber
 \end{eqnarray}
where $f(\mbold{X_i})$ is obtained from the continuous models. 

To characterize the intrinsic complexity of the classification problems, we also calculated the Bayes error rate from a large sample of the continuous outcomes ($n=10^{7}$) and subsequently applied the formula $\textnormal{Error}_{\textrm{Bayes}} =1-E\left(\underset{j}{\textrm{max}} Pr(Y=j|\mbold{X})\right), j\in\{0,1\}$ according to \cite{James2013}. The $j\in\{0,1\}$ refer to the class indicators. 

The five functional relationships $f(\mbold{X_i})$ between the predictors and target for different simulation settings are specified as follows.

1. Friedman. The setup for Friedman was as described in \citet{friedman}.
\begin{eqnarray}
X_{ij} &\sim& Uniform(0, 1), i=1,\ldots,n; j=1,\ldots,p \nonumber\\
\epsilon_i &\sim& N(0,1)\nonumber\\
f(\mbold{X_i}) &=& 10\sin{(\pi X_{i1}X_{i2})} + 20(X_{i3}-0.5)^2+10X_{i4} + 5X_{i5} + \epsilon_i\nonumber
\end{eqnarray}
Bayes error rate for the Friedman classification problem is $0.02$. It is the least complex problem by this measure among those investigated.

2. Checkerboard. In addition to Friedman, we simulated data from a Checkerboard-like model with strong correlation as in Scenario 3 of \citet{zhu2015}.

\begin{eqnarray}
\mbold{X_i} &\sim& N(0, \Sigma_{p\times p}), i=1, \ldots, n \nonumber\\
\epsilon_i &\sim& N(0,1)\nonumber\\
f(\mbold{X_i})&=&2X_{i5}X_{i10}+2X_{i15}X_{i20} + \epsilon_i\nonumber
 \end{eqnarray}
The $(j,k)$ component of $\Sigma$ is equal to $0.9^{|j-k|}$.
Bayes error rate for the Checkerboard classification problem is $0.18$.

3. van der Laan. The setup was studied in van der Laan et al. \cite{van2007super}.
\begin{eqnarray}
X_{ij} &\sim& Uniform(0, 1), i=1,\ldots,n; j=1,\ldots,p \nonumber\\
\epsilon_i &\sim& N(0,0.5)\nonumber\\
f(\mbold{X_i}) &=& \tilde{X}_{i1}\tilde{X}_{i2}+\tilde{X}_{i3}^2+\tilde{X}_{i8}\tilde{X}_{i10}-
\tilde{X}_{i6}^2+ \epsilon_i\nonumber\\
\tilde{X}_i &= &2(X_i-0.5)
\end{eqnarray}
Bayes error rate for the van der Laan classification problem is $0.34$, making it the most complex by this measure among those investigated.

4. Meier 1. This setup was investigated in  Meier et al. \cite{meier}.
\begin{eqnarray}
X_{ij} &\sim& Uniform(0, 1), i=1,\ldots,n; j=1,\ldots,p \nonumber\\
\epsilon_i &\sim& N(0,0.5)\nonumber\\
f(\mbold{X_i}) &=& -\sin(2\tilde{X}_{i1})+\tilde{X}_{i2}^2+\tilde{X}_{i3}-\exp(\tilde{X}_{i4})+ \epsilon_i\nonumber\\
\tilde{X}_i &= &2(X_i-0.5)
\end{eqnarray}
Bayes error rate for the Meier 1 classification problem is $0.28$.

5. Meier 2. This setup was investigated in  Meier et al. \cite{meier} as well.
\begin{eqnarray}
X_{ij} &\sim& Uniform(0, 1), i=1,\ldots,n; j=1,\ldots,p \nonumber\\
\epsilon_i &\sim& N(0,0.5)\nonumber\\
f(\mbold{X_i}) &=& -\tilde{X}_{i1}+(2\tilde{X}_{i2}-1)^2+\frac{\sin(2\pi\tilde{X}_{i3})}
{2-\sin(2\pi\tilde{X}_{i4})}+2\cos(2\pi\tilde{X}_{i4})+4\cos^2(2\pi\tilde{X}_{i4})+ \epsilon_i\nonumber\\
\tilde{X}_i &= &2(X_i-0.5)
\end{eqnarray}
Bayes error  rate for the Meier 2 classification problem is $0.19$.

We used mean squared error (MSE) and classification accuracy to measure the prediction performance for continuous and binary data, respectively. For continuous and binary data, we prefer smaller MSE and higher accuracy, respectively.

For each functional relationship $f(\mbold{X_i})$ (Friedman, Checkerboard, Meier 1, Meier 2, and van der Laan) and each outcome (continuous or binary), we simulated data from four scenarios with different samples sizes $n=800$ and $n=1600$ and number of covariates $p=20$ and $p=40$. Within each scenario, we simulated 200 data sets and for each data set we randomly chose 75\% of samples as training data and remaining 25\% as test data.

\subsection{Simulation Results}\label{sec3sub2}
 
\begin{figure}[ht]
\begin{subfigure}{.45\textwidth}
  \centering
  \includegraphics[height=0.20\textheight]{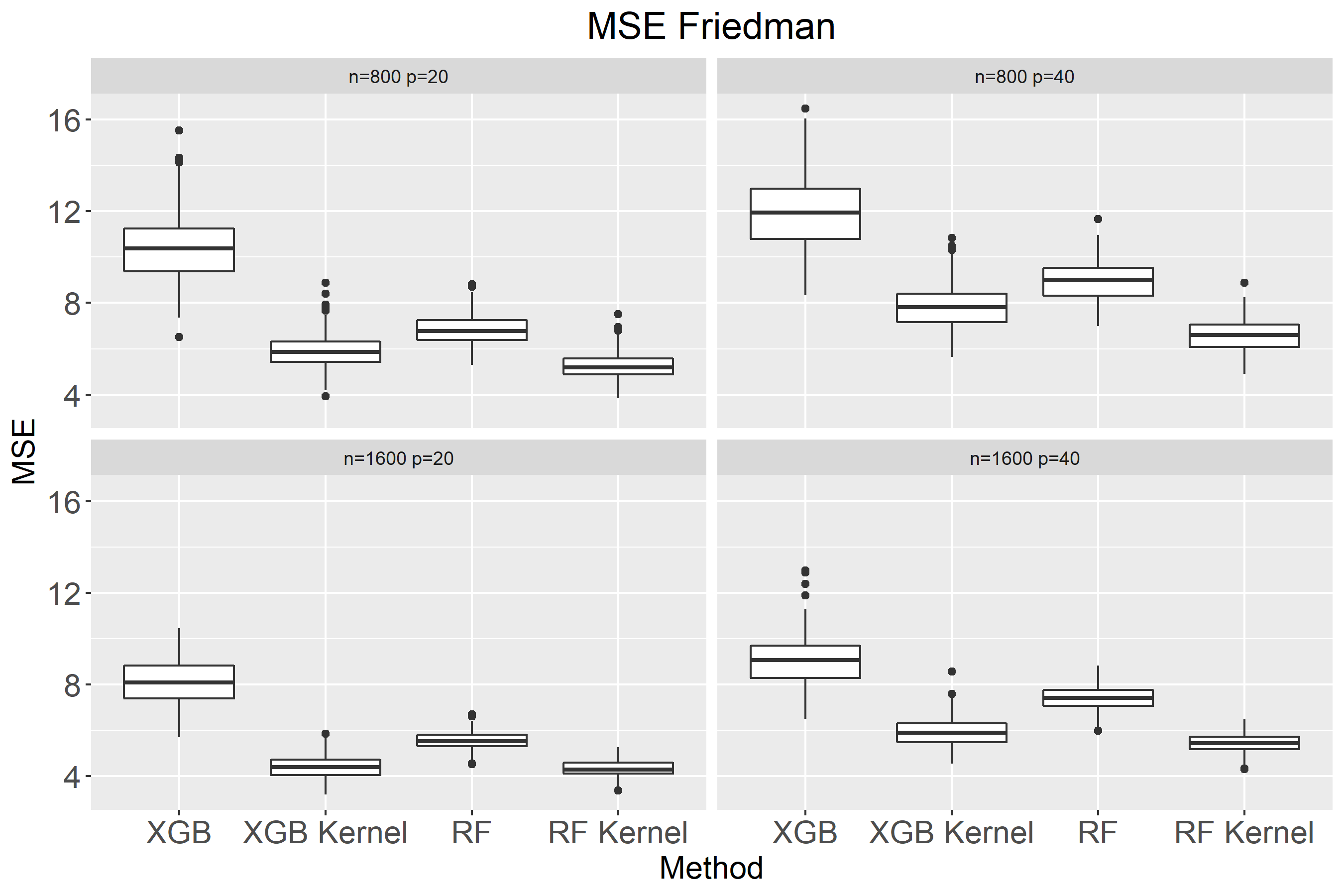}  
  \subcaption{}
  \label{fig:sub-first}
\end{subfigure}
\begin{subfigure}{.45\textwidth}
  \centering
  \includegraphics[height=0.2\textheight]{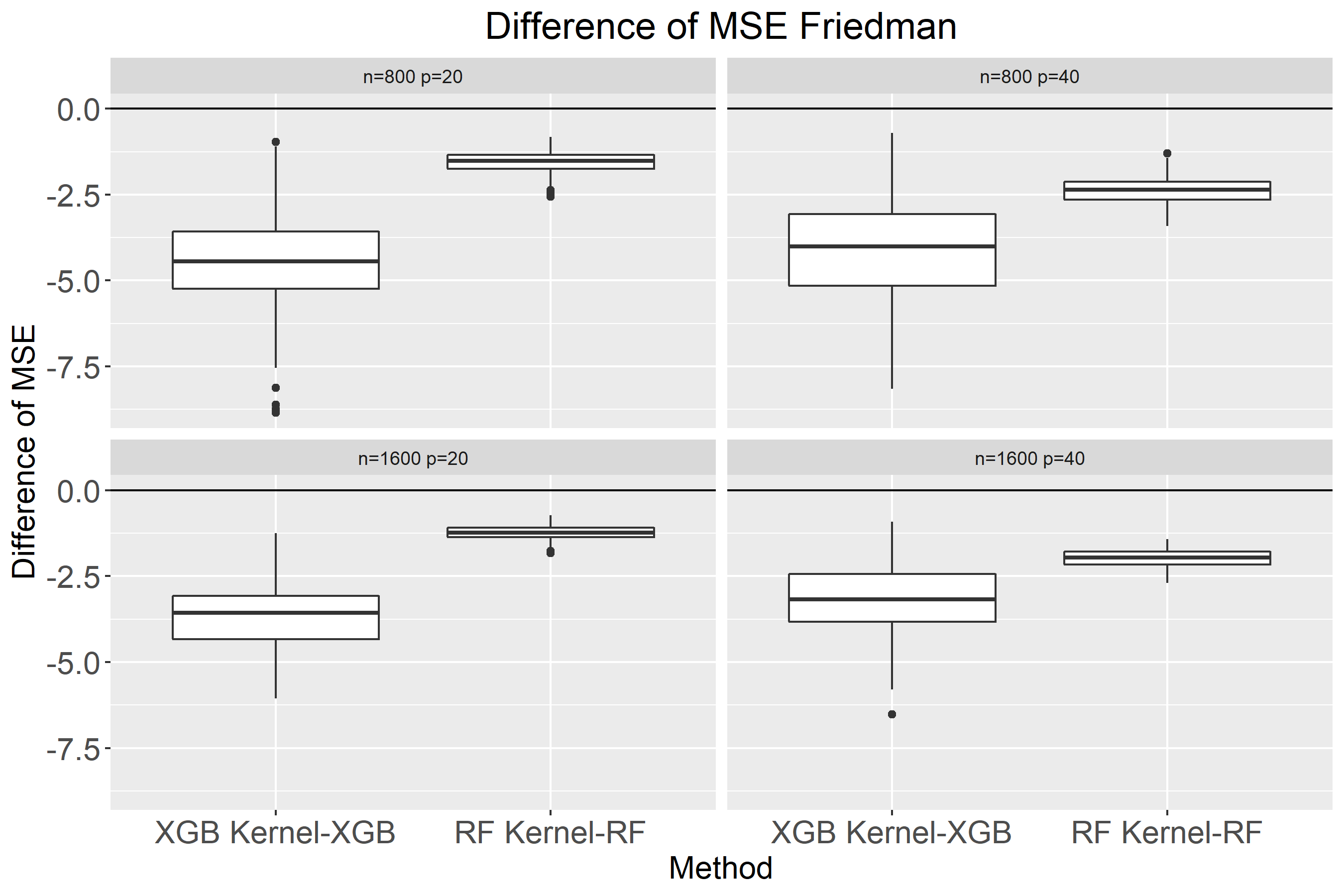}   
    \subcaption{}
  \label{fig:sub-first}
\end{subfigure}\\
\begin{subfigure}{.45\textwidth}
  \centering
  \includegraphics[height=0.2\textheight, width=0.9\textwidth]{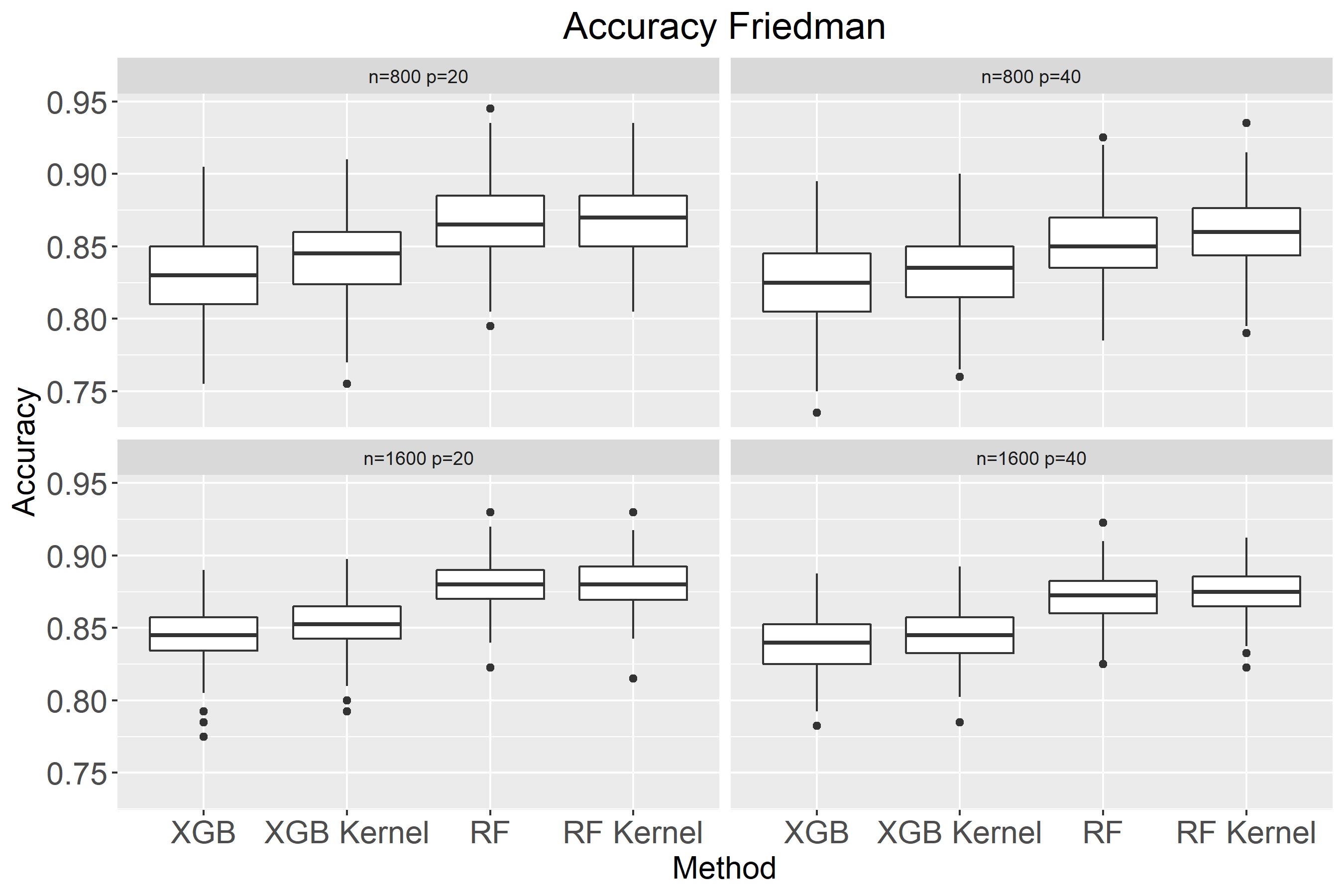} 
    \subcaption{}
  \label{fig:sub-first}
\end{subfigure}
\begin{subfigure}{.45\textwidth}
  \centering
  \includegraphics[height=0.2\textheight]{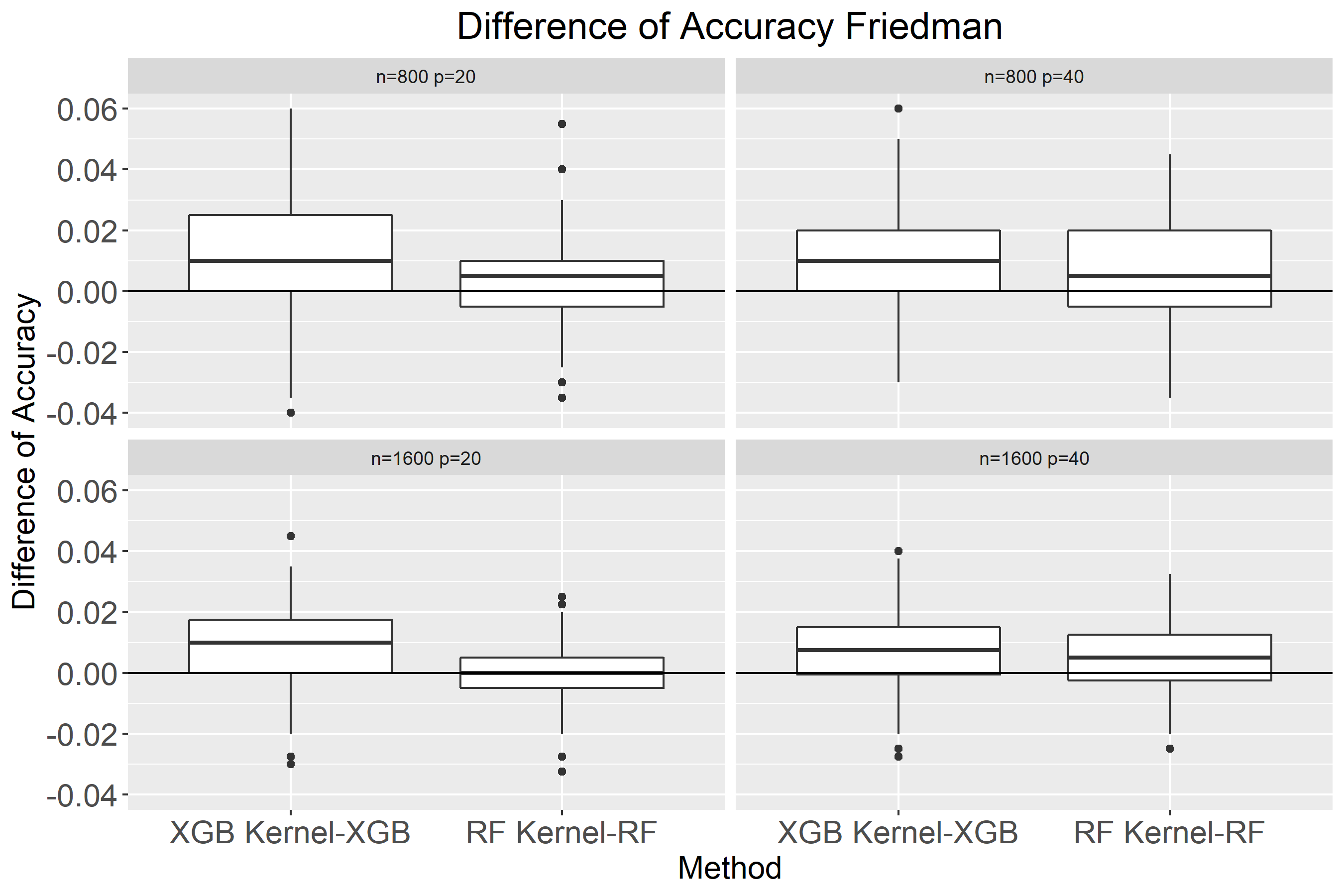} 
   \subcaption{}
  \label{fig:sub-first}
\end{subfigure}\\
\begin{subfigure}{.45\textwidth}
  \centering
  \includegraphics[height=0.20\textheight]{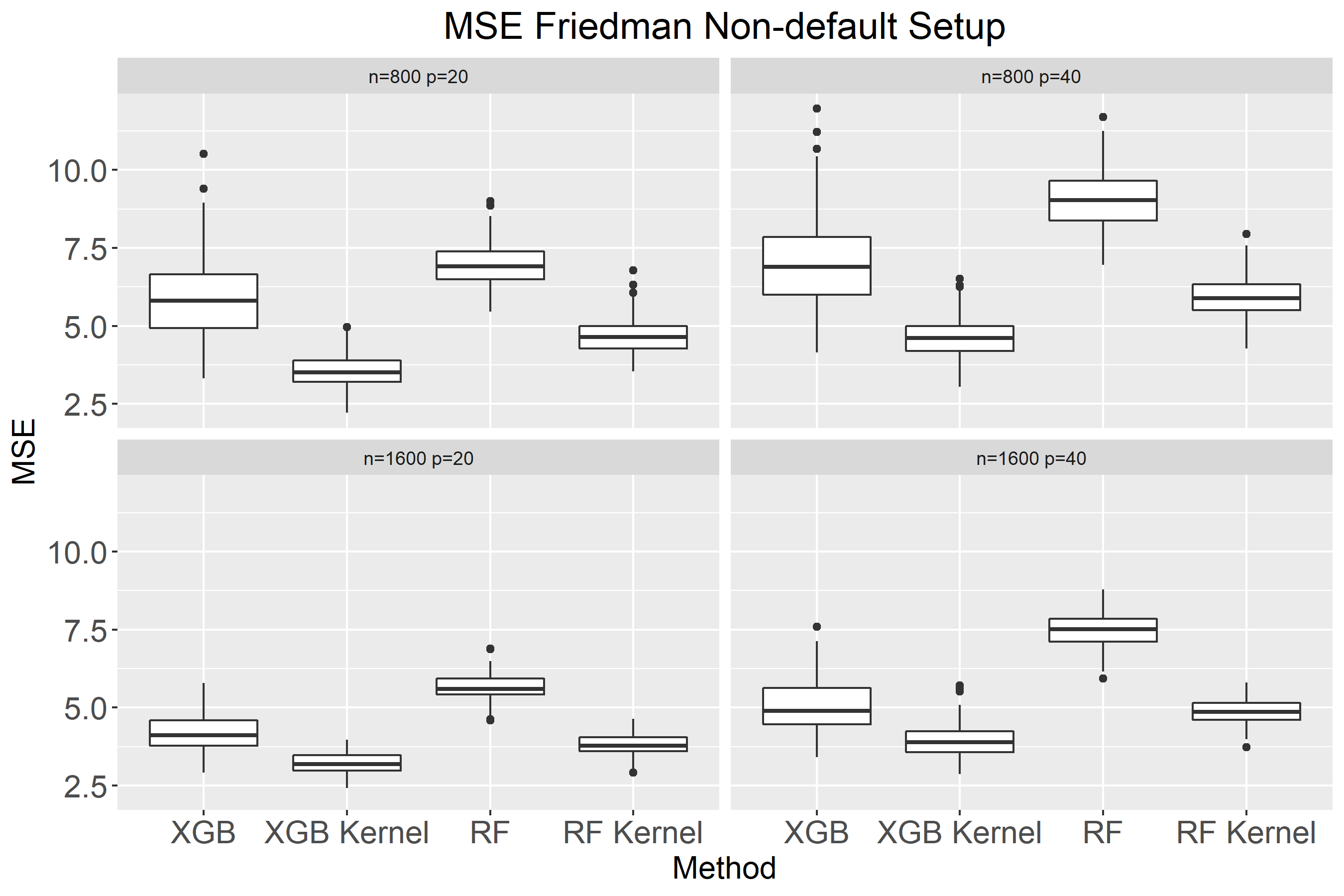}  
   \subcaption{}
  \label{fig:sub-first}
\end{subfigure}
\begin{subfigure}{.45\textwidth}
  \centering
  \includegraphics[height=0.2\textheight]{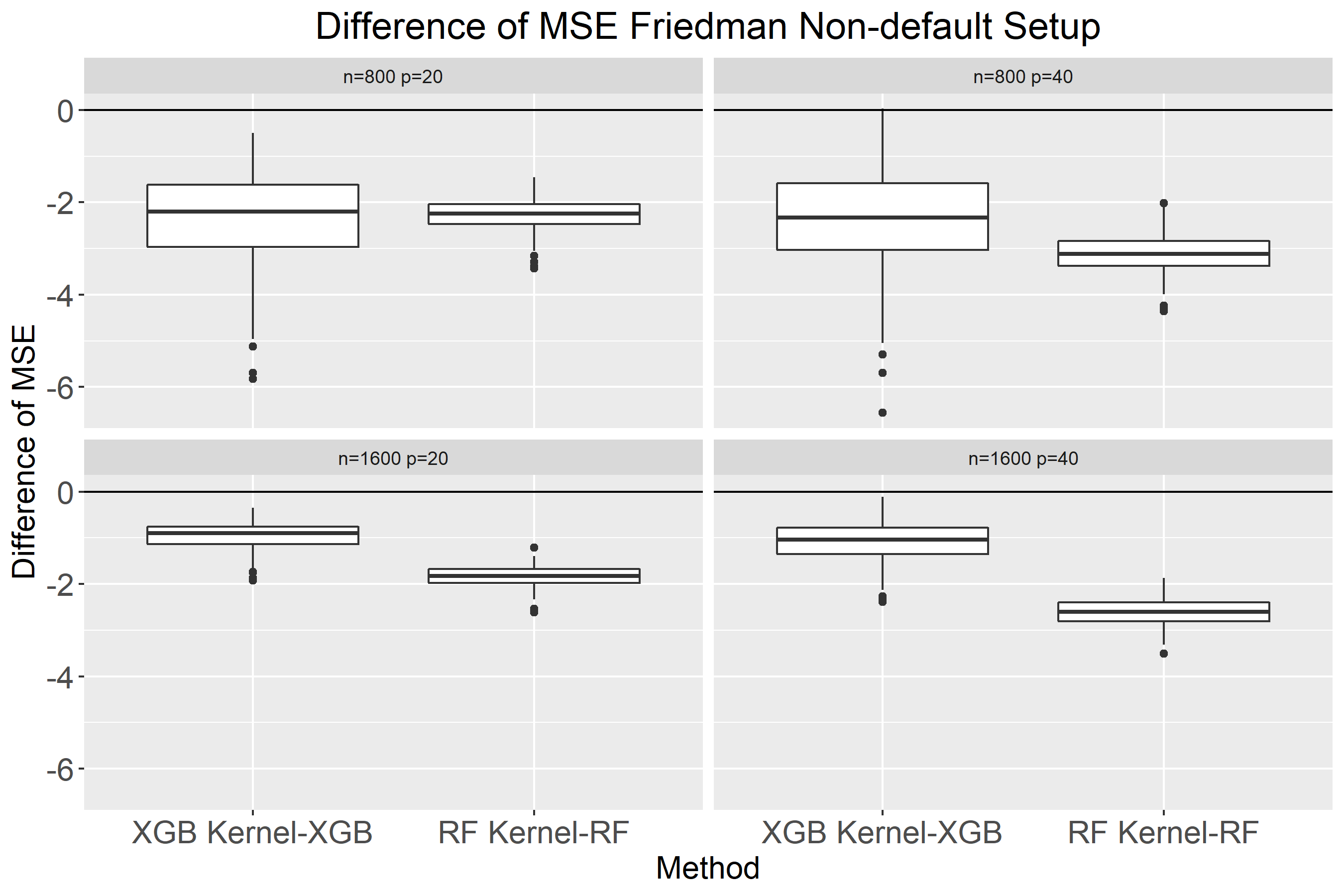}   
   \subcaption{}
  \label{fig:sub-first}
\end{subfigure}\\
\begin{subfigure}{.45\textwidth}
  \centering
  \includegraphics[height=0.2\textheight, width=0.9\textwidth]{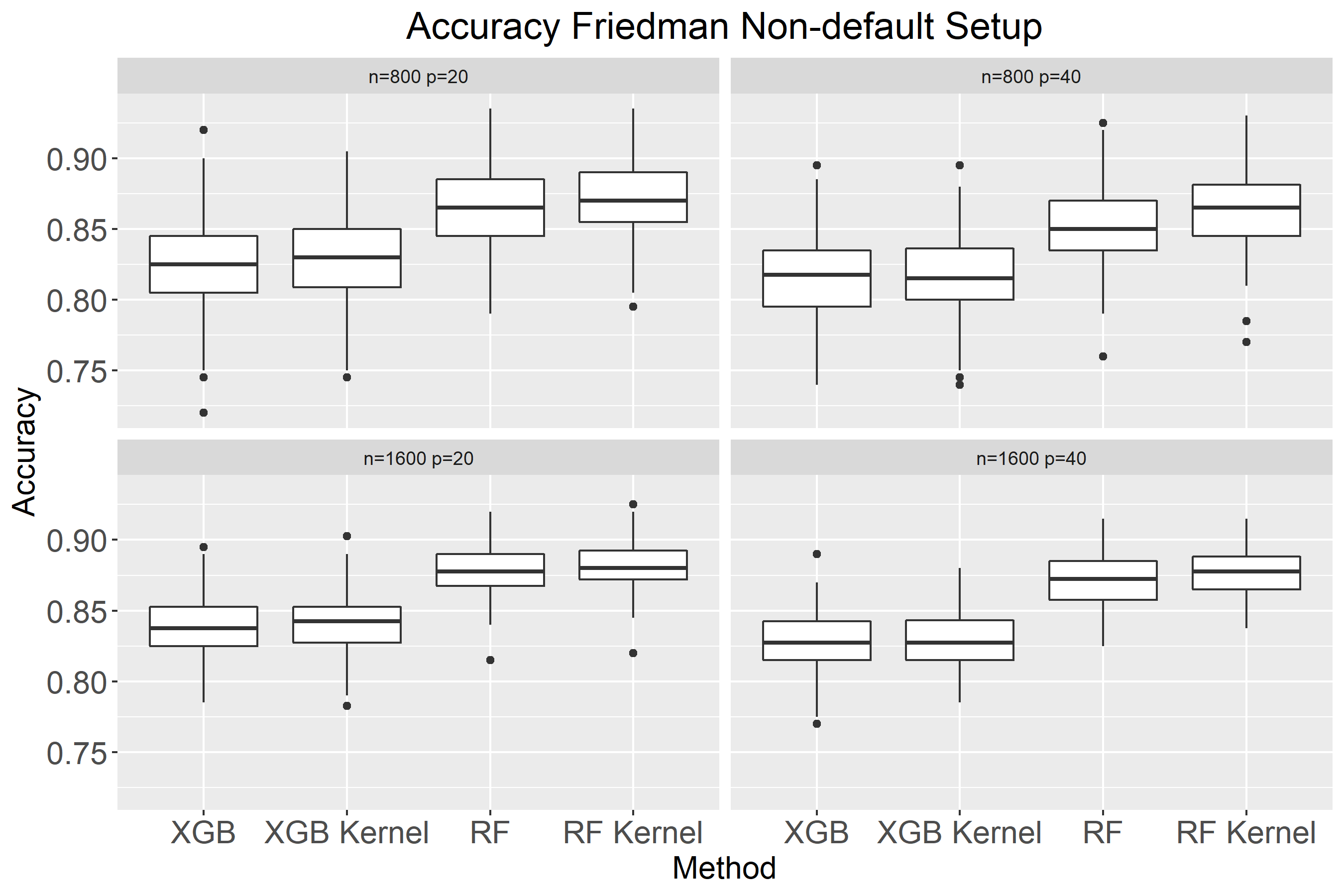} 
  \label{fig:sub-first}
    \subcaption{}
\end{subfigure}
\begin{subfigure}{.45\textwidth}
  \centering
  \includegraphics[height=0.2\textheight]{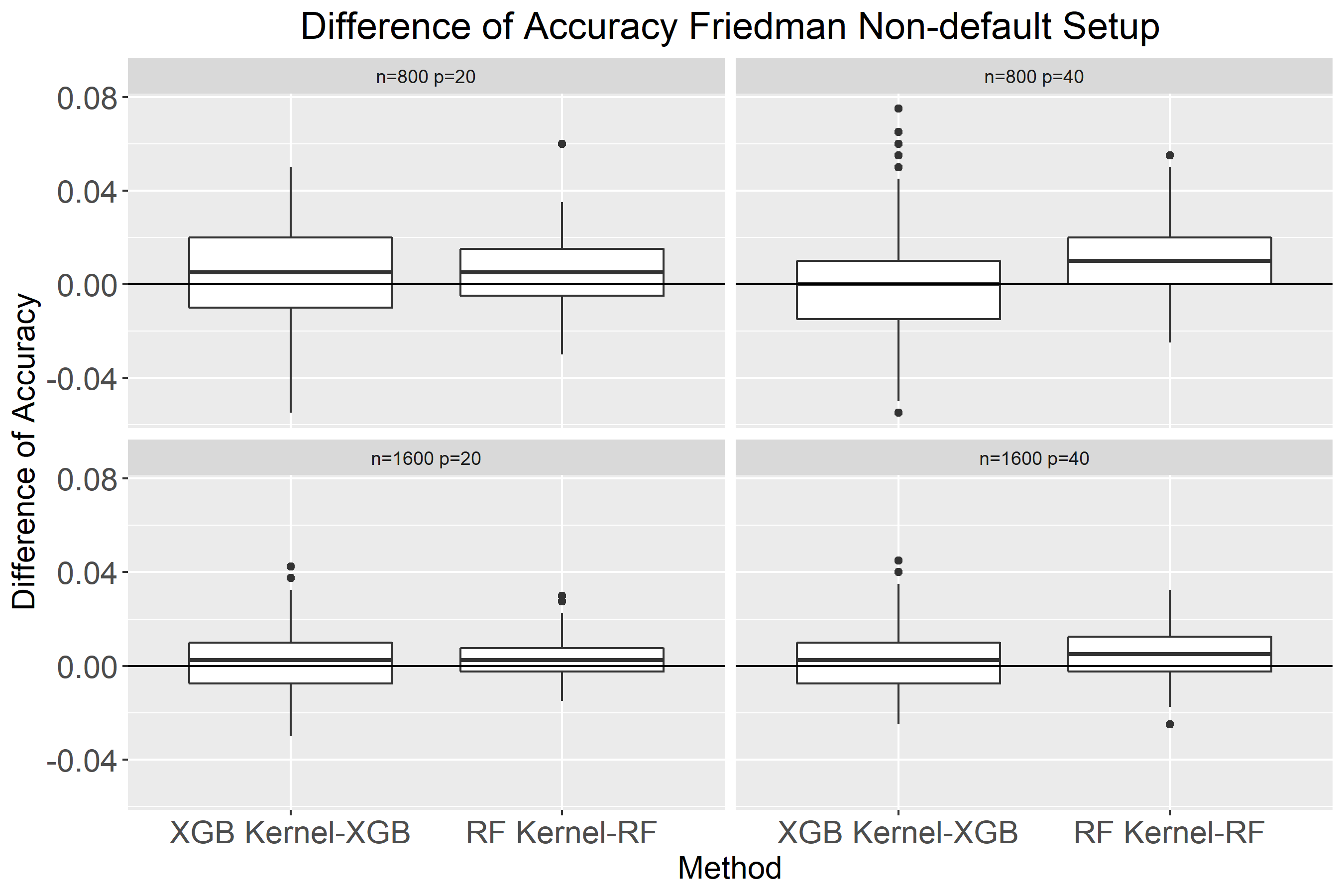} 
  \label{fig:sub-first}
   \subcaption{}
\end{subfigure}\\
\caption{Comparison of MSE and classification accuracy using RF, RF kernel, XGB, and XGB kernel using default and non-default setup in RF and XGB for data simulated from Friedman setting}
\label{fig:Friedman}
\end{figure}

\begin{figure}[ht]
\begin{subfigure}{.45\textwidth}
  \centering
  \includegraphics[height=0.20\textheight]{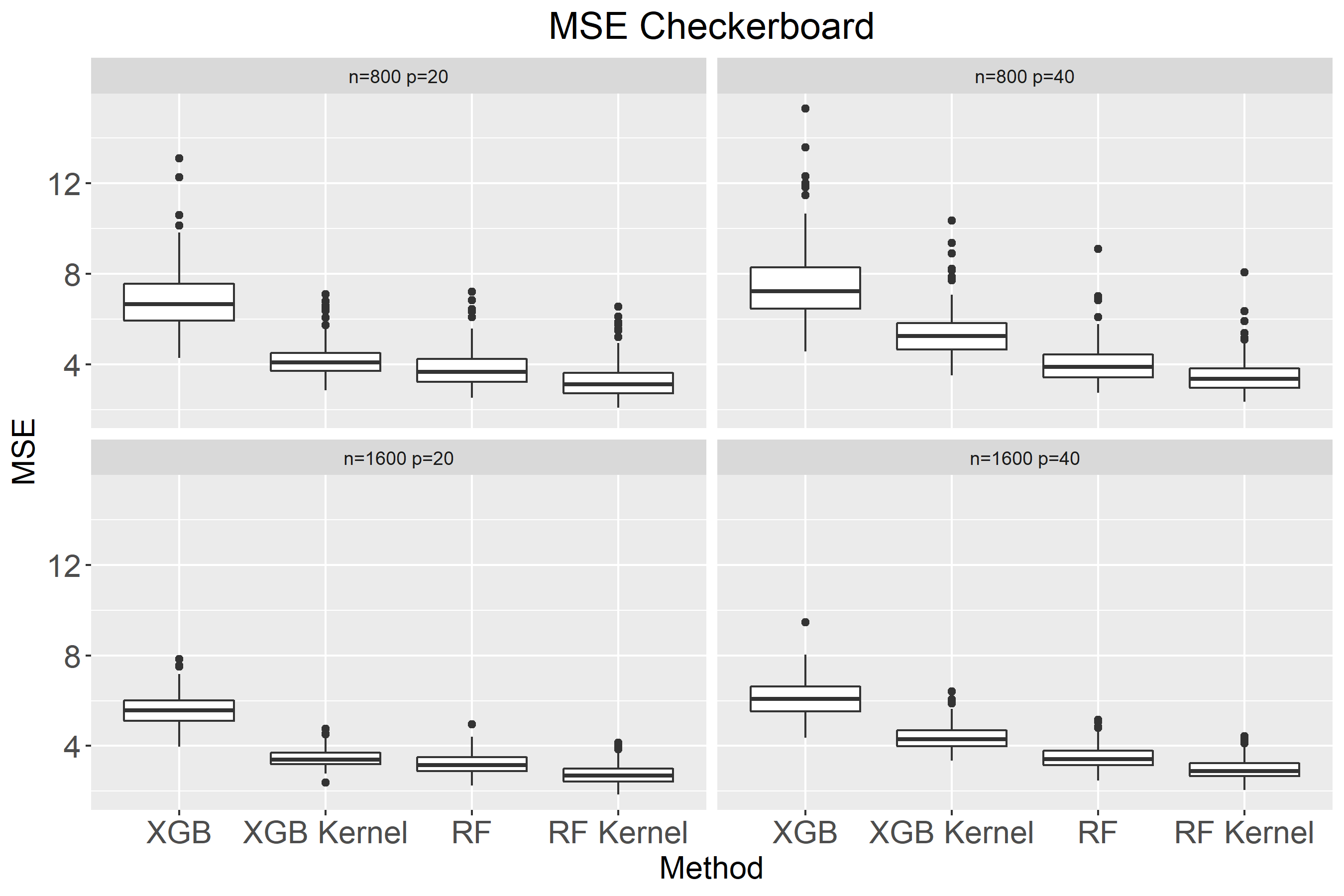}  
  \label{fig:sub-first}
    \subcaption{}
\end{subfigure}
\begin{subfigure}{.45\textwidth}
  \centering
  \includegraphics[height=0.2\textheight]{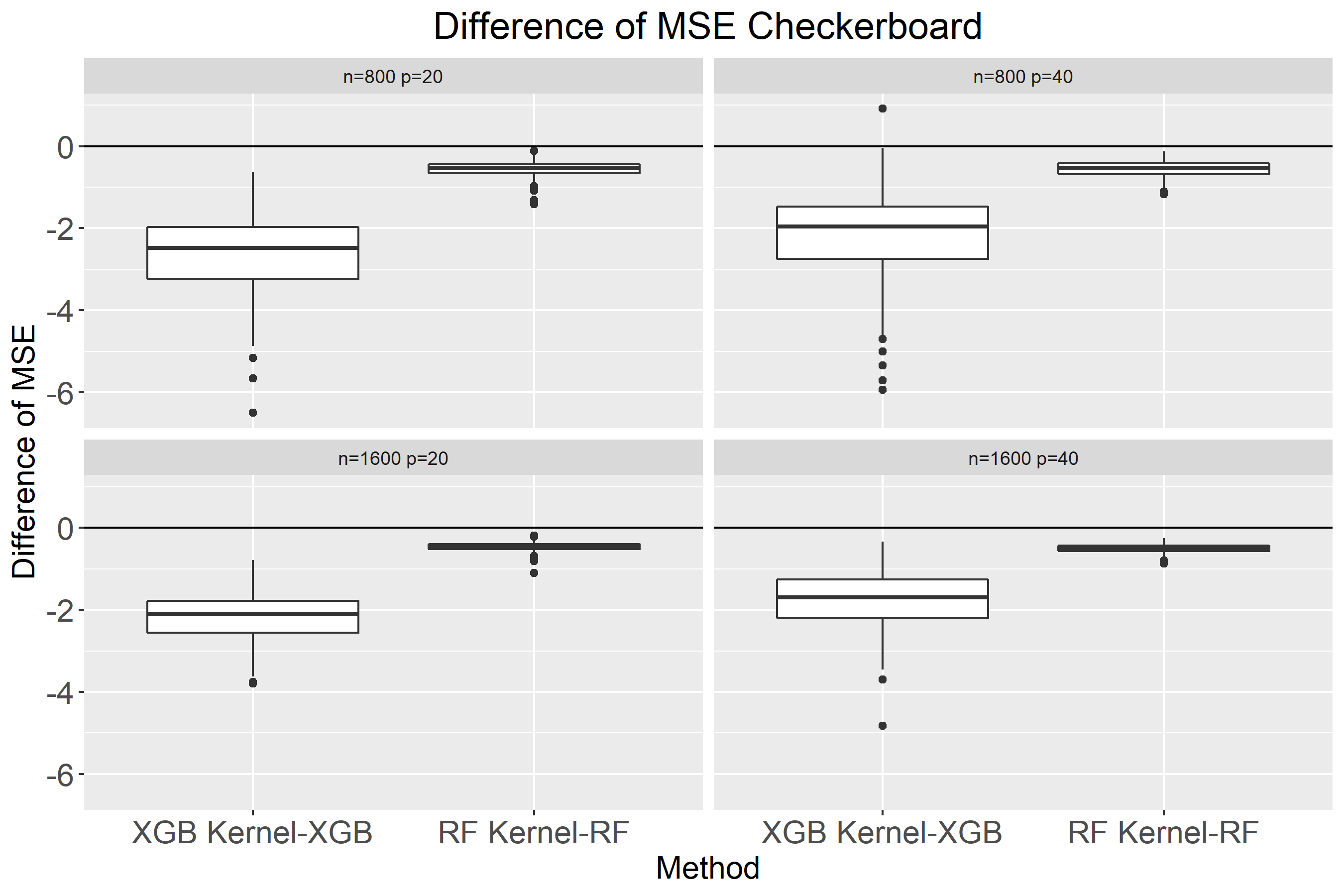}   
  \label{fig:sub-first}
    \subcaption{}
\end{subfigure}\\
\begin{subfigure}{.45\textwidth}
  \centering
  \includegraphics[height=0.2\textheight, width=0.9\textwidth]{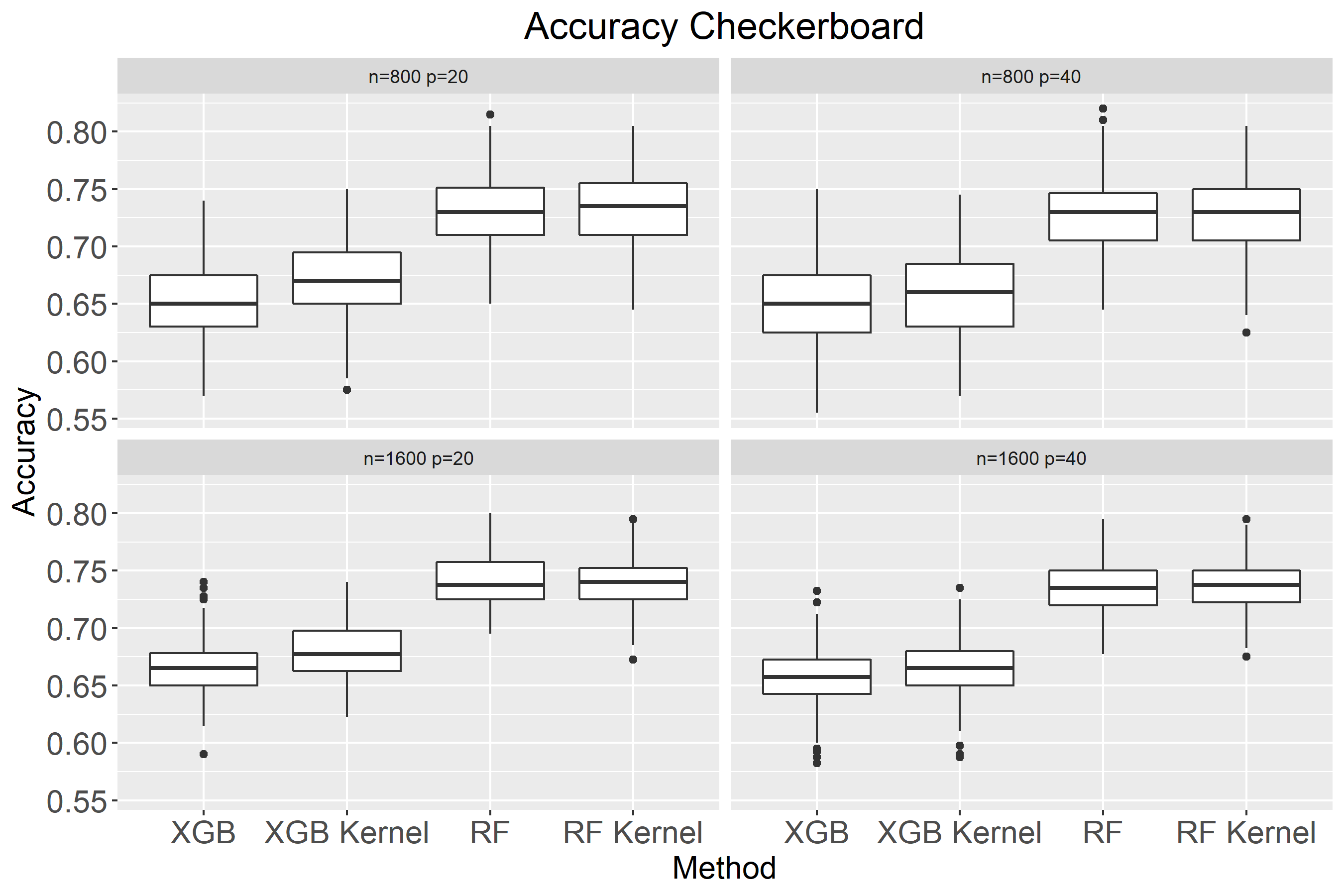}  
  \label{fig:sub-first}
    \subcaption{}
\end{subfigure}
\begin{subfigure}{.45\textwidth}
  \centering
  \includegraphics[height=0.2\textheight]{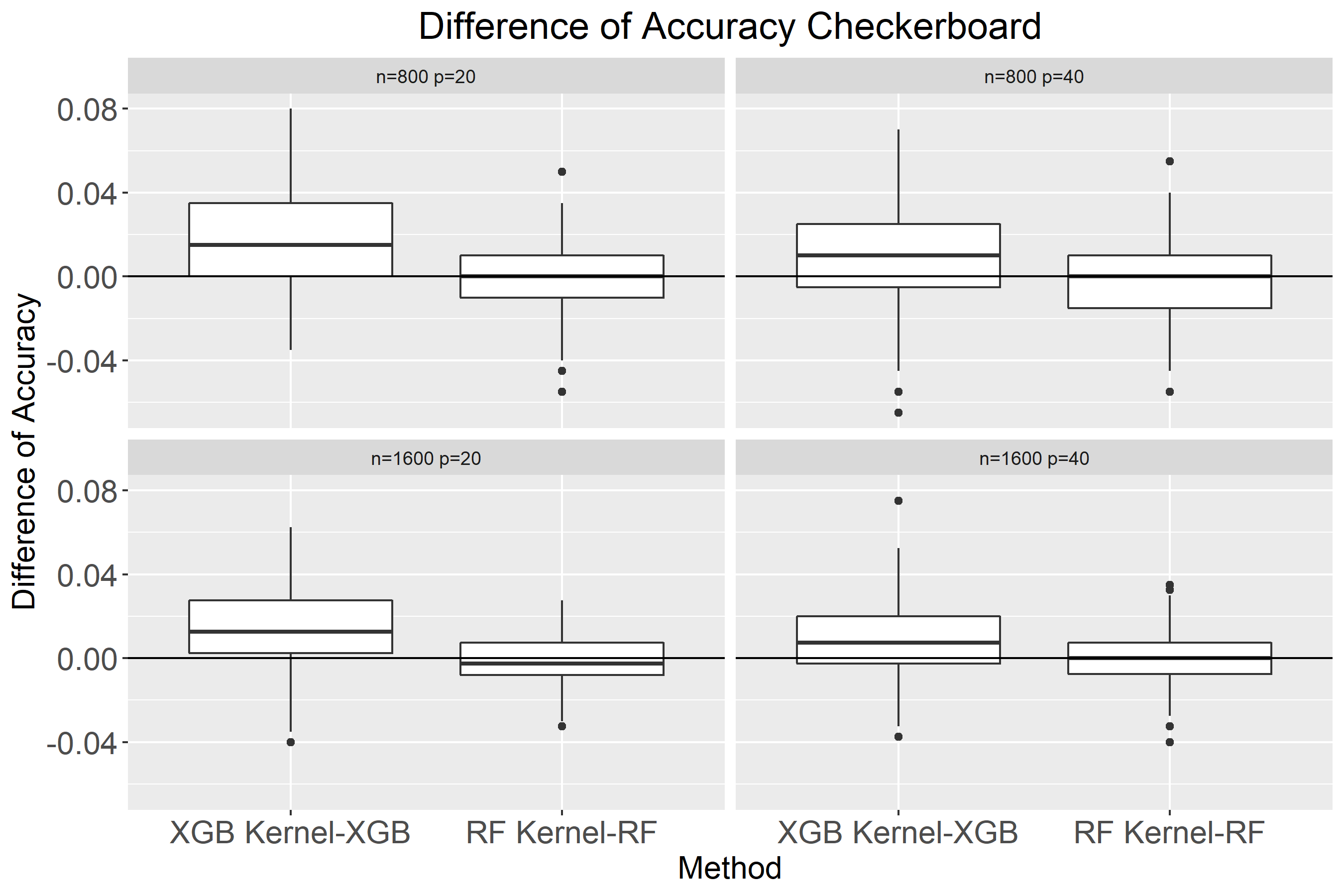} 
  \label{fig:sub-first}
    \subcaption{}
\end{subfigure}\\
\begin{subfigure}{.45\textwidth}
  \centering
  \includegraphics[height=0.20\textheight]{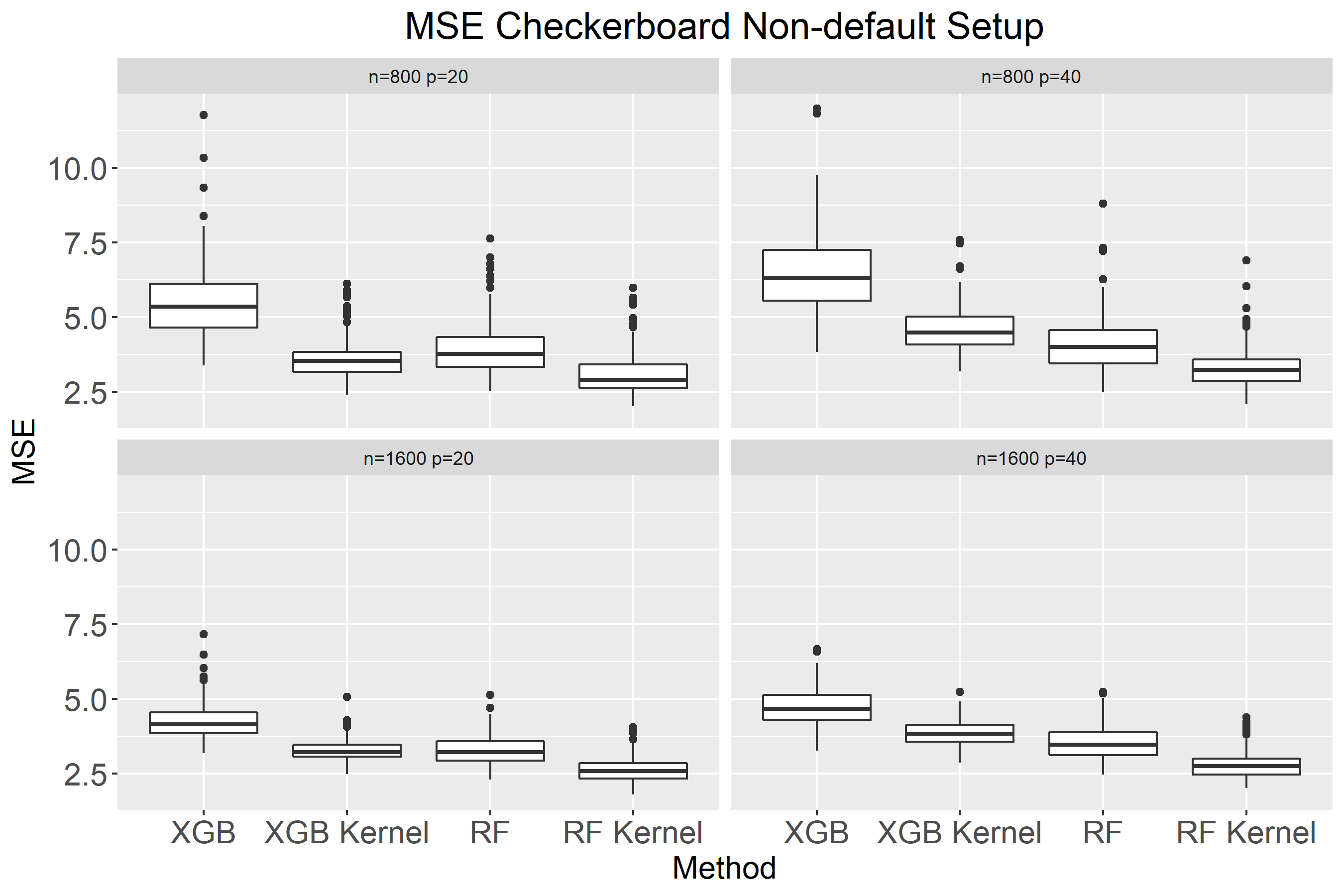}  
  \label{fig:sub-first}
    \subcaption{}
\end{subfigure}
\begin{subfigure}{.45\textwidth}
  \centering
  \includegraphics[height=0.2\textheight]{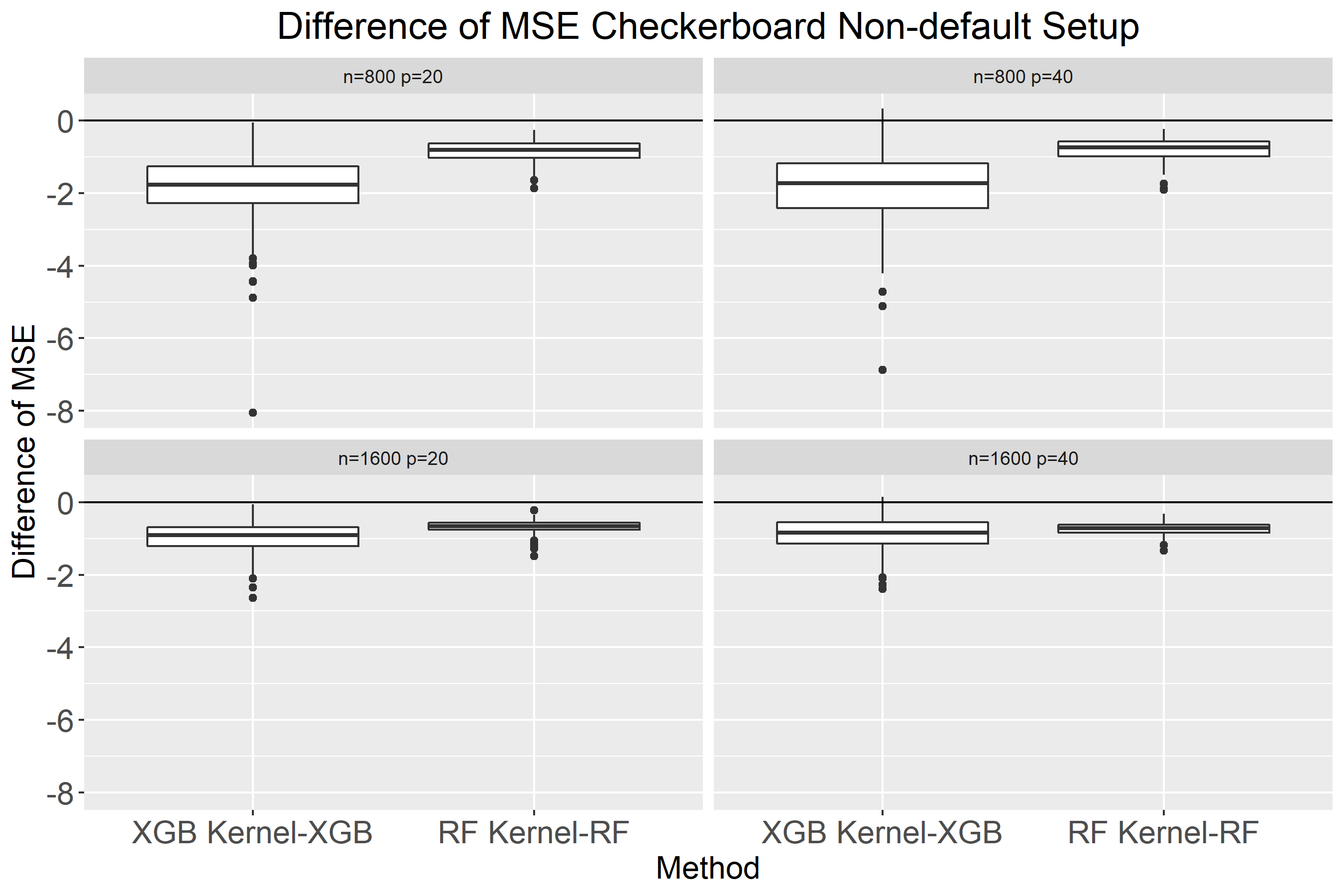}   
  \label{fig:sub-first}
    \subcaption{}
\end{subfigure}\\
\begin{subfigure}{.45\textwidth}
  \centering
  \includegraphics[height=0.2\textheight, width=0.9\textwidth]{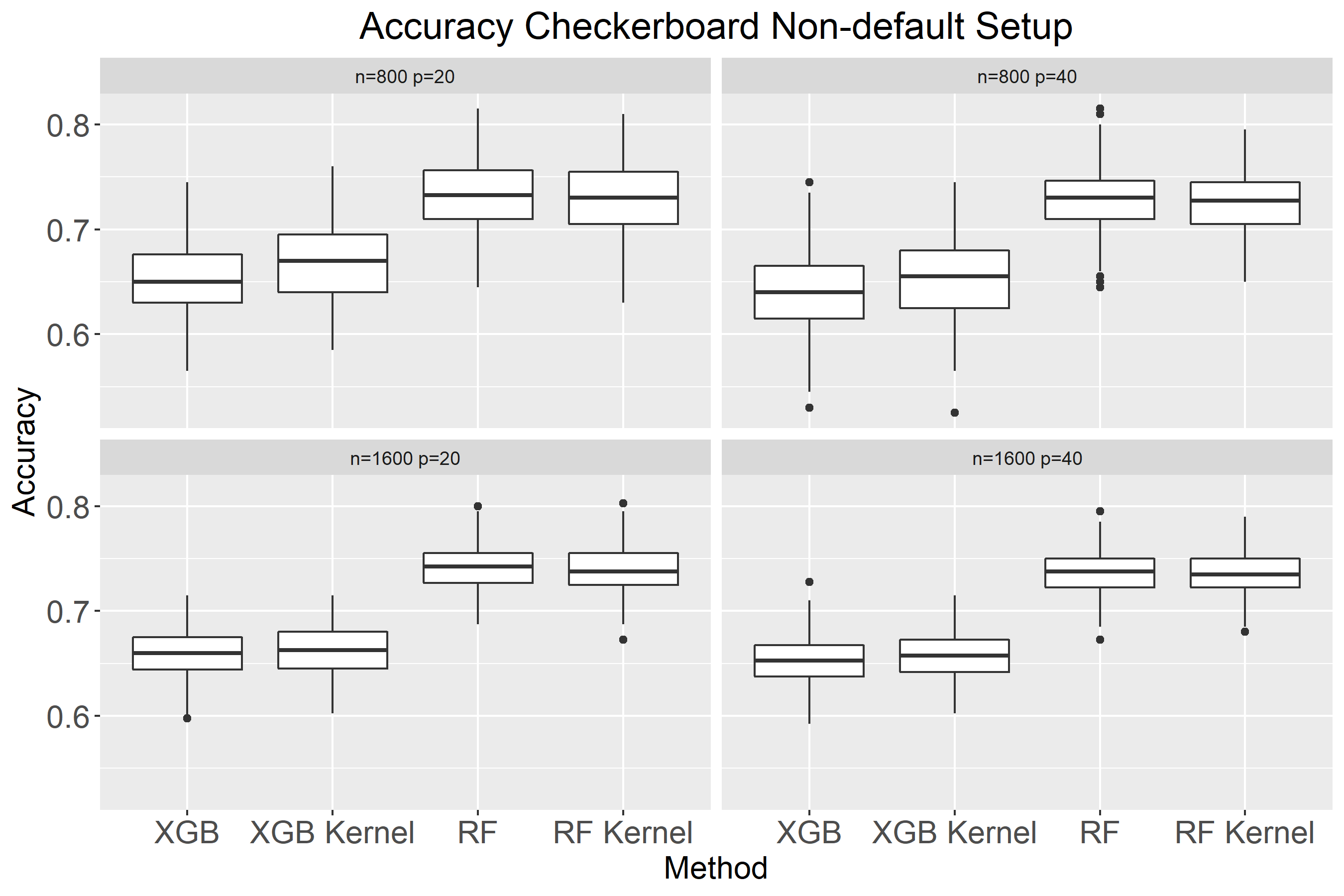} 
  \label{fig:sub-first}
    \subcaption{}
\end{subfigure}
\begin{subfigure}{.45\textwidth}
  \centering
  \includegraphics[height=0.2\textheight]{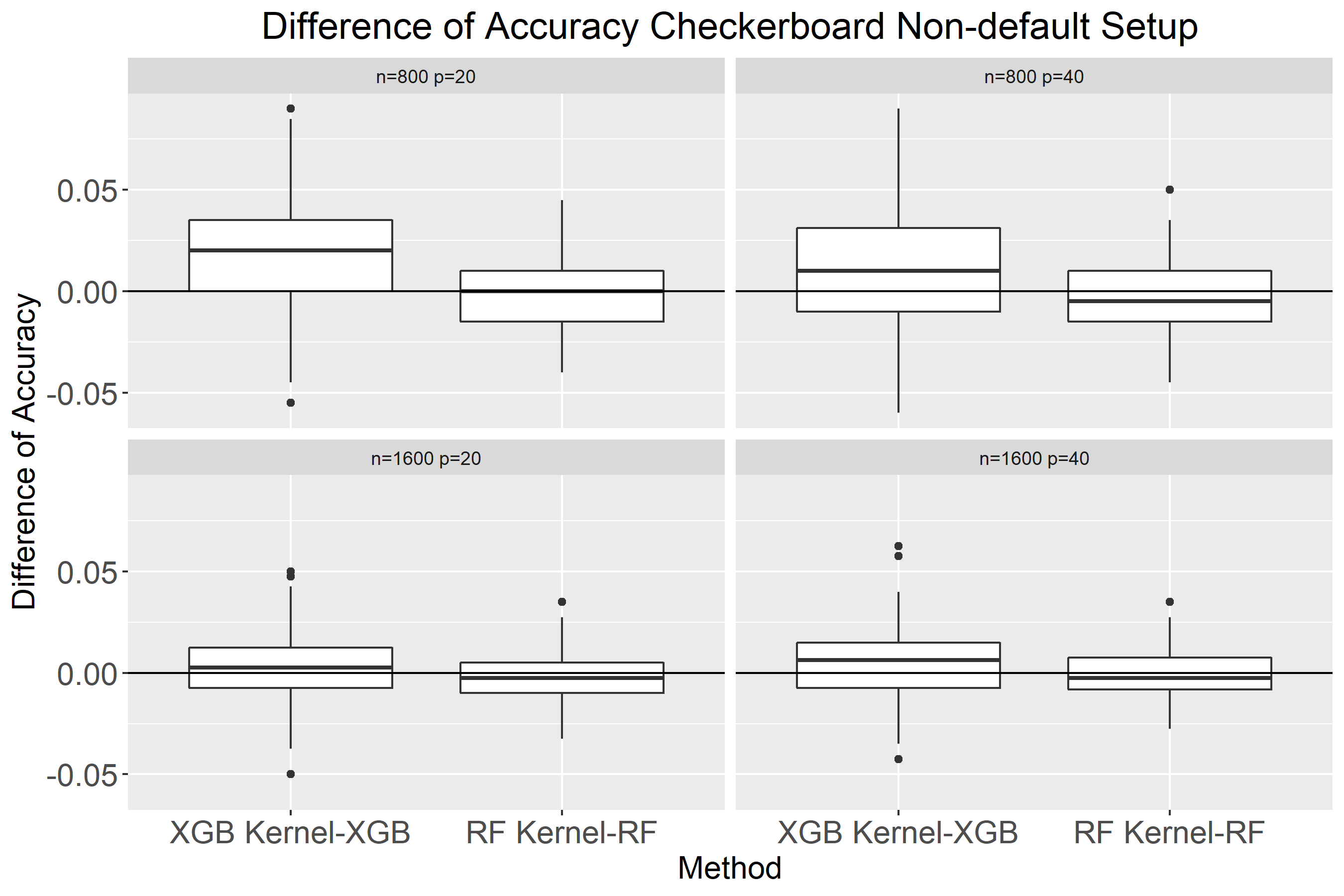} 
  \label{fig:sub-first}
    \subcaption{}
\end{subfigure}\\

\caption{Comparison of MSE and classification accuracy using RF, RF kernel, XGB, and XGB kernel using default and non-default setup in RF and XGB for data simulated from Checkerboard setting}
\label{fig:suppCheckerboard}
\end{figure}

\begin{figure}[ht]
\begin{subfigure}{.45\textwidth}
  \centering
  \includegraphics[height=0.20\textheight]{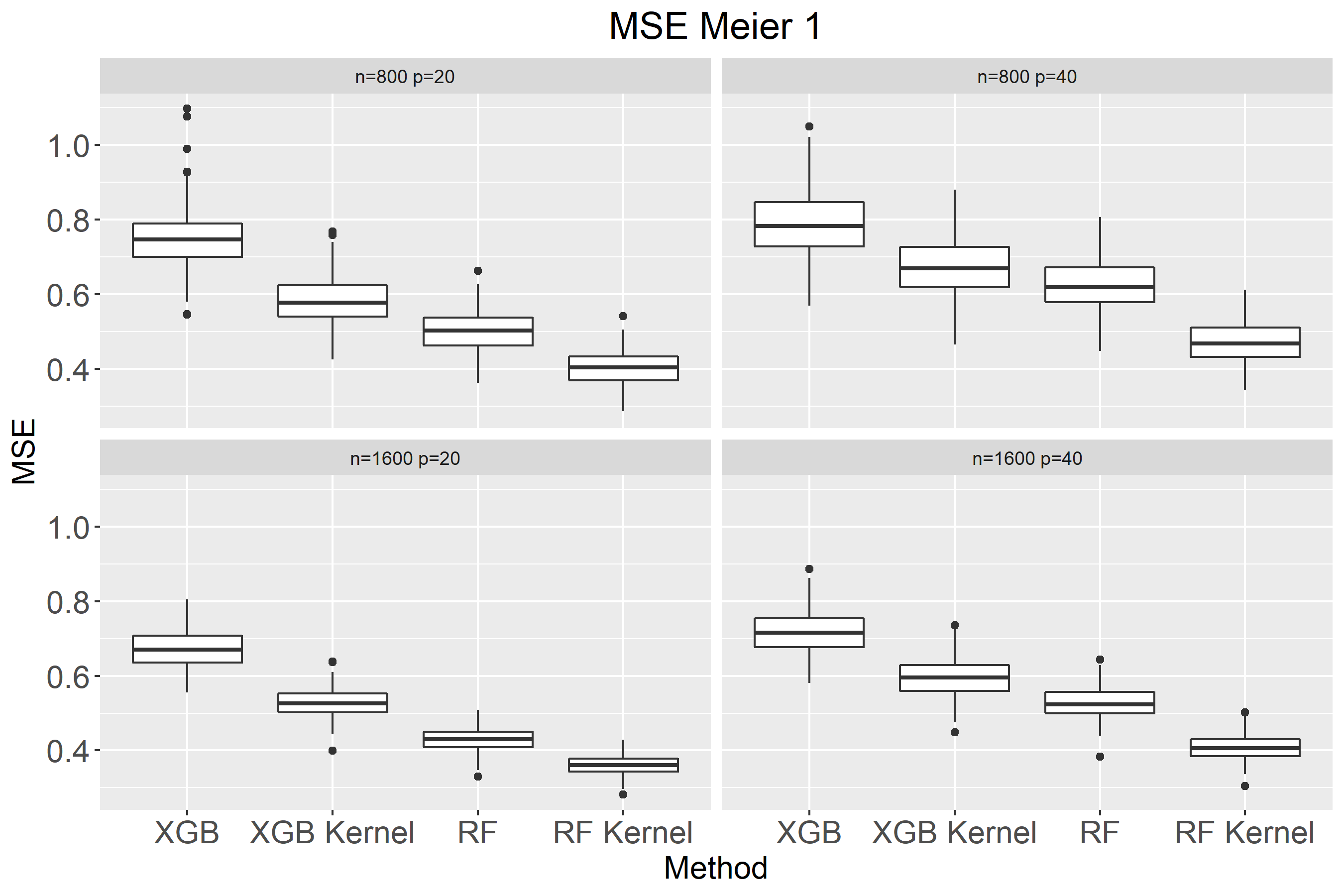}  
    \subcaption{}
  \label{fig:sub-first}
\end{subfigure}
\begin{subfigure}{.45\textwidth}
  \centering
  \includegraphics[height=0.2\textheight]{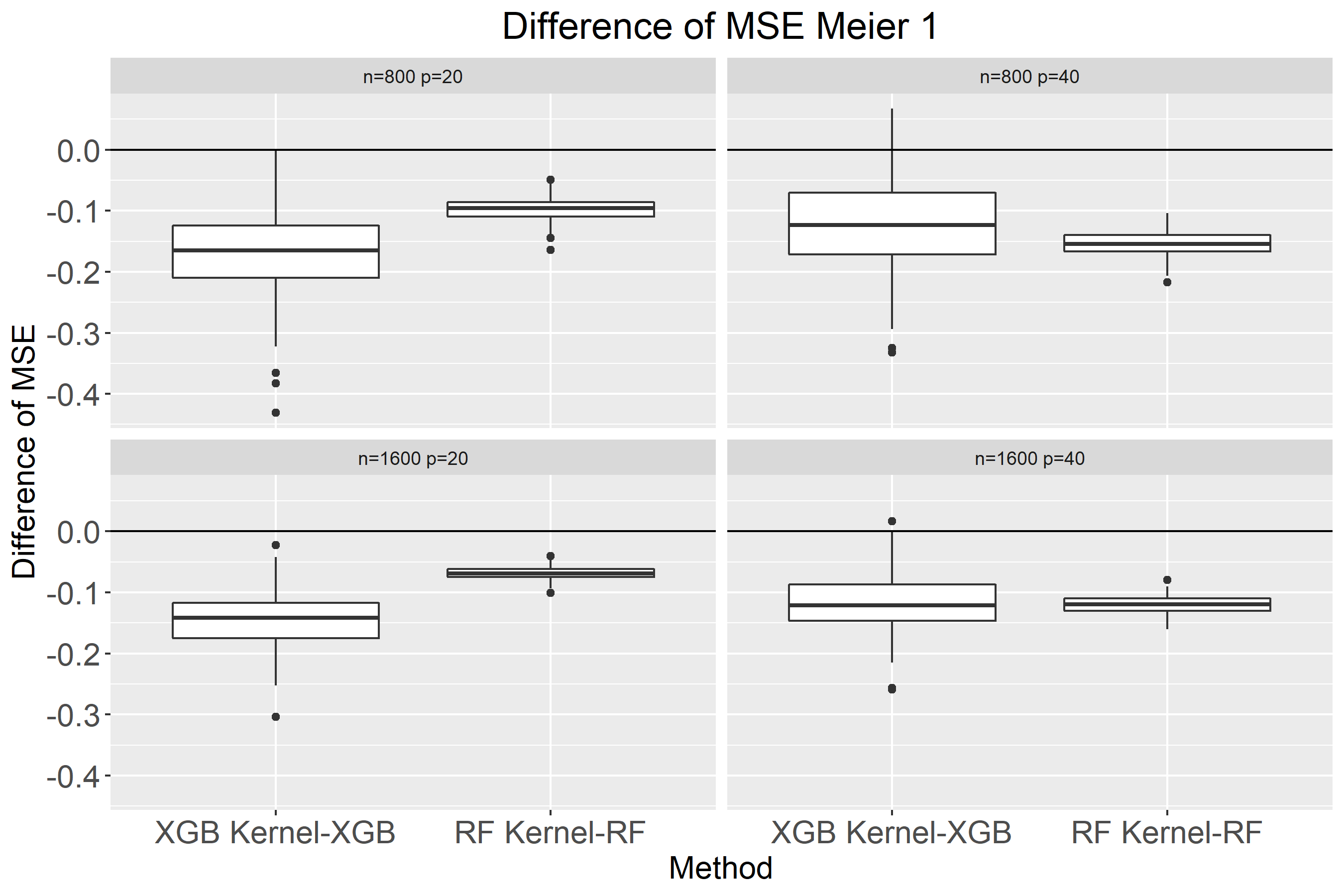}   
    \subcaption{}
  \label{fig:sub-first}
\end{subfigure}\\
\begin{subfigure}{.45\textwidth}
  \centering
  \includegraphics[height=0.2\textheight, width=0.9\textwidth]{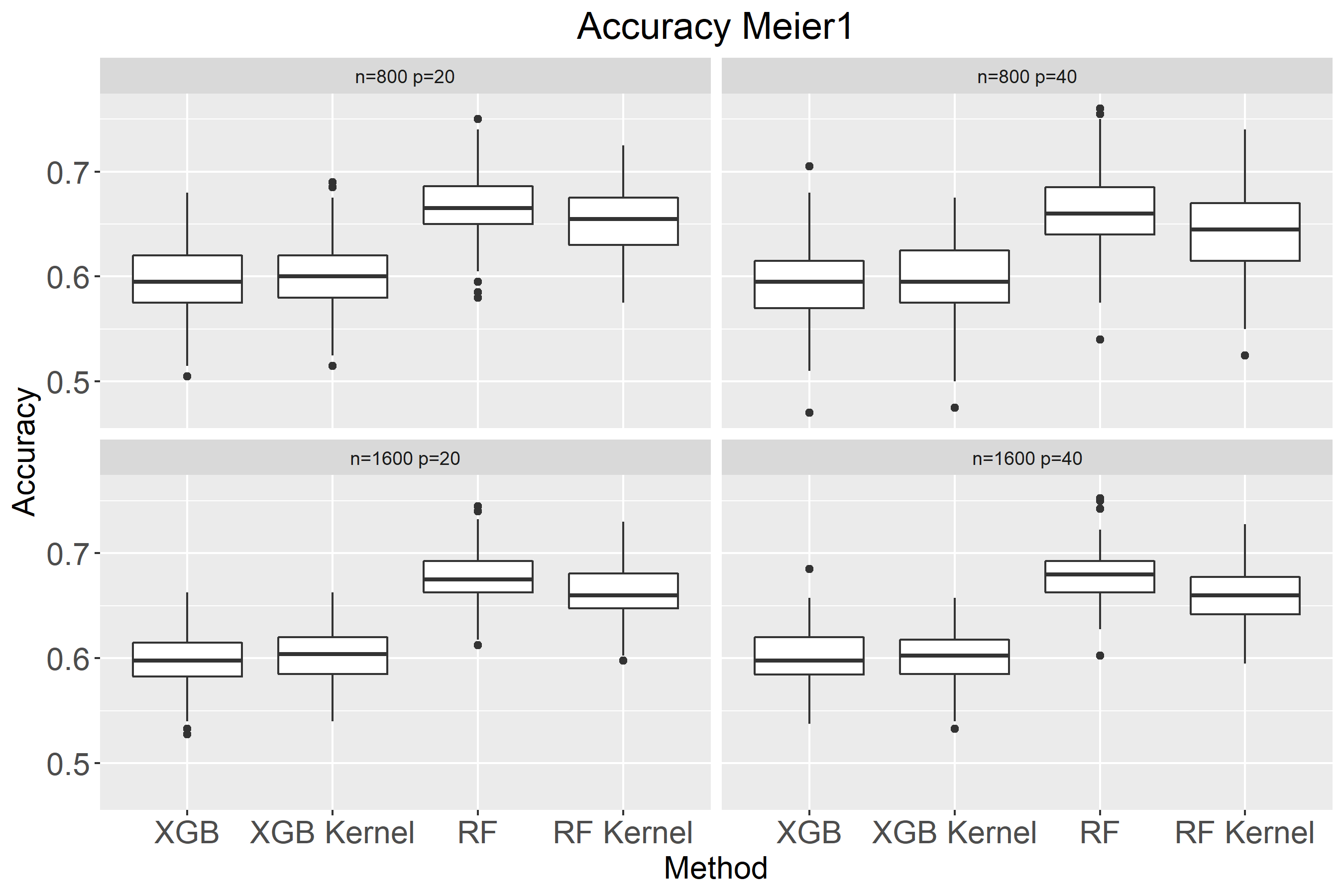}  
  \label{fig:sub-first}
    \subcaption{}
\end{subfigure}
\begin{subfigure}{.45\textwidth}
  \centering
  \includegraphics[height=0.2\textheight]{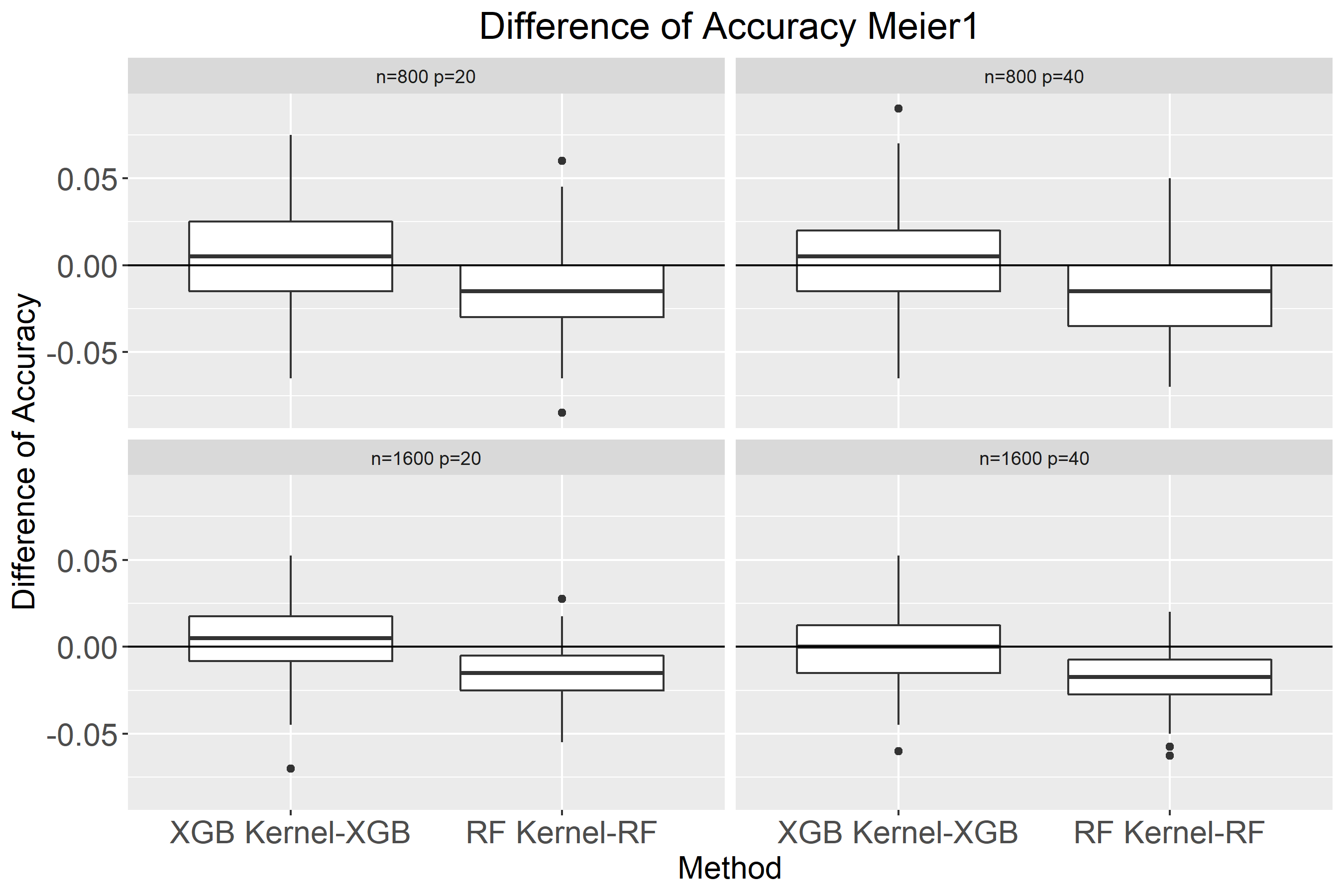} 
    \subcaption{}
  \label{fig:sub-first}
\end{subfigure}\\
\begin{subfigure}{.45\textwidth}
  \centering
  \includegraphics[height=0.20\textheight]{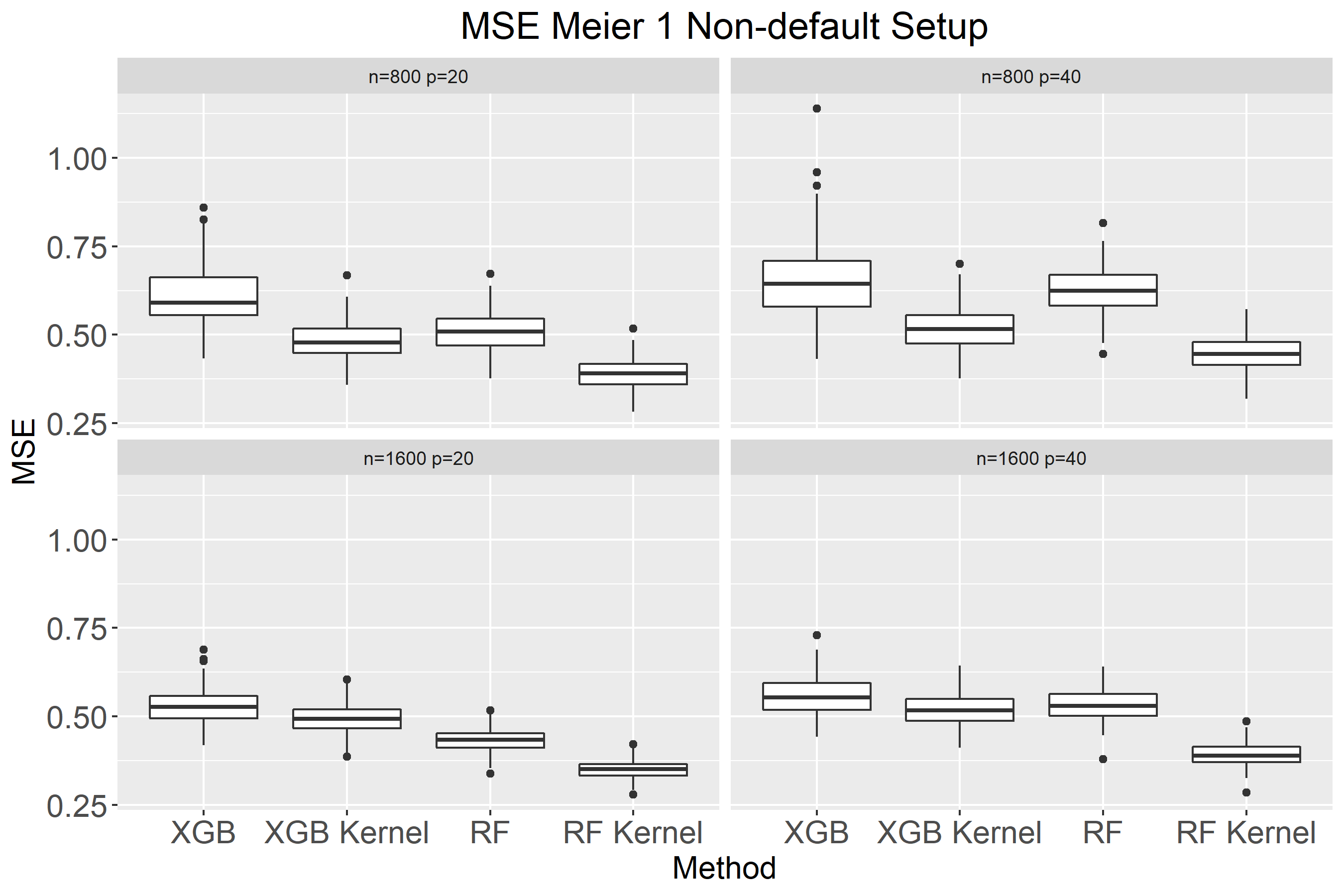}  
    \subcaption{}
  \label{fig:sub-first}
\end{subfigure}
\begin{subfigure}{.45\textwidth}
  \centering
  \includegraphics[height=0.2\textheight]{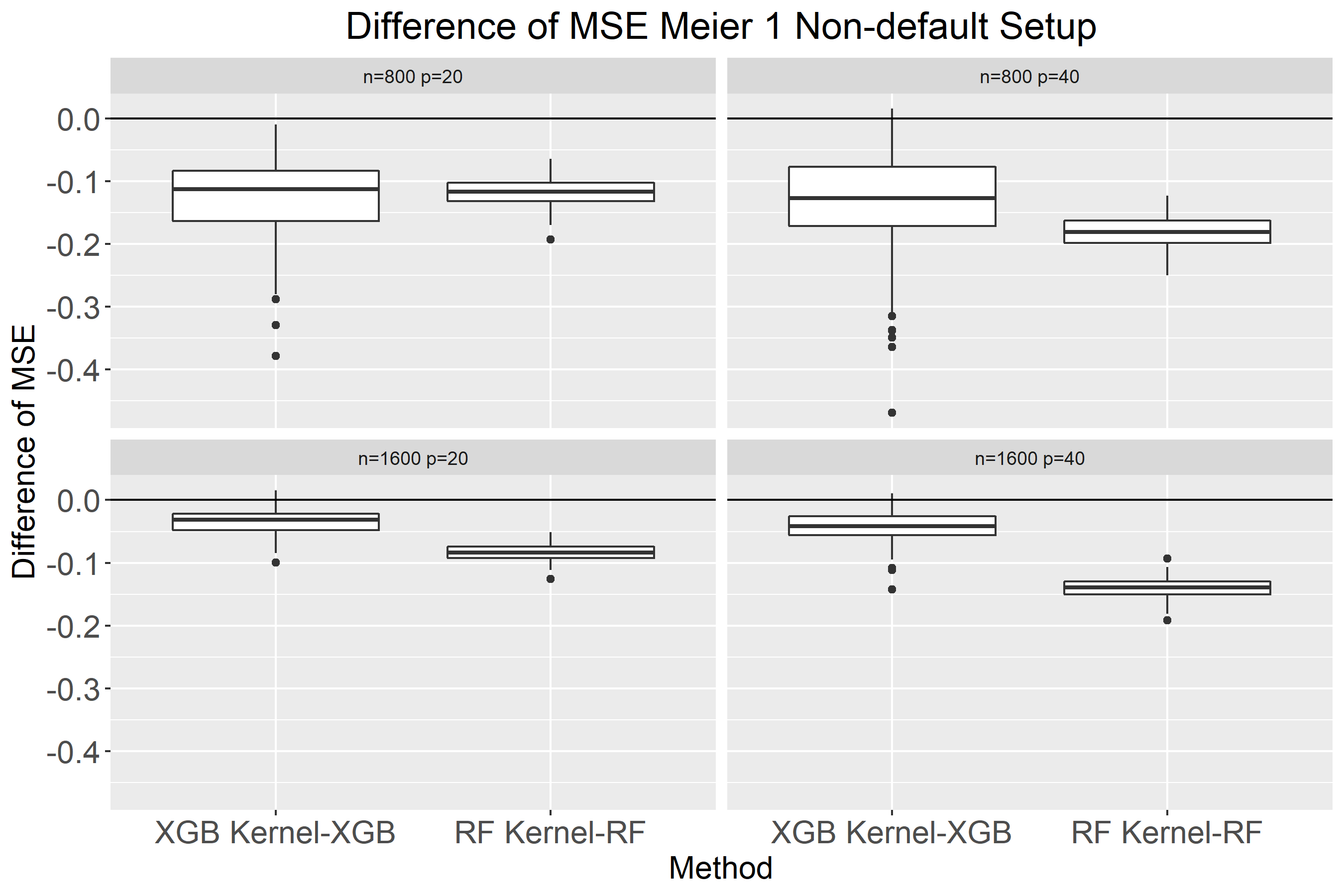}   
    \subcaption{}
  \label{fig:sub-first}
\end{subfigure}\\
\begin{subfigure}{.45\textwidth}
  \centering
  \includegraphics[height=0.2\textheight, width=0.9\textwidth]{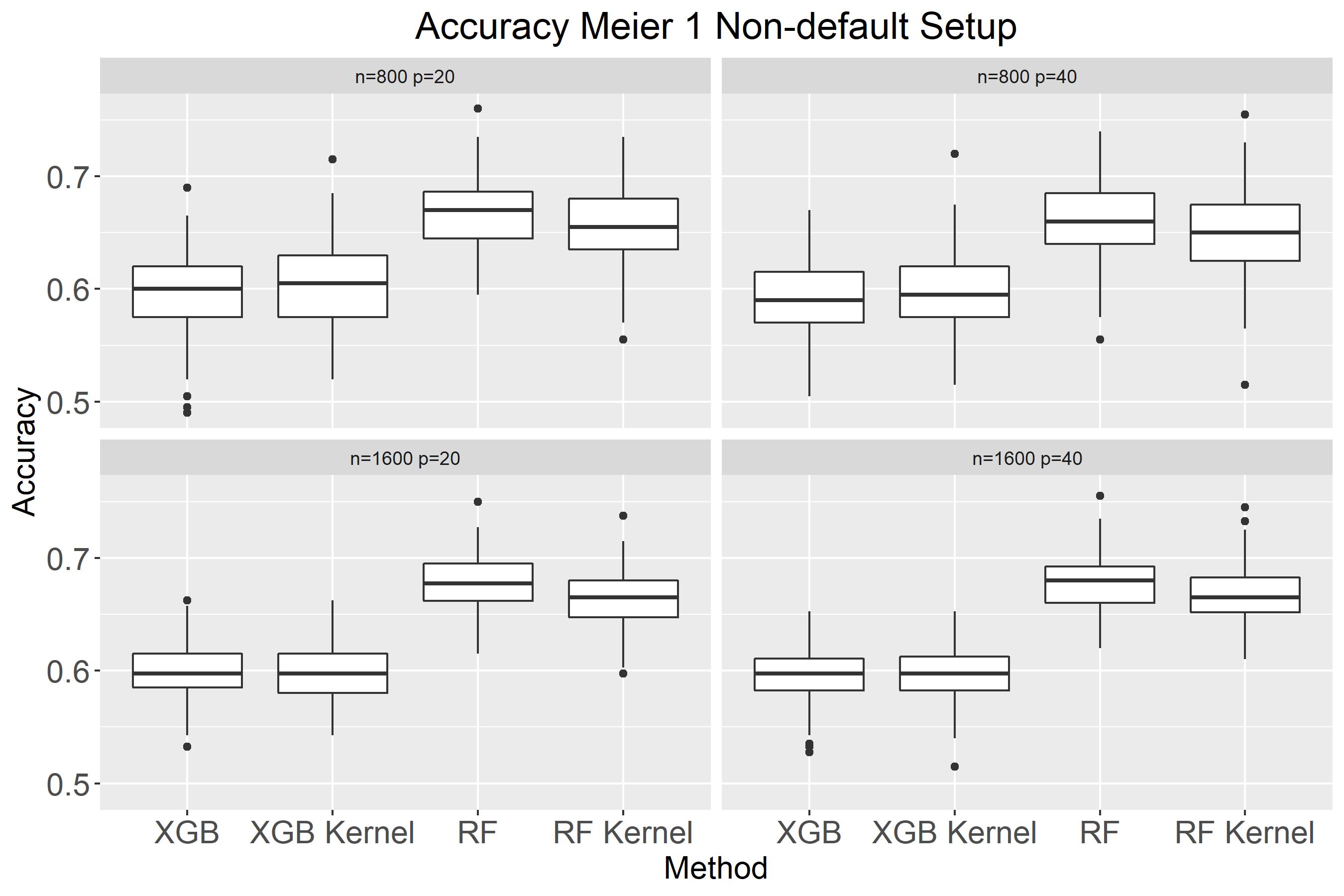} 
    \subcaption{}
  \label{fig:sub-first}
\end{subfigure}
\begin{subfigure}{.45\textwidth}
  \centering
  \includegraphics[height=0.2\textheight]{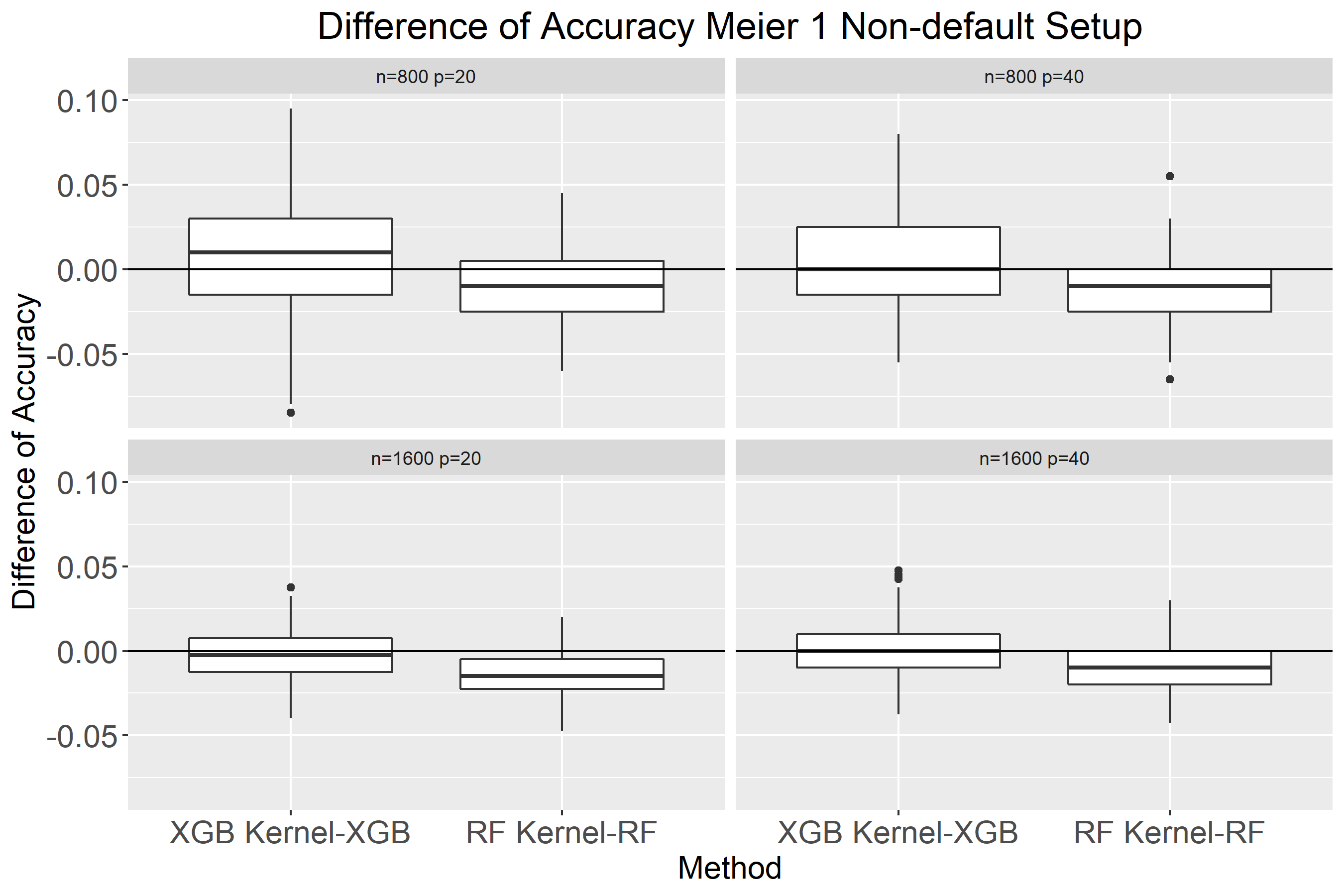} 
    \subcaption{}
  \label{fig:sub-first}
\end{subfigure}\\

\caption{Comparison of MSE and classification accuracy using RF, RF kernel, XGB, and XGB kernel using default and non-default setup in RF and XGB for data simulated from Meier 1 setting}
\label{fig:suppMeier1}
\end{figure}

\begin{figure}[ht]
\begin{subfigure}{.45\textwidth}
  \centering
  \includegraphics[height=0.20\textheight]{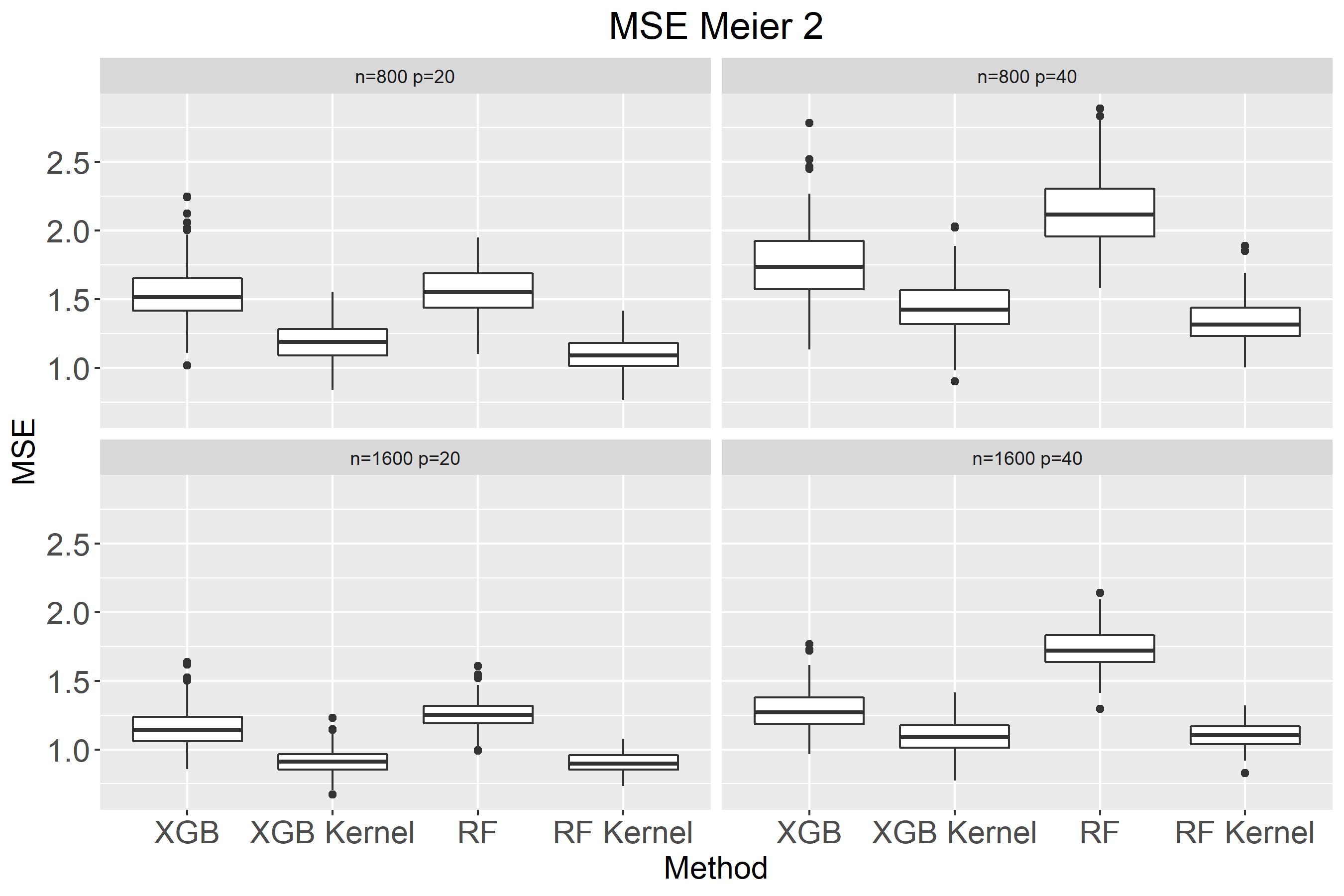}  
  \label{fig:sub-first}
    \subcaption{}
\end{subfigure}
\begin{subfigure}{.45\textwidth}
  \centering
  \includegraphics[height=0.2\textheight]{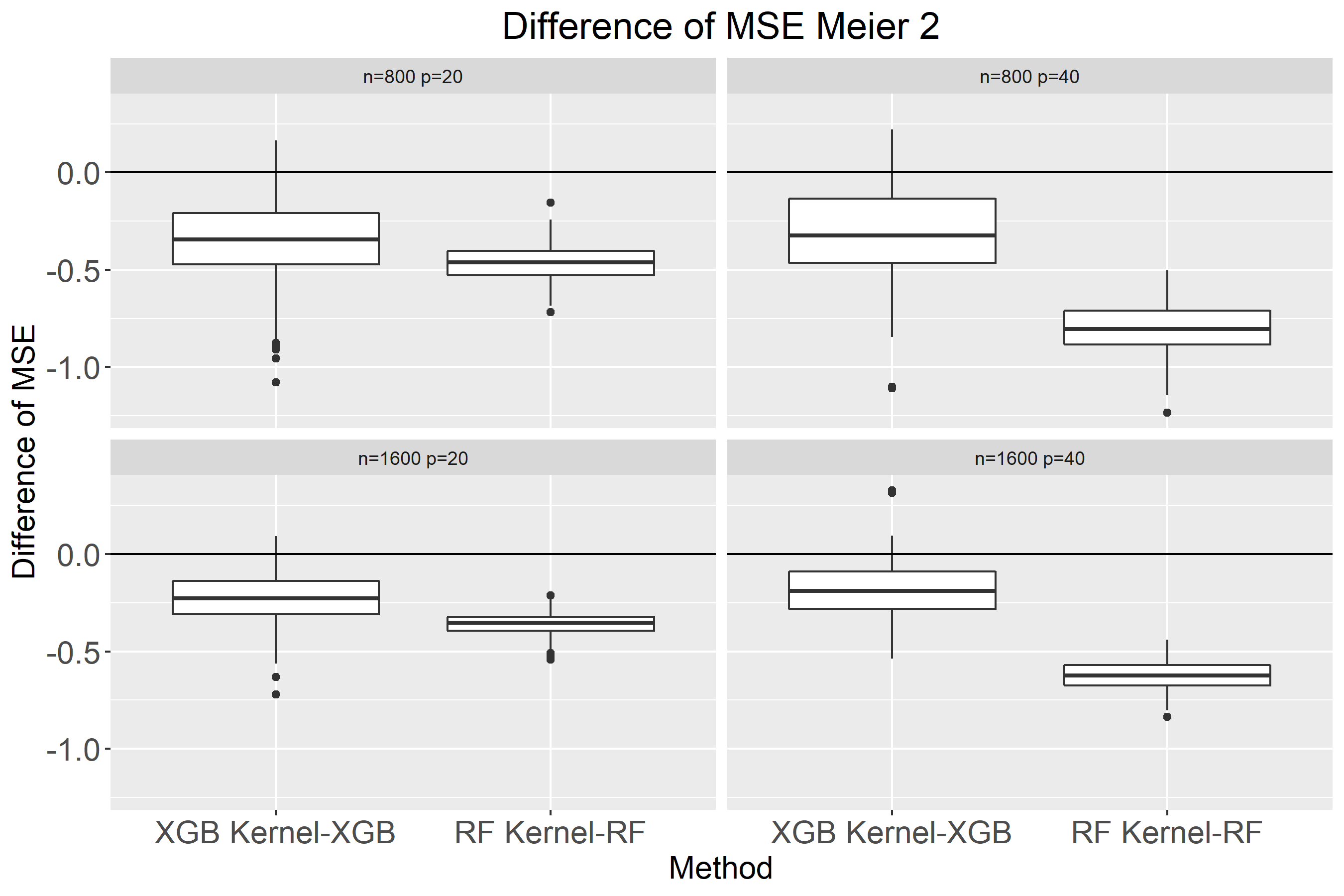}   
  \label{fig:sub-first}
    \subcaption{}
\end{subfigure}\\
\begin{subfigure}{.45\textwidth}
  \centering
  \includegraphics[height=0.2\textheight, width=0.9\textwidth]{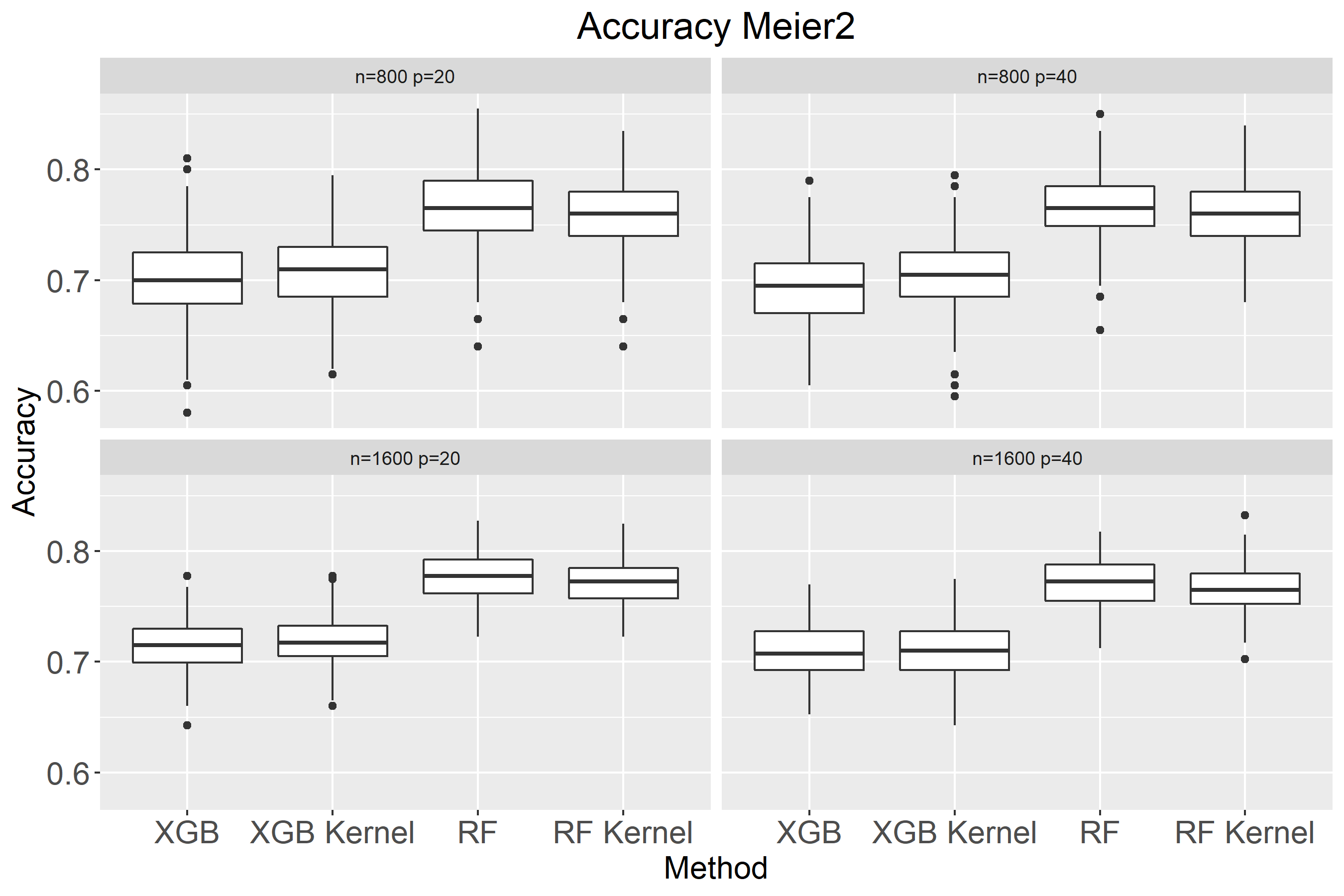}  
  \label{fig:sub-first}
    \subcaption{}
\end{subfigure}
\begin{subfigure}{.45\textwidth}
  \centering
  \includegraphics[height=0.2\textheight]{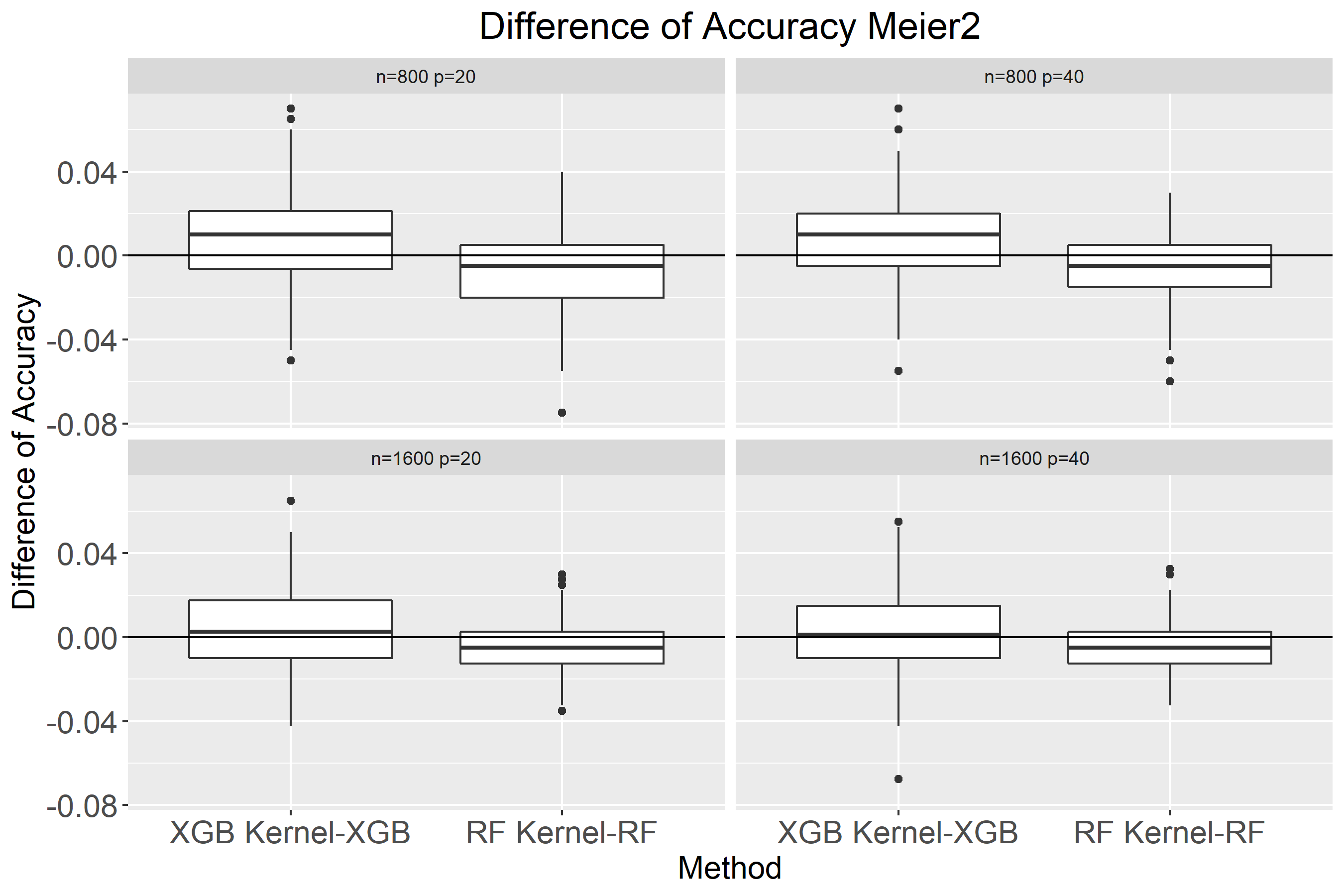} 
  \label{fig:sub-first}
    \subcaption{}
\end{subfigure}\\
\begin{subfigure}{.45\textwidth}
  \centering
  \includegraphics[height=0.20\textheight]{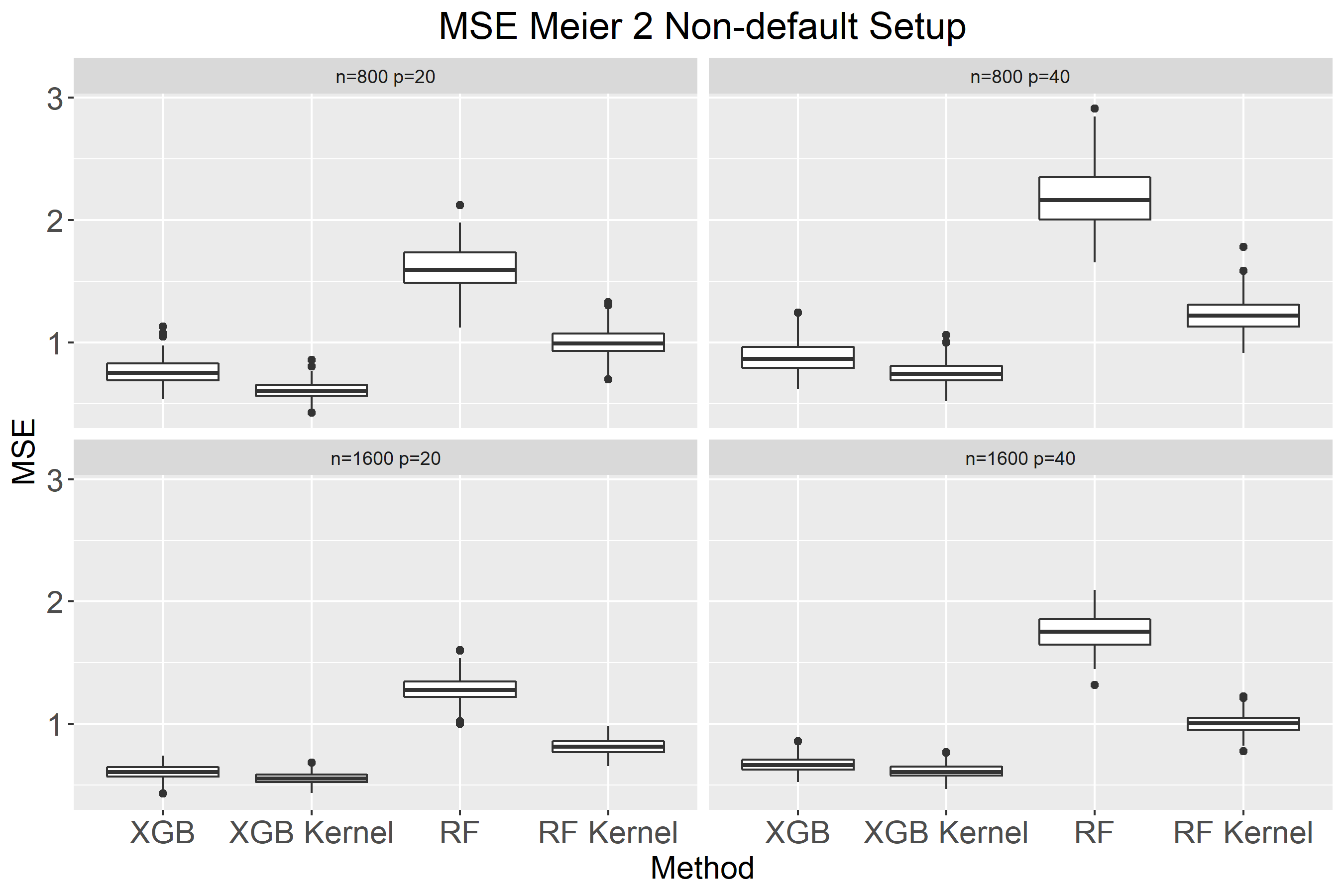}  
  \label{fig:sub-first}
    \subcaption{}
\end{subfigure}
\begin{subfigure}{.45\textwidth}
  \centering
  \includegraphics[height=0.2\textheight]{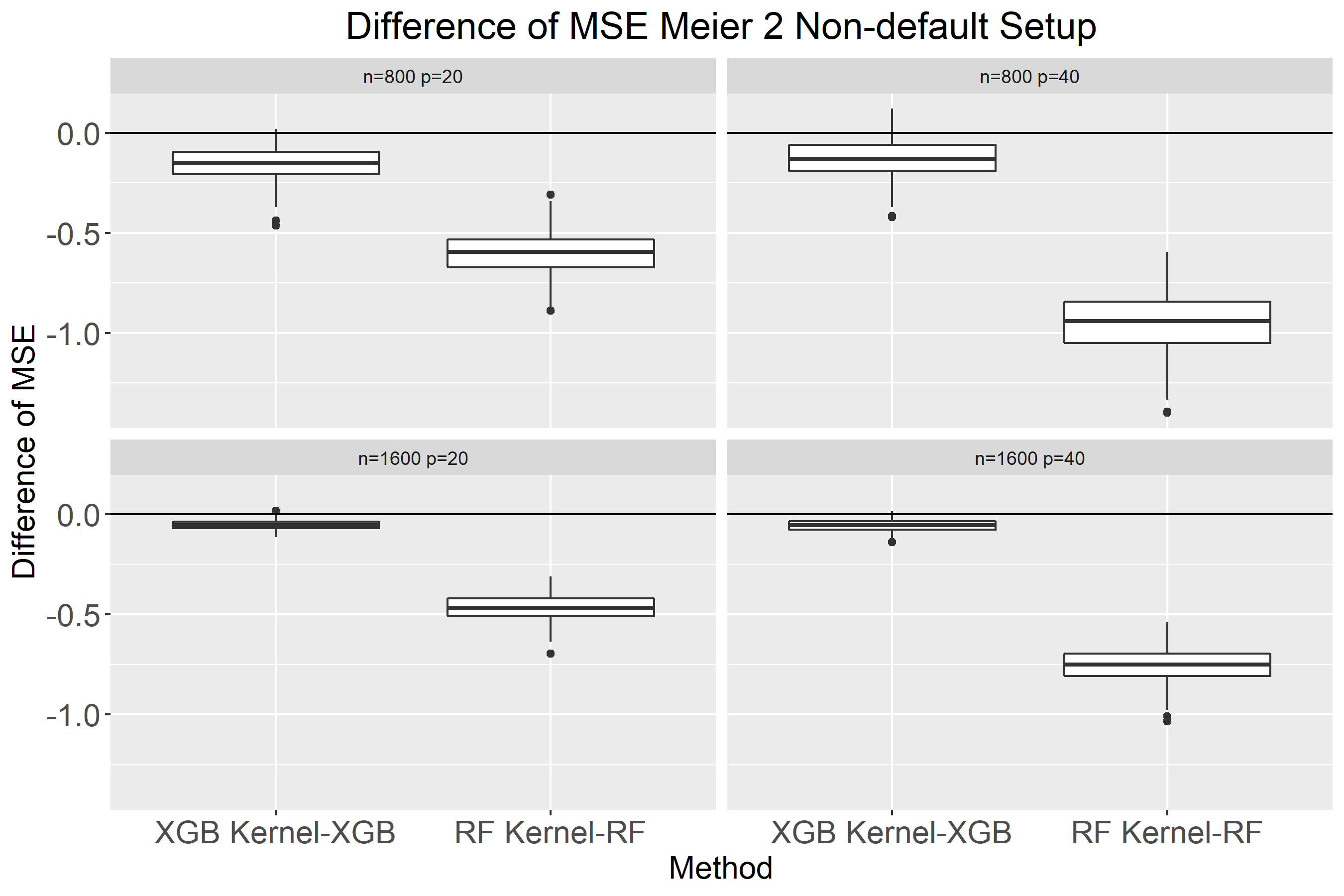}   
  \label{fig:sub-first}
    \subcaption{}
\end{subfigure}\\
\begin{subfigure}{.45\textwidth}
  \centering
  \includegraphics[height=0.2\textheight, width=0.9\textwidth]{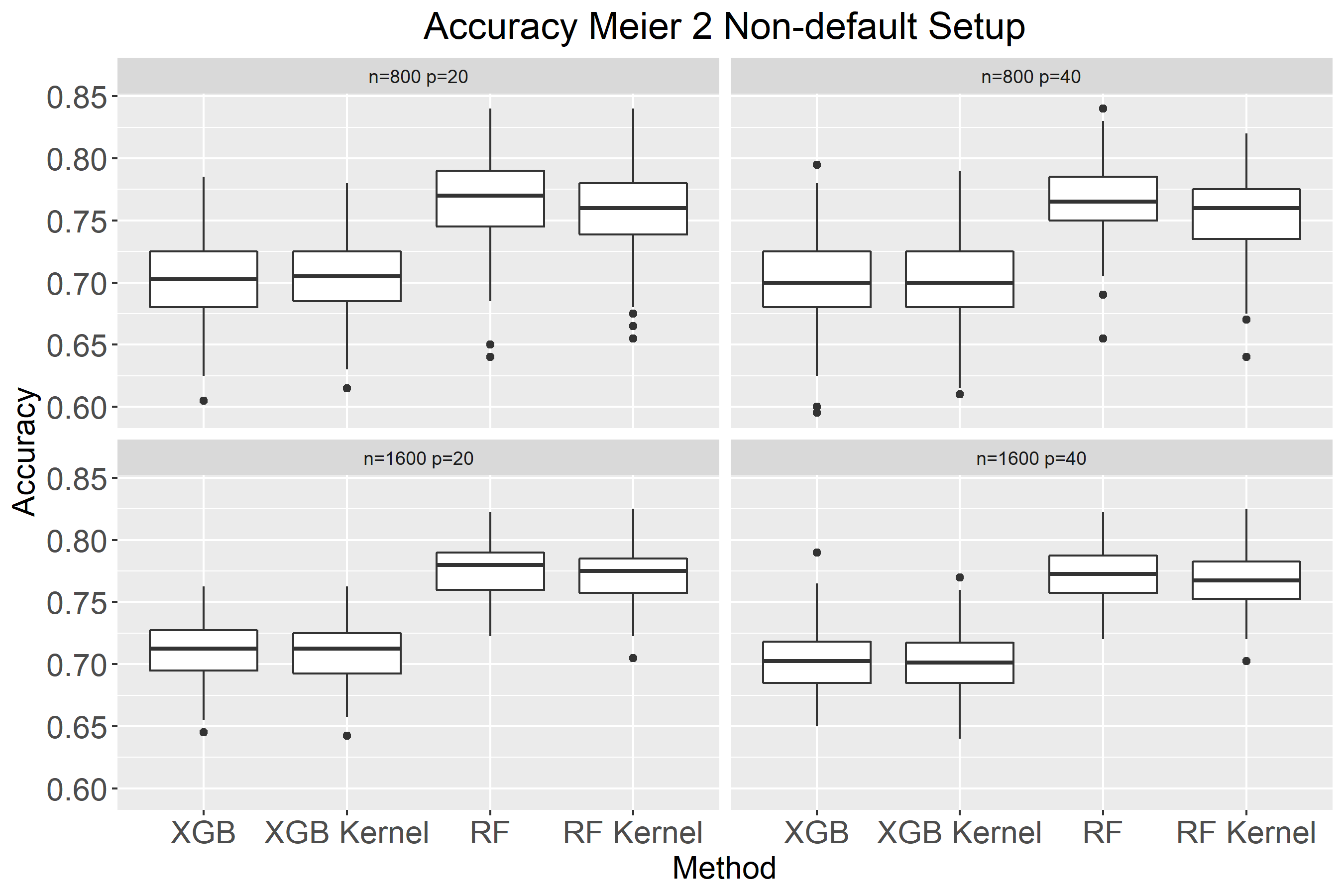} 
  \label{fig:sub-first}
    \subcaption{}
\end{subfigure}
\begin{subfigure}{.45\textwidth}
  \centering
  \includegraphics[height=0.2\textheight]{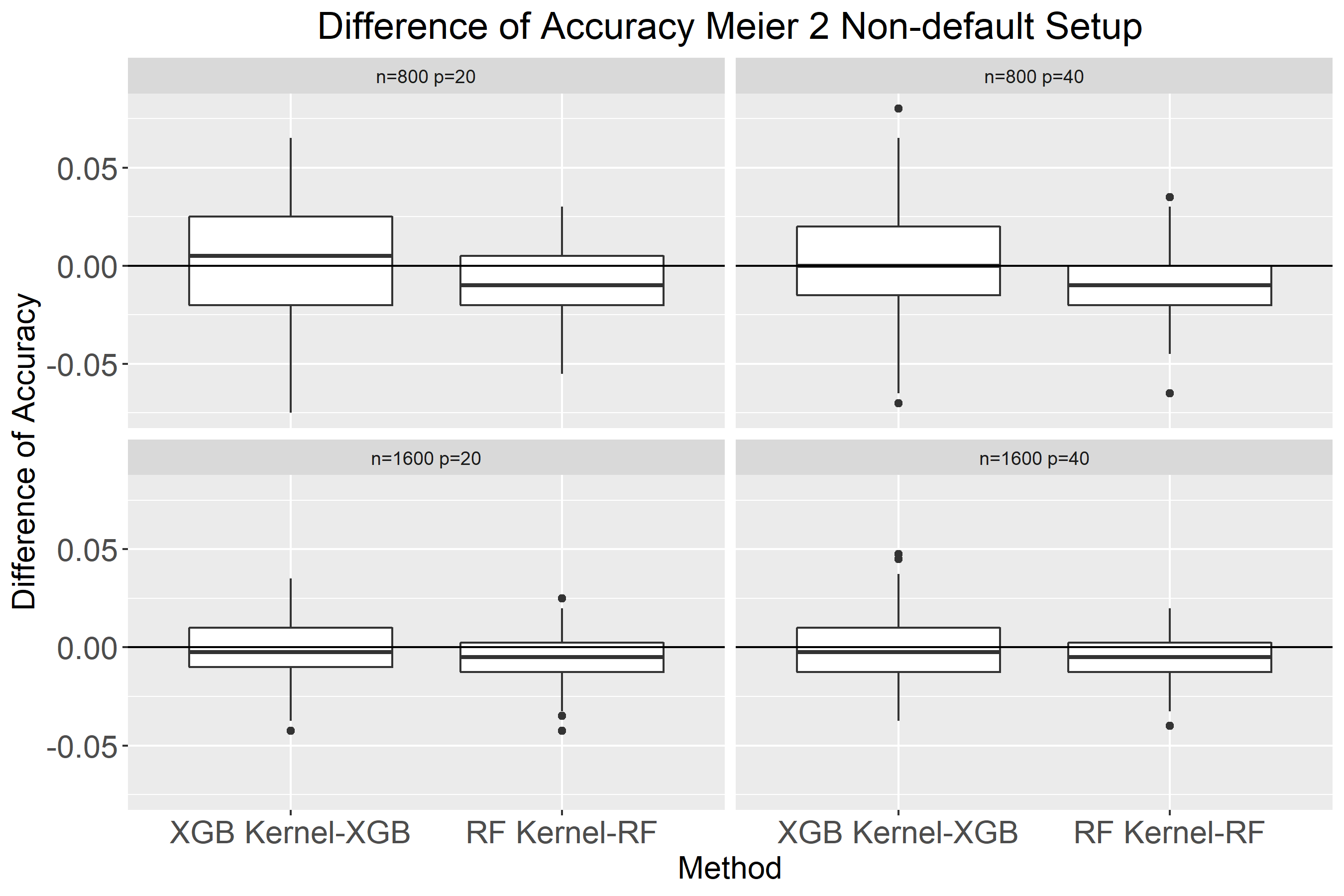} 
  \label{fig:sub-first}
    \subcaption{}
\end{subfigure}\\

\caption{Comparison of MSE and classification accuracy using RF, RF kernel, XGB, and XGB kernel using default and non-default setup in RF and XGB for data simulated from Meier 2 setting}
\label{fig:suppMeier2}
\end{figure}

\begin{figure}[ht]
\begin{subfigure}{.45\textwidth}
  \centering
  \includegraphics[height=0.20\textheight]{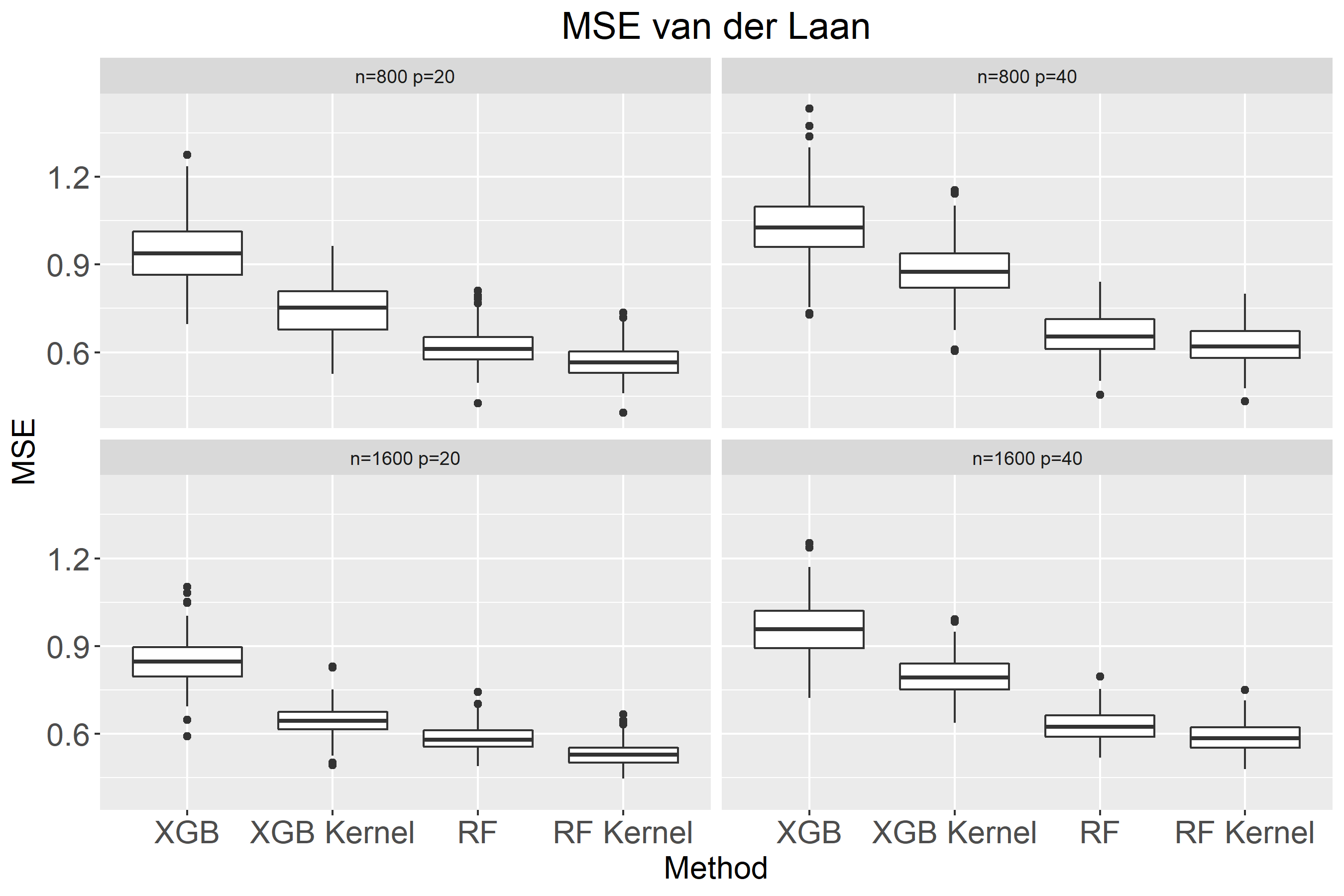}  
  \label{fig:sub-first}
    \subcaption{}
\end{subfigure}
\begin{subfigure}{.45\textwidth}
  \centering
  \includegraphics[height=0.2\textheight]{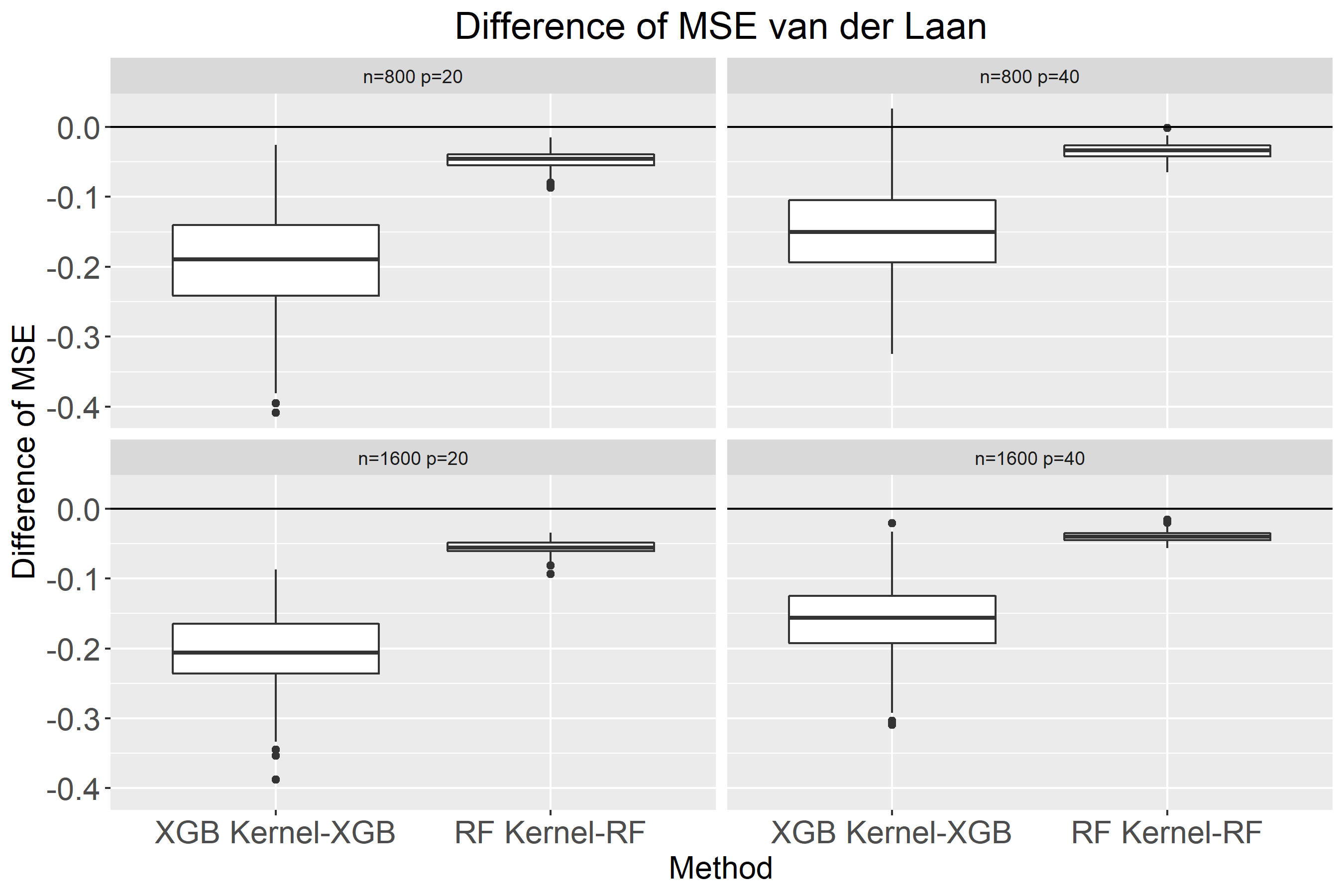}   
  \label{fig:sub-first}
    \subcaption{}
\end{subfigure}\\
\begin{subfigure}{.45\textwidth}
  \centering
  \includegraphics[height=0.2\textheight, width=0.9\textwidth]{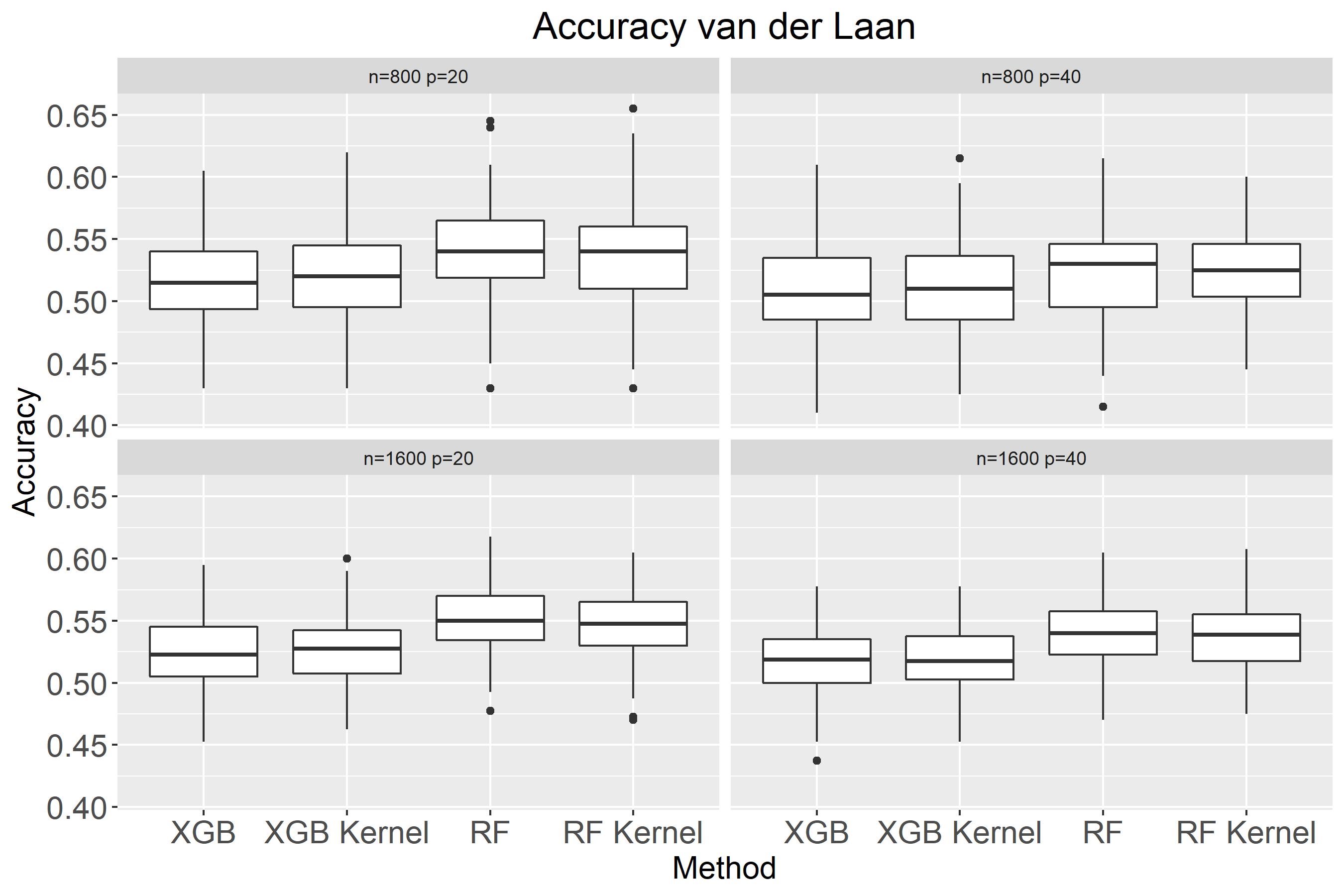}  
  \label{fig:sub-first}
    \subcaption{}
\end{subfigure}
\begin{subfigure}{.45\textwidth}
  \centering
  \includegraphics[height=0.2\textheight]{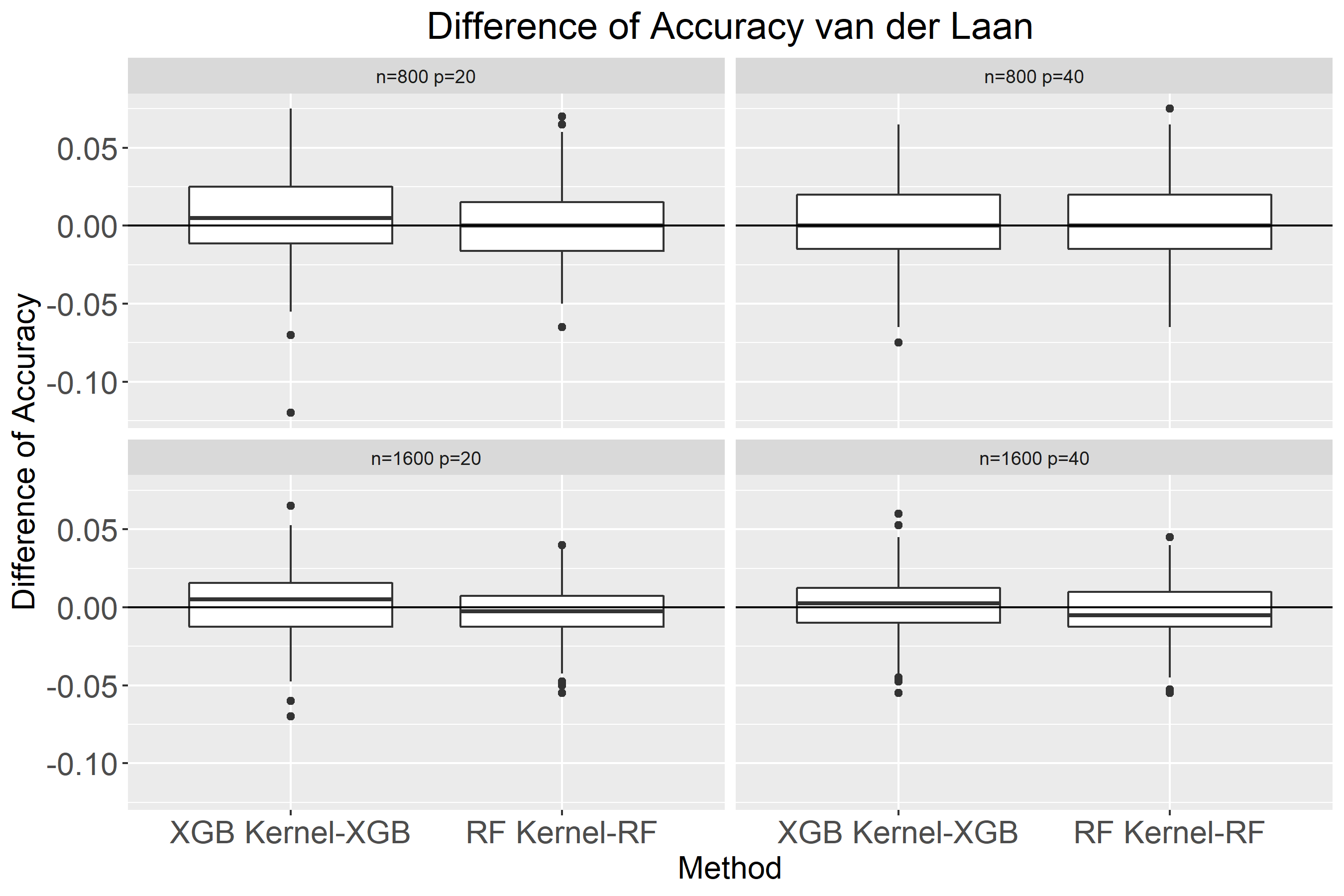} 
  \label{fig:sub-first}
    \subcaption{}
\end{subfigure}\\
\begin{subfigure}{.45\textwidth}
  \centering
  \includegraphics[height=0.20\textheight]{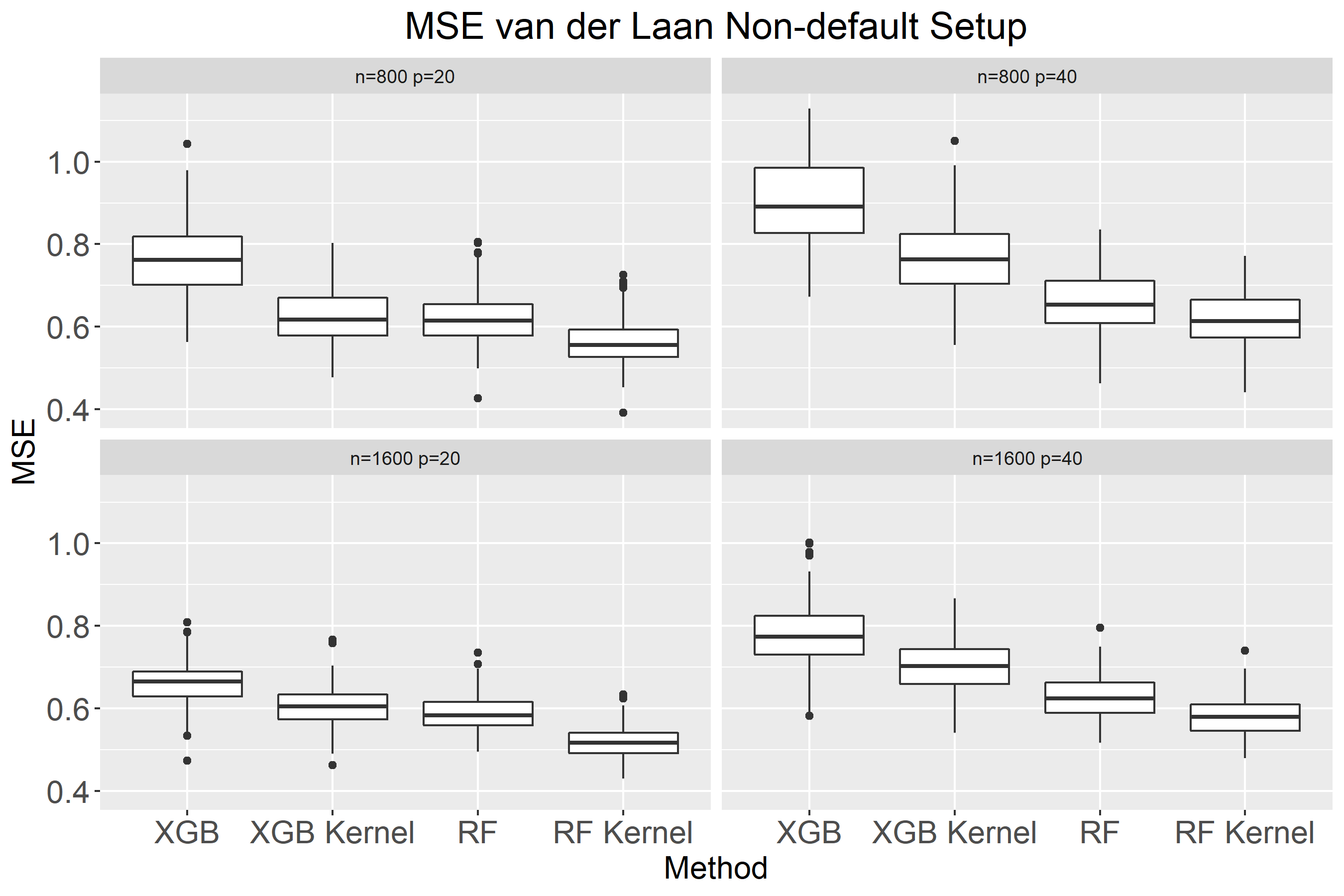}  
  \label{fig:sub-first}
    \subcaption{}
\end{subfigure}
\begin{subfigure}{.45\textwidth}
  \centering
  \includegraphics[height=0.2\textheight]{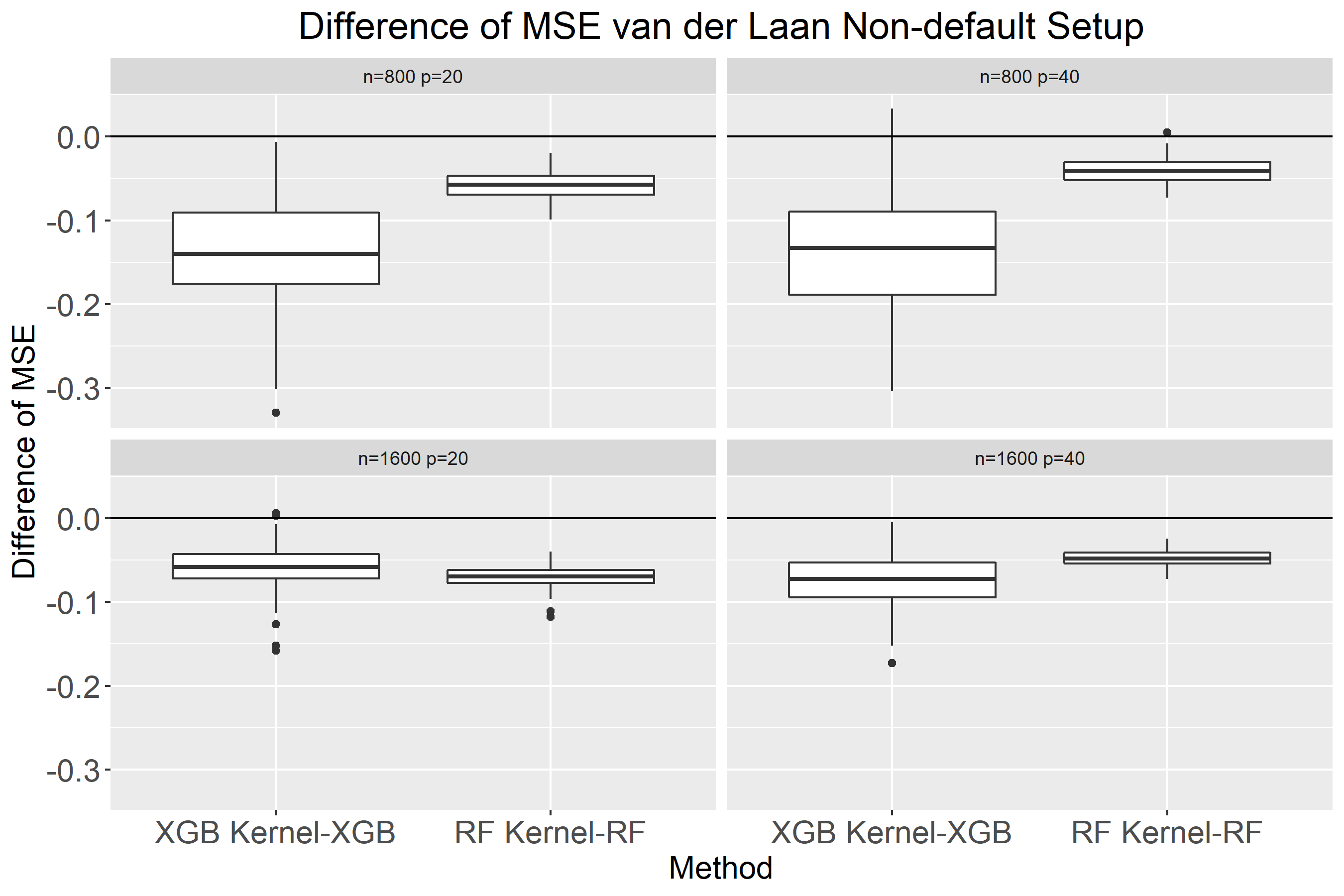}   
  \label{fig:sub-first}
    \subcaption{}
\end{subfigure}\\
\begin{subfigure}{.45\textwidth}
  \centering
  \includegraphics[height=0.2\textheight, width=0.9\textwidth]{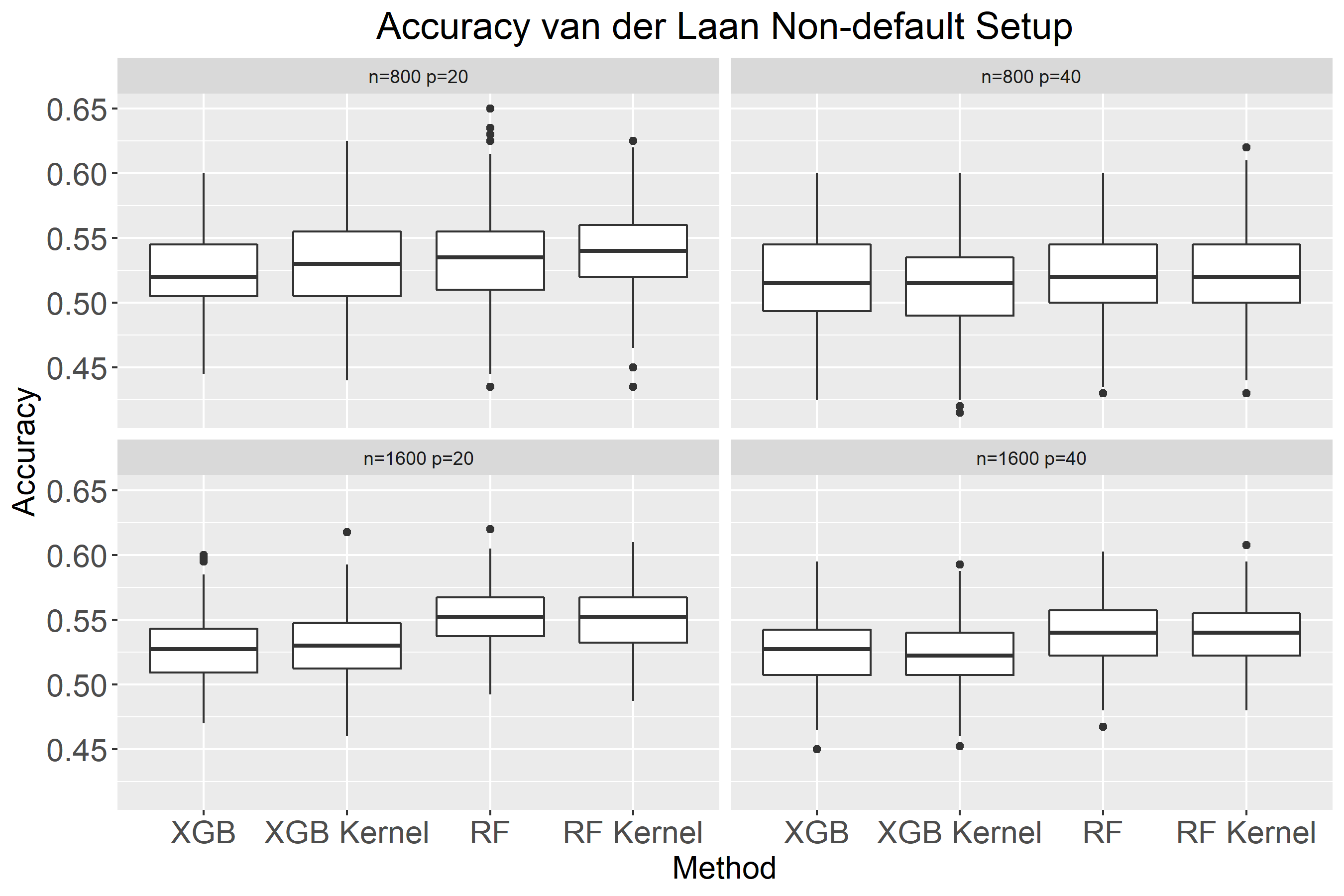} 
  \label{fig:sub-first}
    \subcaption{}
\end{subfigure}
\begin{subfigure}{.45\textwidth}
  \centering
  \includegraphics[height=0.2\textheight]{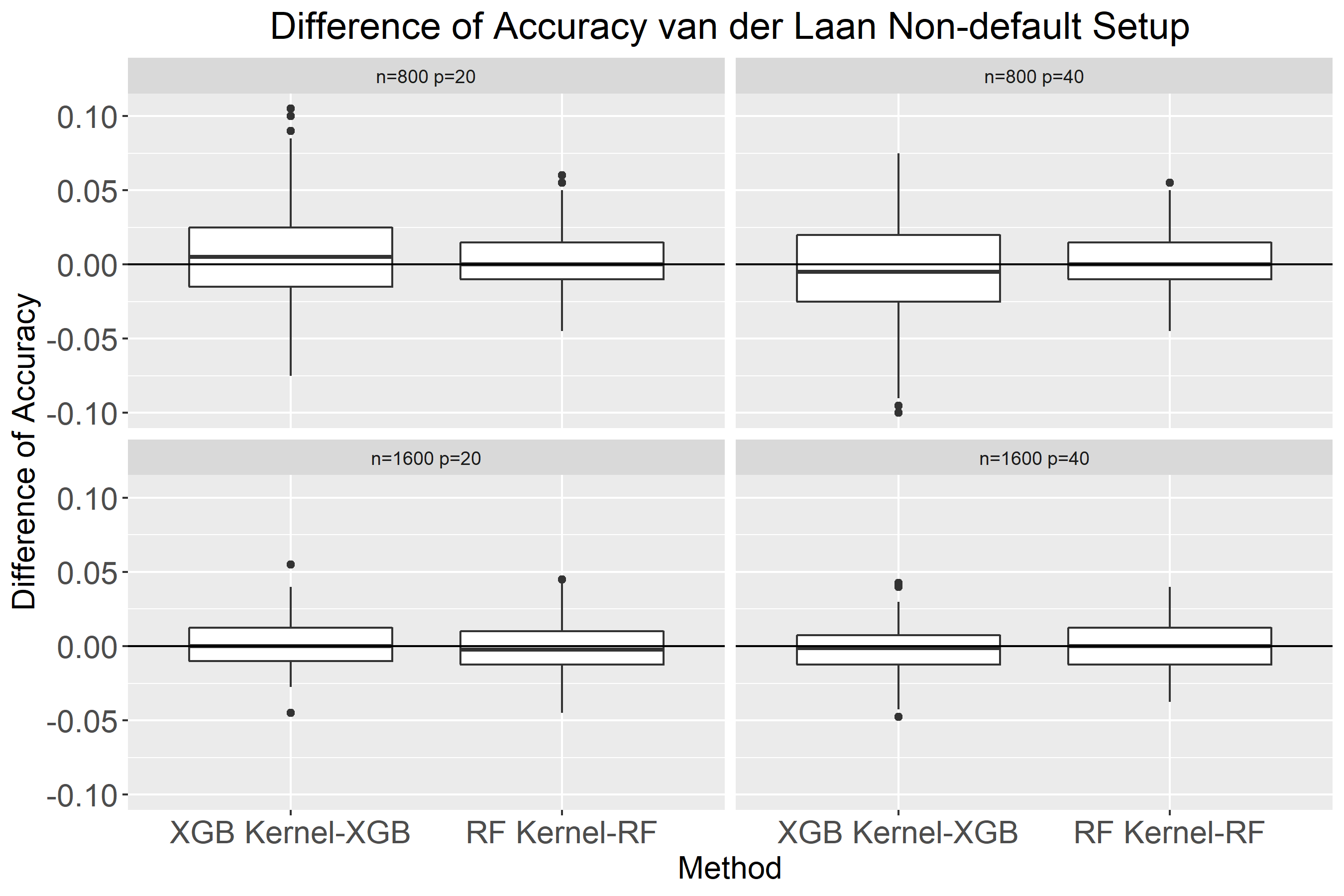} 
  \label{fig:sub-first}
    \subcaption{}
\end{subfigure}\\

\caption{Comparison of MSE and classification accuracy using RF, RF kernel, XGB, and XGB kernel using default and non-default setup in RF and XGB for data simulated from van der Laan setting}
\label{fig:suppVanderLaan}
\end{figure}

 The performance of the RF/GBT methods and RF/GBT kernels on test data for the Friedman generative model for regression and classification are shown in Fig. \ref{fig:Friedman} (a-h). The results for the default setup are shown in Fig. \ref{fig:Friedman} (a-d) and for the sensitivity analysis in Fig. \ref{fig:Friedman} (e-h). The performance metrics for the default setup for regression and classification are shown in Fig. \ref{fig:Friedman} (a) and Figure \ref{fig:Friedman} (c), respectively. To demonstrate the superiority of RF/GBT vs RF/GBT kernel, we showed also the box plots of difference in performance measures between RF kernel, GBT kernel and the corresponding RF, GBT methods in Fig. \ref{fig:Friedman} (b) and Fig. \ref{fig:Friedman} (d), respectively. A reference horizontal line with y-axis value equal to zero was drawn in each plot. The further away the box plot to the reference line (downward for MSE and upward for accuracy), the better the results of RF/GBT kernel compared to RF/GBT. The Figures showing the boxplots of the performance metrics differences for Checkerboard, Meier 1, Meier 2, and van der Laan are provided in Supporting Information (Figures \ref{fig:suppCheckerboard},\ref{fig:suppMeier1},\ref{fig:suppMeier2} and \ref{fig:suppVanderLaan}). For completeness, the overall summary of the performance results from primary analysis across all setups for continuous and binary targets are provided in Tables \ref{tab:tableContinuous} and \ref{tab:tableBinary}, respectively.

With respect to the RF/GBT kernel vs. RF/GBT comparison, the kernels generally outperformed RF/GBT for regression. Furthermore, for regression, it tended to be the case that with the same sample size (fixed $n$), the smaller the signal-to-noise ratio (the larger the value of $p$), the larger the improvement of adopting RF/GBT kernel approach after using RF/GBT. In addition, with a fixed number of covariates, the results from RF/GBT kernel was more accurate compared to the RF/GBT as the sample size decreased.
For classification, the performance was impacted by the target dichotomization and generally RF/GBT was found comparable with kernels. Specifically, for the Friedman data the RF/GBT kernels were performing slightly better than RF/GBT (Fig.\ref{fig:Friedman}(d)). For Meier 1 (Fig.\ref{fig:suppMeier1}(d)) and Meier 2 (Fig.\ref{fig:suppMeier2}(d)), the RF kernel was marginally worse the RF. The XGB kernel for Meier 1 and Meier 2 performed slightly better than XGB. 
For the checkerboard (Fig.\ref{fig:suppCheckerboard}(d)) the RF kernel and RF performed about the same and the XGB kernel slightly outperformed XGB. 
For the the van der Laan data (Fig.\ref{fig:suppVanderLaan}(d)), the RF/GBT and RF/GBT kernel classification performances were about the same.

When comparing XGB based methods vs. those based on RF for regression, the XGB based methods performed slightly worse than the RF based methods (Figs.\ref{fig:Friedman}(a), \ref{fig:suppCheckerboard}(a), \ref{fig:suppMeier1}(a), and \ref{fig:suppVanderLaan}(a)) across the different simulation scenarios.  Data sets Meier 2 yielded comparable performance (Fig. \ref{fig:suppMeier2}(a)).

The results from the sensitivity analysis for Checkerboard, Meier 1, Meier 2, and van der Laan are shown sub-figures (e-h) in Figures \ref{fig:suppCheckerboard},\ref{fig:suppMeier1},\ref{fig:suppMeier2} and \ref{fig:suppVanderLaan}. The corresponding numerical results are given in Supplementary information in Tables
\ref{tab:tableContinuousNode2} and \ref{tab:tableBinaryNode2} for regression and classification, respectively. These results were in line with those from the primary analysis. 

\noindent

\section{Real Data}\label{sec4}
\subsection{Regression and Classification (Continuous and binary outcome)}

The performance of the RF/GBT vs RF/GBT kernels was evaluated on seven benchmark data sets obtained from the UCI repository and a Kaggle California house price data set obtained from \url{https://archive.ics.uci.edu/ml/index.php},\cite{Dua2019} and \url{https://www.kaggle.com/camnugent/california-housing-prices}, respectively. The utilized data sets are summarized in Table 
\ref{tab:tableDataSets}. Using these data sets, for the regression task we predicted the continuous targets. For classification task we dichotomized the continuous targets around their median that rendered a two class classification problem.

\begin{table}

\centering
   \caption{Summary of the real life datasets}
   \begin{tabular}{lrr}
  \hline
  Dataset & n & p \\
  \hline
  California housing \cite{pace1997} & 20640 & 9 \\
  Boston housing \cite{harrison1978}& 506 & 13 \\
  Concrete Compressive Strength \cite{yeh1998}& 1030 & 9 \\
  Servo System \cite{Quinlan1993} & 167 & 5\\
  Conventional and Social Movie (CSM) \cite{ahmed2015}& 187 & 12\\
  Protein Tertiary Structure \cite{Dua2019} & 45730 & 9\\
  Auto mpg \cite{Quinlan1993}& 392 & 9\\\hline
   \end{tabular}
\label{tab:tableDataSets}   
\end{table}

For regression, for the larger data sets (California and Protein) we randomly selected 2000 samples and split them into training and test set, with 1500 and 500 samples, respectively. We repeated the analysis 200 times to evaluate the performance of RF(GBT) and RF(GBT) kernel algorithms, respectively. For the other data sets we split the data into training and test set in the ratio 3 to 1, respectively and repeated the analysis 200 times. 
Similarly for classification we randomly split the data with dichotomized target. We evaluated the performance of RF(GBT) and RF(GBT) kernel, in a binary classification setting, respectively. We repeated the analysis 200 times.

\begin{figure}[ht]
\begin{subfigure}{.45\textwidth}
  \centering
  \includegraphics[height=0.2\textheight]{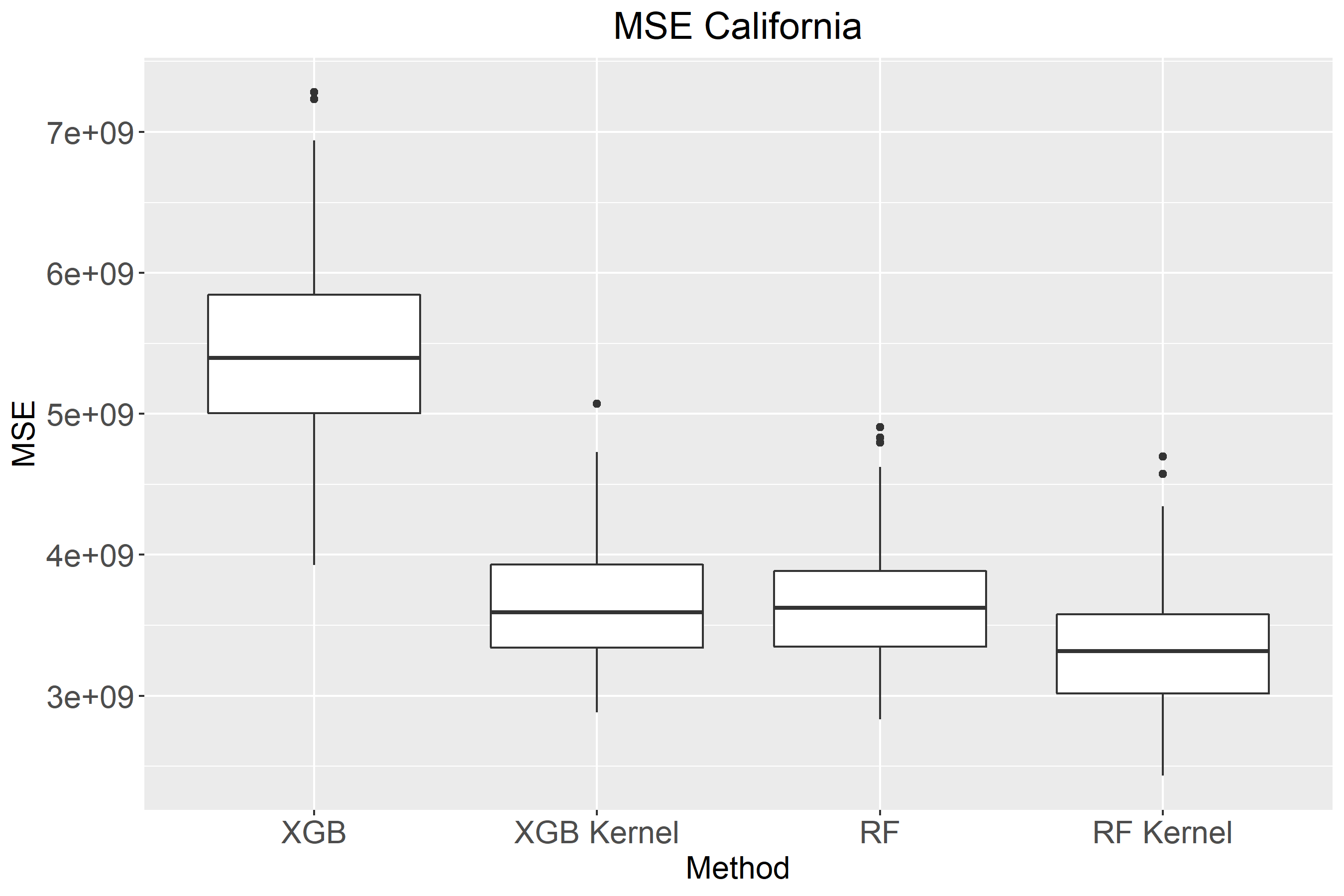}  
  \subcaption{}
  \label{fig:sub-first}
\end{subfigure}
\begin{subfigure}{.45\textwidth}
  \centering
  \includegraphics[height=0.2\textheight]{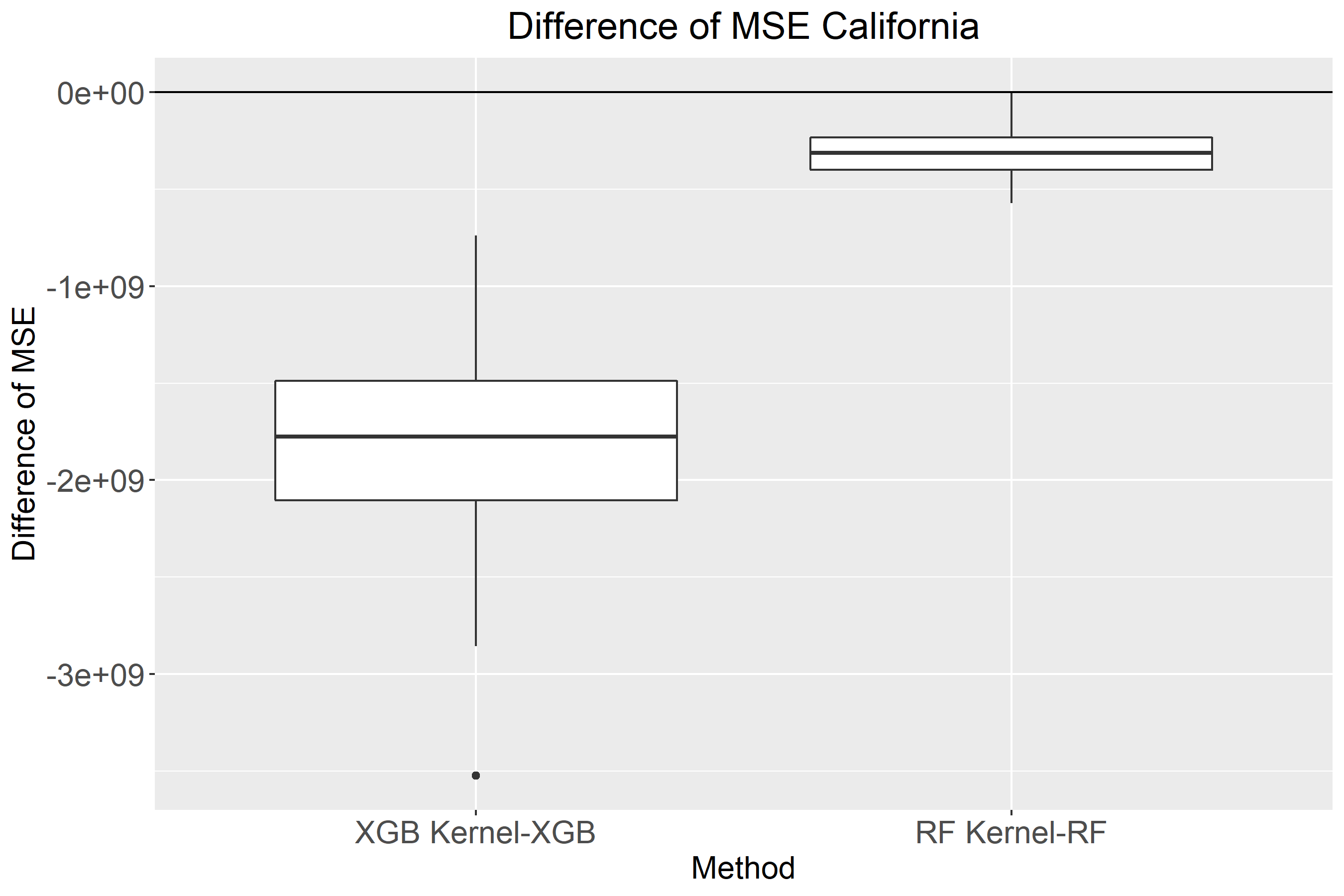}  
 \subcaption{}
  \label{fig:sub-first}
\end{subfigure}\\
\begin{subfigure}{.45\textwidth}
  \centering
  \includegraphics[height=0.2\textheight, width=0.9\textwidth]{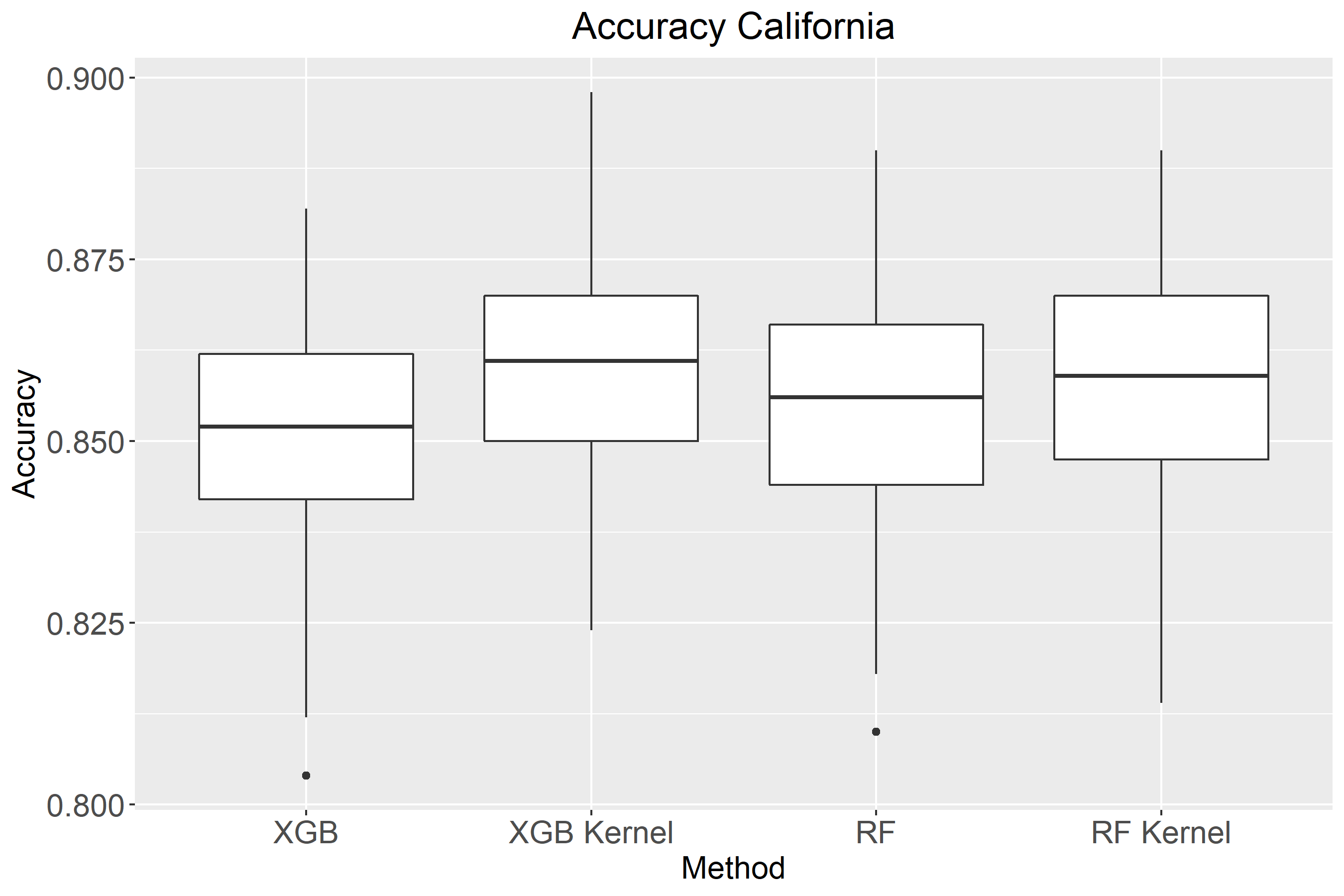}  
   \subcaption{}
  \label{fig:sub-first}
\end{subfigure}
\begin{subfigure}{.45\textwidth}
  \centering
  \includegraphics[height=0.2\textheight]{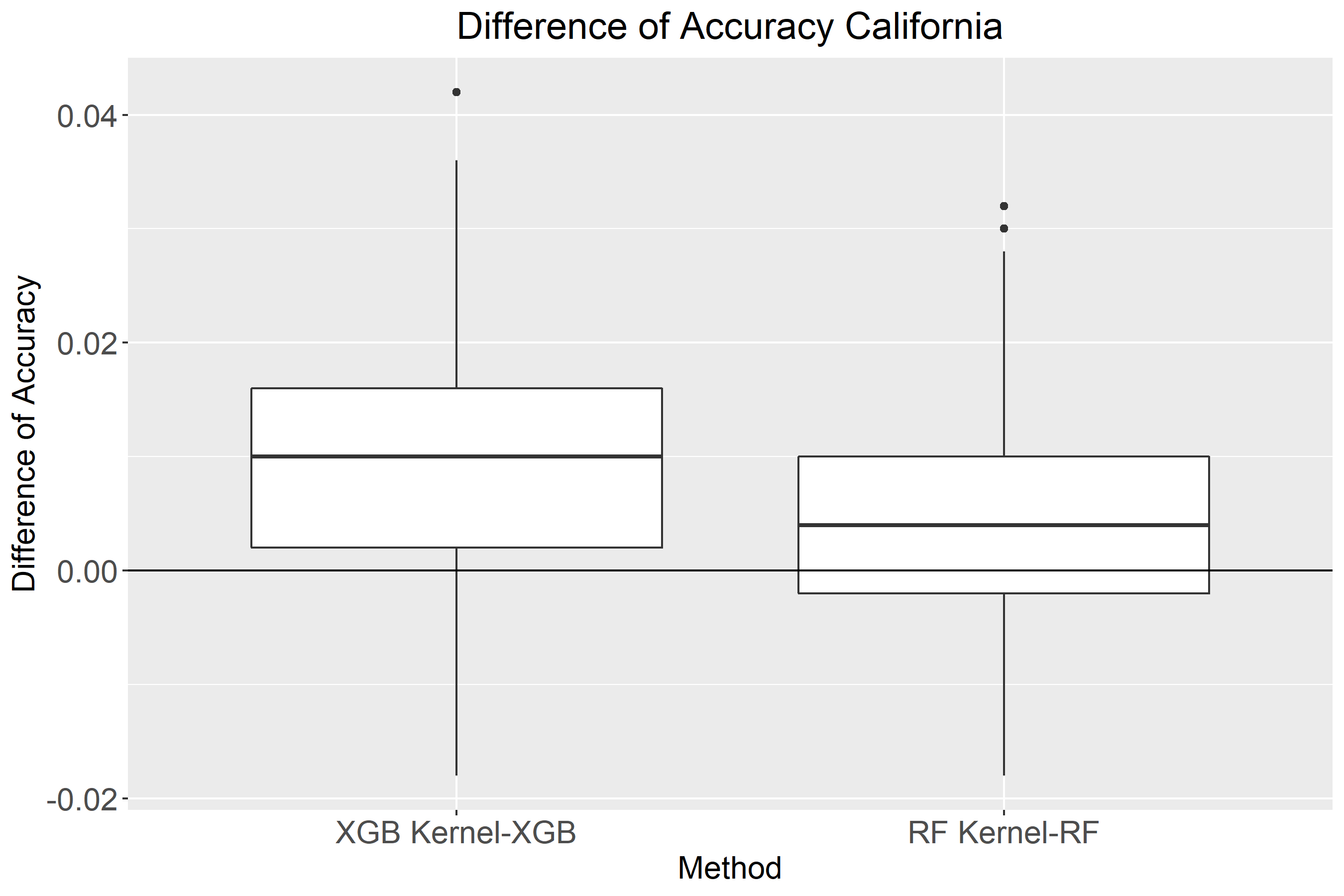}  
   \subcaption{}
  \label{fig:sub-first}
\end{subfigure}
\caption{Comparison of MSE, classification accuracy using RF, RF kernel, XGB, and XGB Kernel for the California housing data}
\label{fig:california}
\end{figure}

\begin{figure}[ht]
\begin{subfigure}{.45\textwidth}
  \centering
  \includegraphics[height=0.2\textheight]{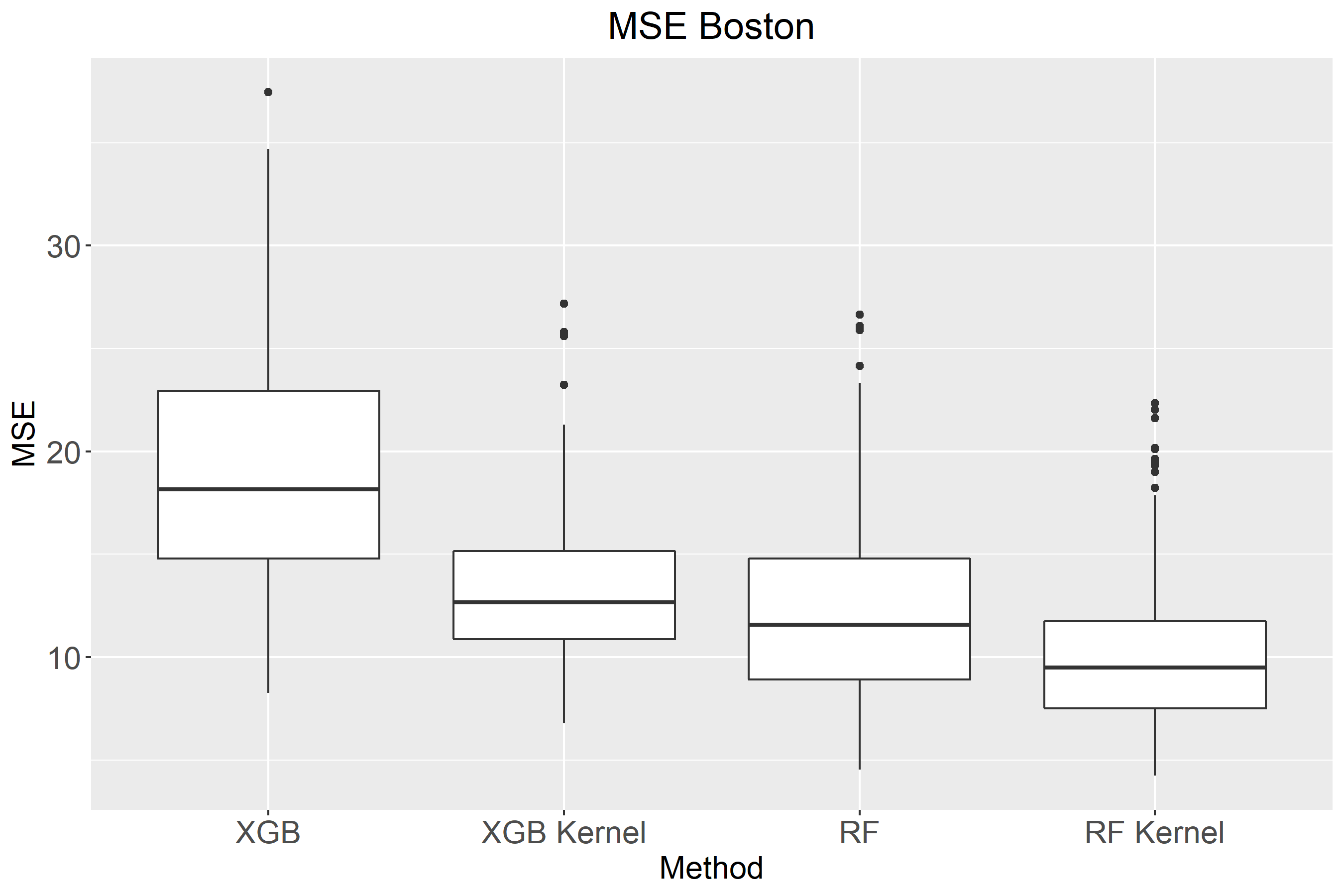}  
    \subcaption{}
  \label{fig:sub-first}
\end{subfigure}
\begin{subfigure}{.45\textwidth}
  \centering
  \includegraphics[height=0.2\textheight]{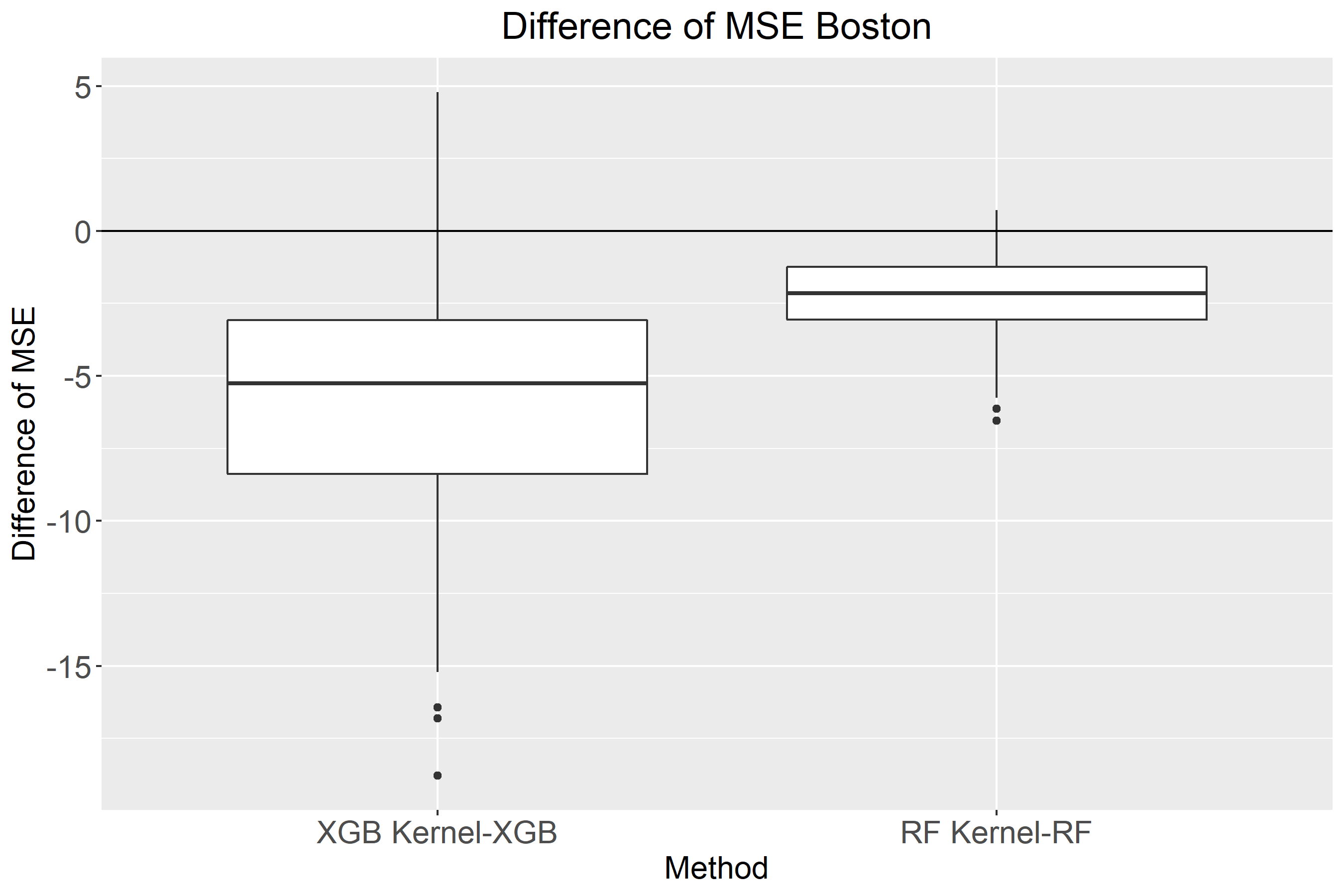}  
  \label{fig:sub-first}
     \subcaption{}
\end{subfigure}\\
\begin{subfigure}{.45\textwidth}
  \centering
  \includegraphics[height=0.2\textheight, width=0.9\textwidth]{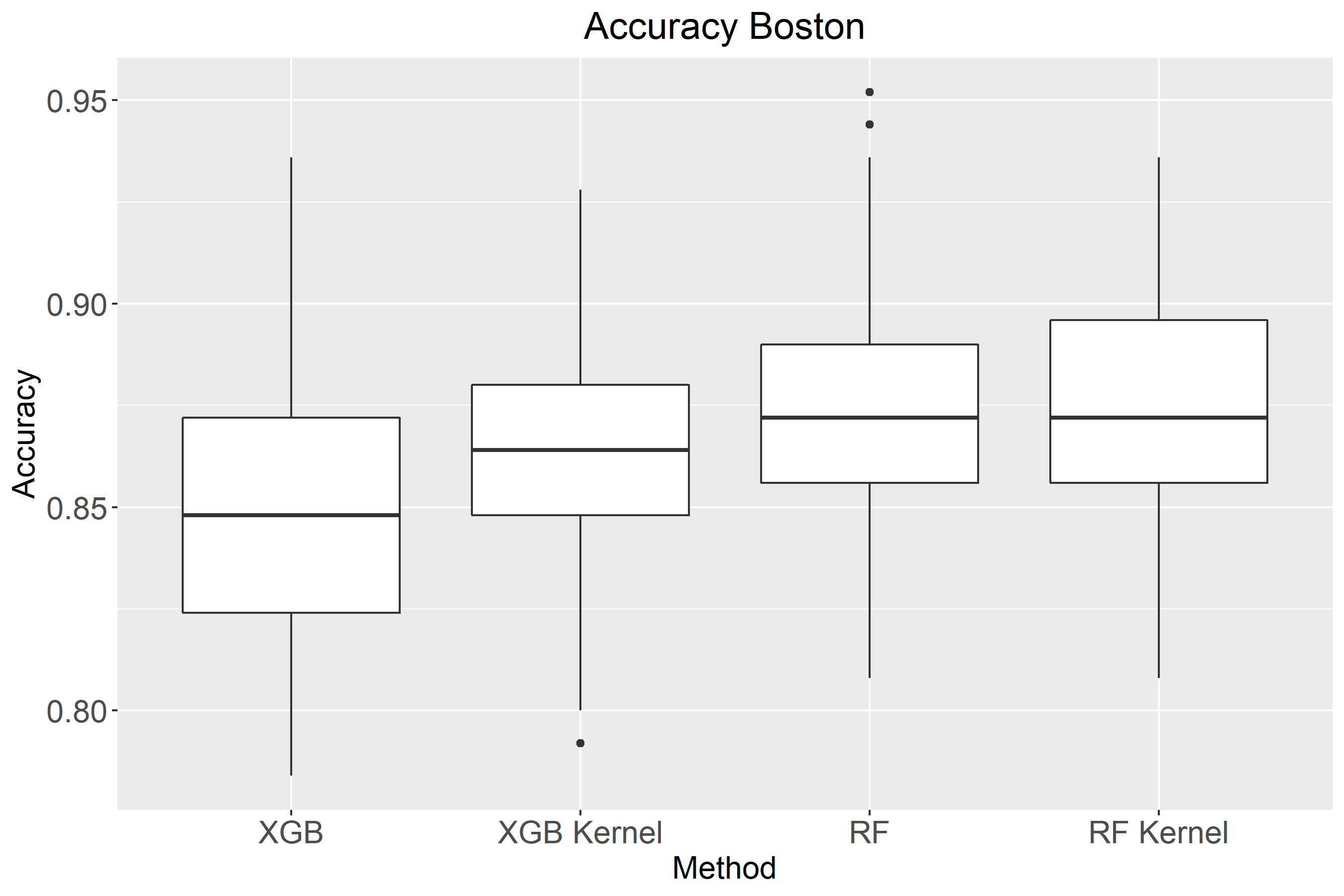}  
    \subcaption{}
  \label{fig:sub-first}
\end{subfigure}
\begin{subfigure}{.45\textwidth}
  \centering
  \includegraphics[height=0.2\textheight]{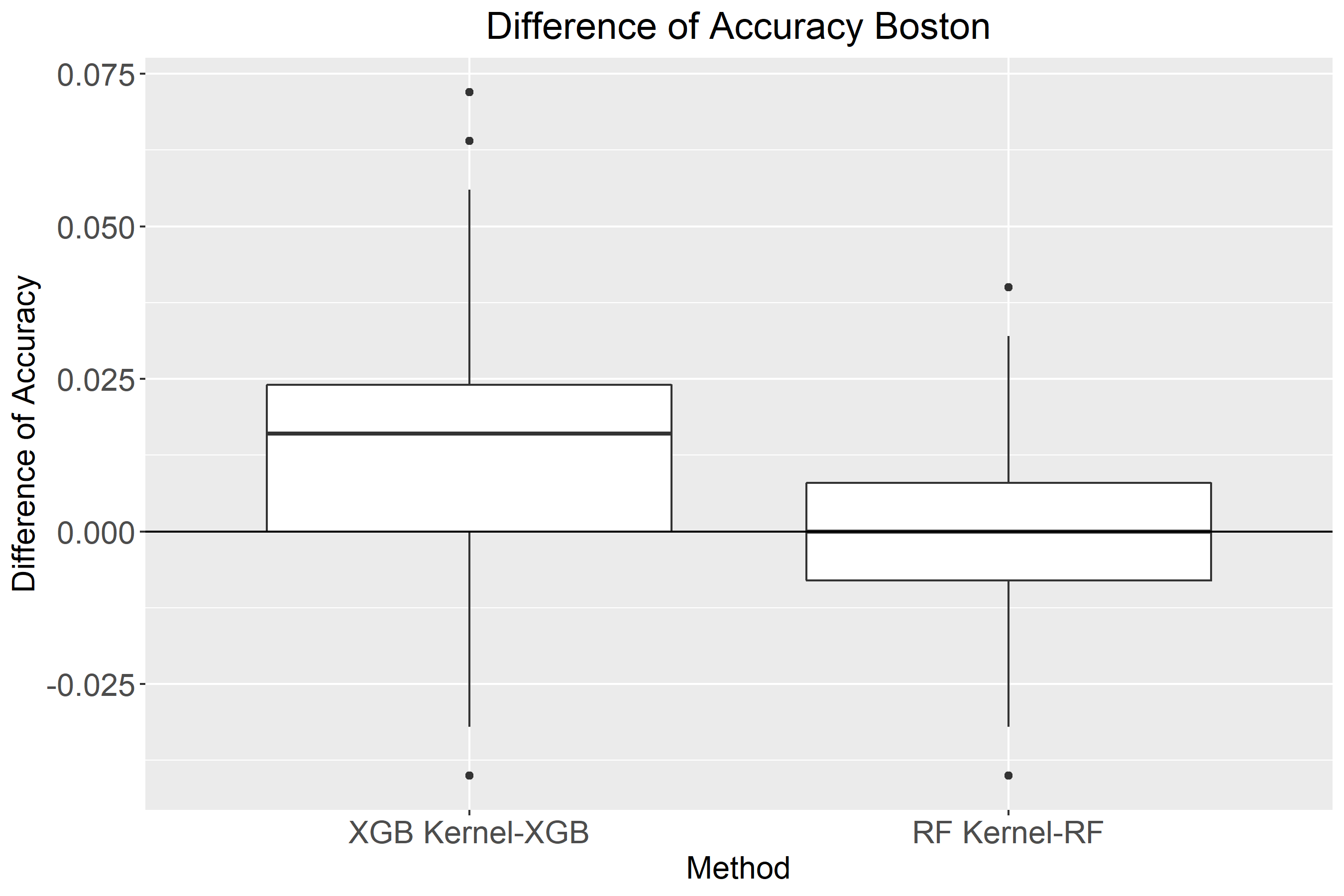}  
     \subcaption{}
  \label{fig:sub-first}
\end{subfigure}
\caption{Comparison of MSE, classification accuracy using RF, RF kernel, XGB, and XGB Kernel for the Boston housing data}
\label{fig:boston}
\end{figure}

\begin{figure}[ht]
\begin{subfigure}{.45\textwidth}
  \centering
  \includegraphics[height=0.2\textheight]{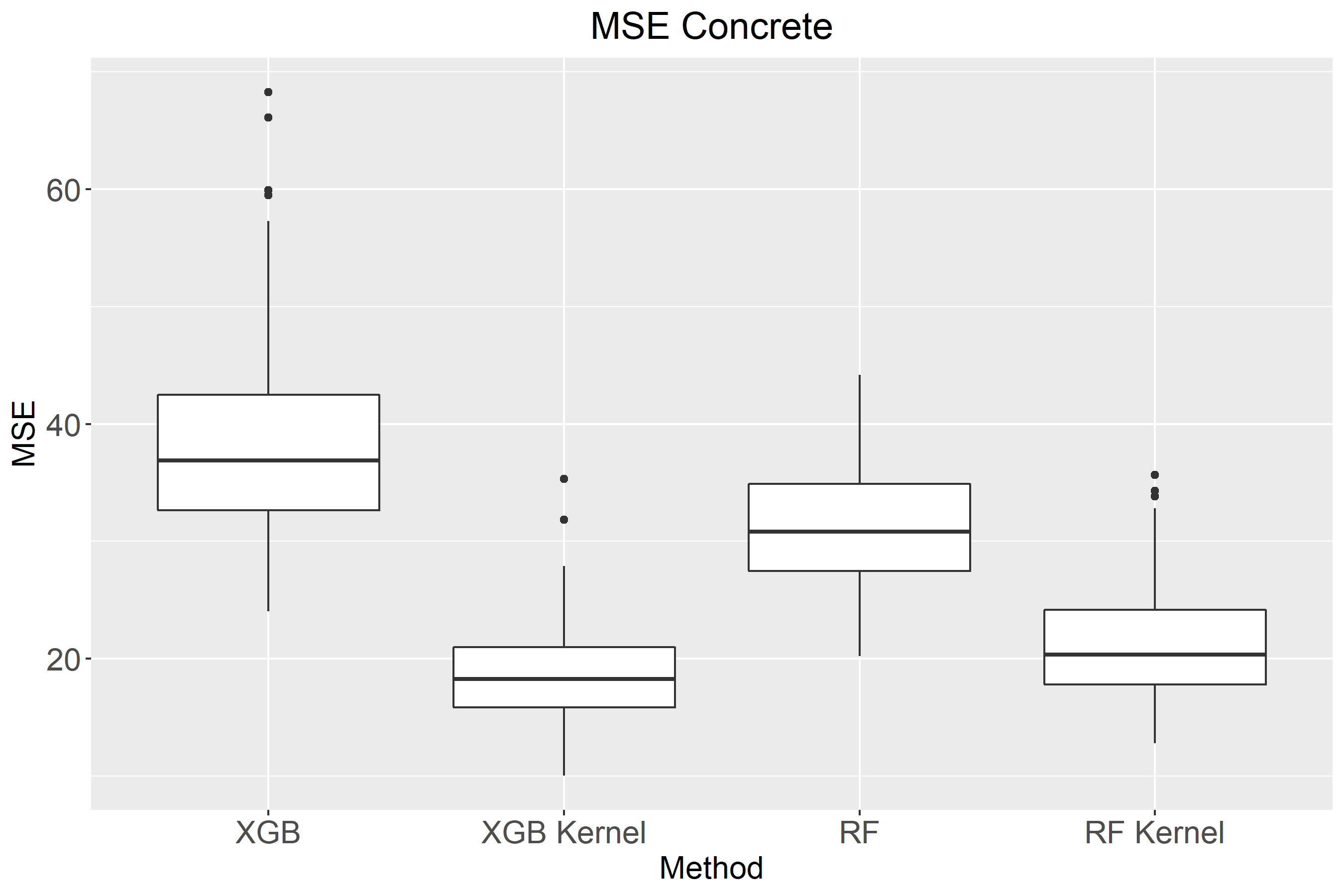}  
     \subcaption{}
  \label{fig:sub-first}
\end{subfigure}
\begin{subfigure}{.45\textwidth}
  \centering
  \includegraphics[height=0.2\textheight]{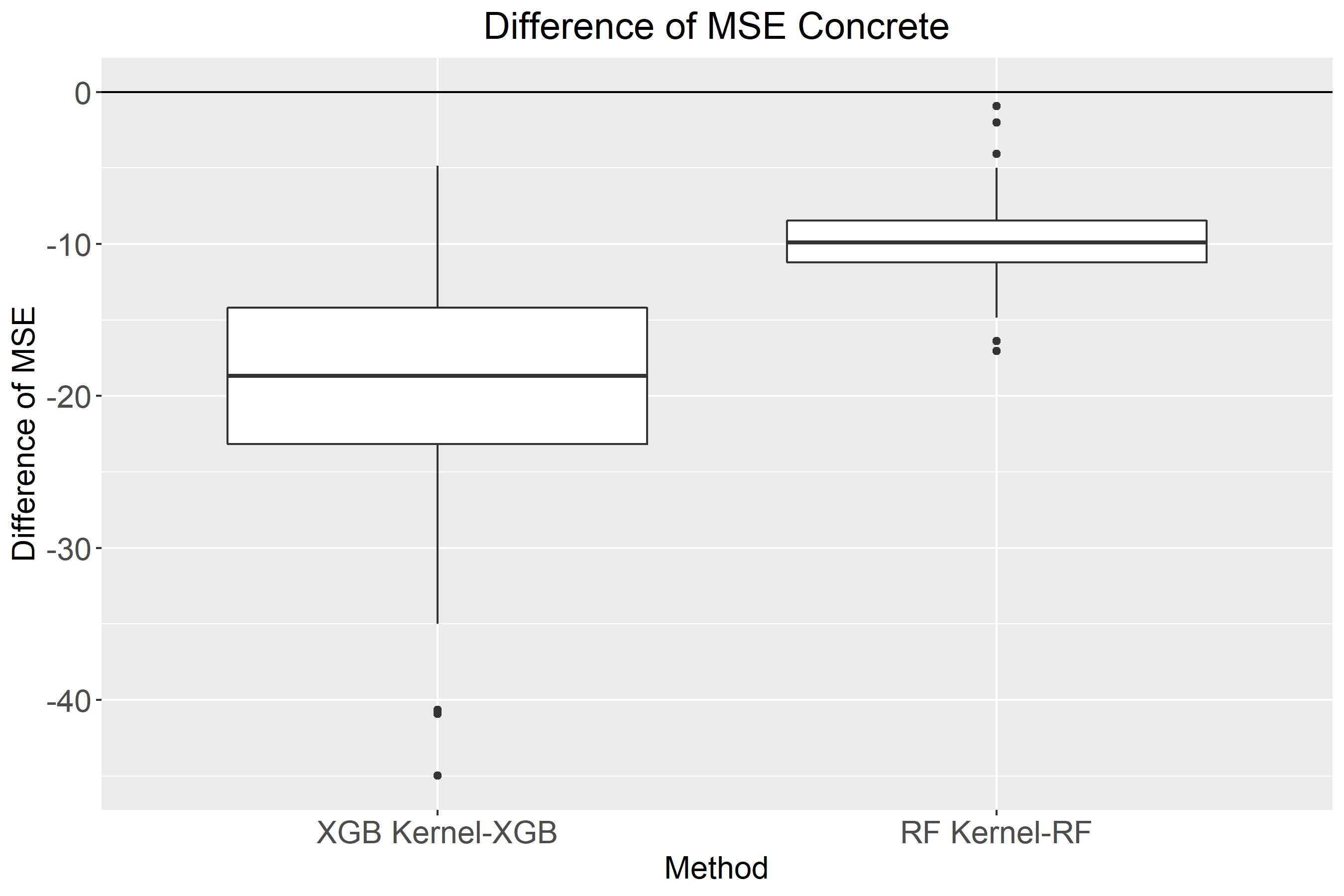}  
     \subcaption{}
  \label{fig:sub-first}
\end{subfigure}\\
\begin{subfigure}{.45\textwidth}
  \centering
  \includegraphics[height=0.2\textheight, width=0.9\textwidth]{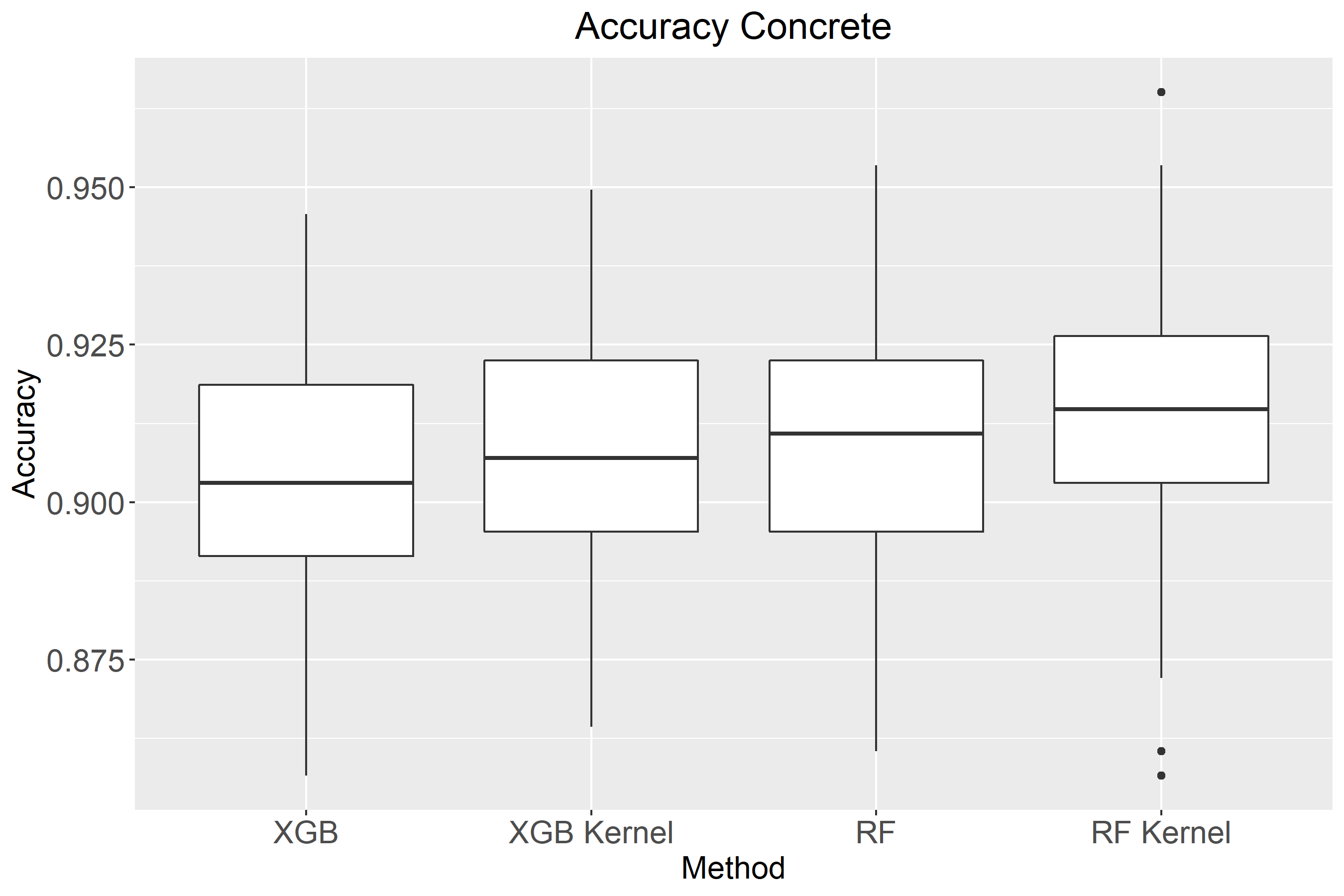}  
     \subcaption{}
  \label{fig:sub-first}
\end{subfigure}
\begin{subfigure}{.45\textwidth}
  \centering
  \includegraphics[height=0.2\textheight]{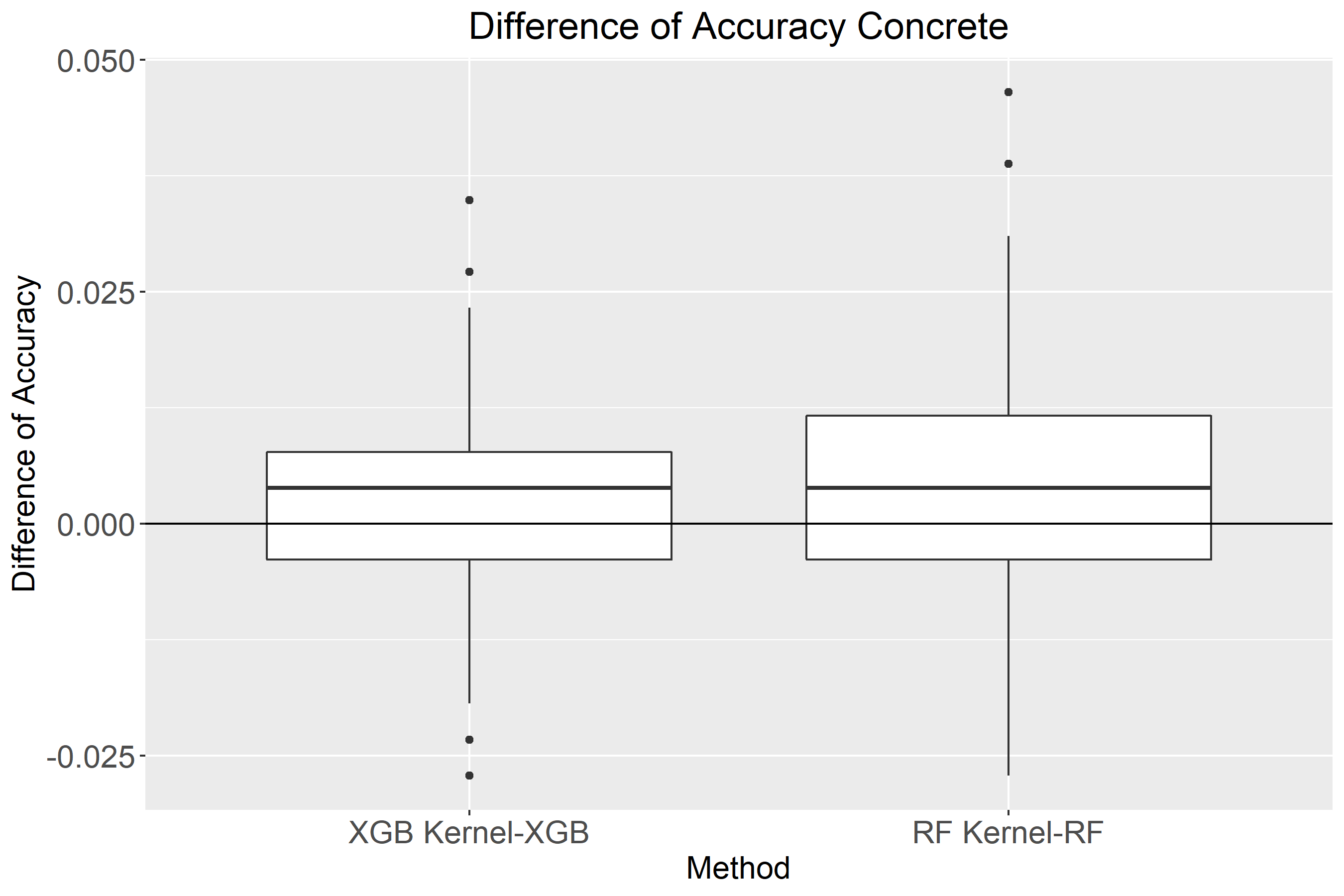}  
     \subcaption{}
  \label{fig:sub-first}
\end{subfigure}
\caption{Comparison of MSE, classification accuracy using RF, RF kernel, XGB, and XGB Kernel for the Concrete Compressive Strength data}
\label{fig:concrete}
\end{figure}

\begin{figure}[ht]
\begin{subfigure}{.45\textwidth}
  \centering
  \includegraphics[height=0.2\textheight]{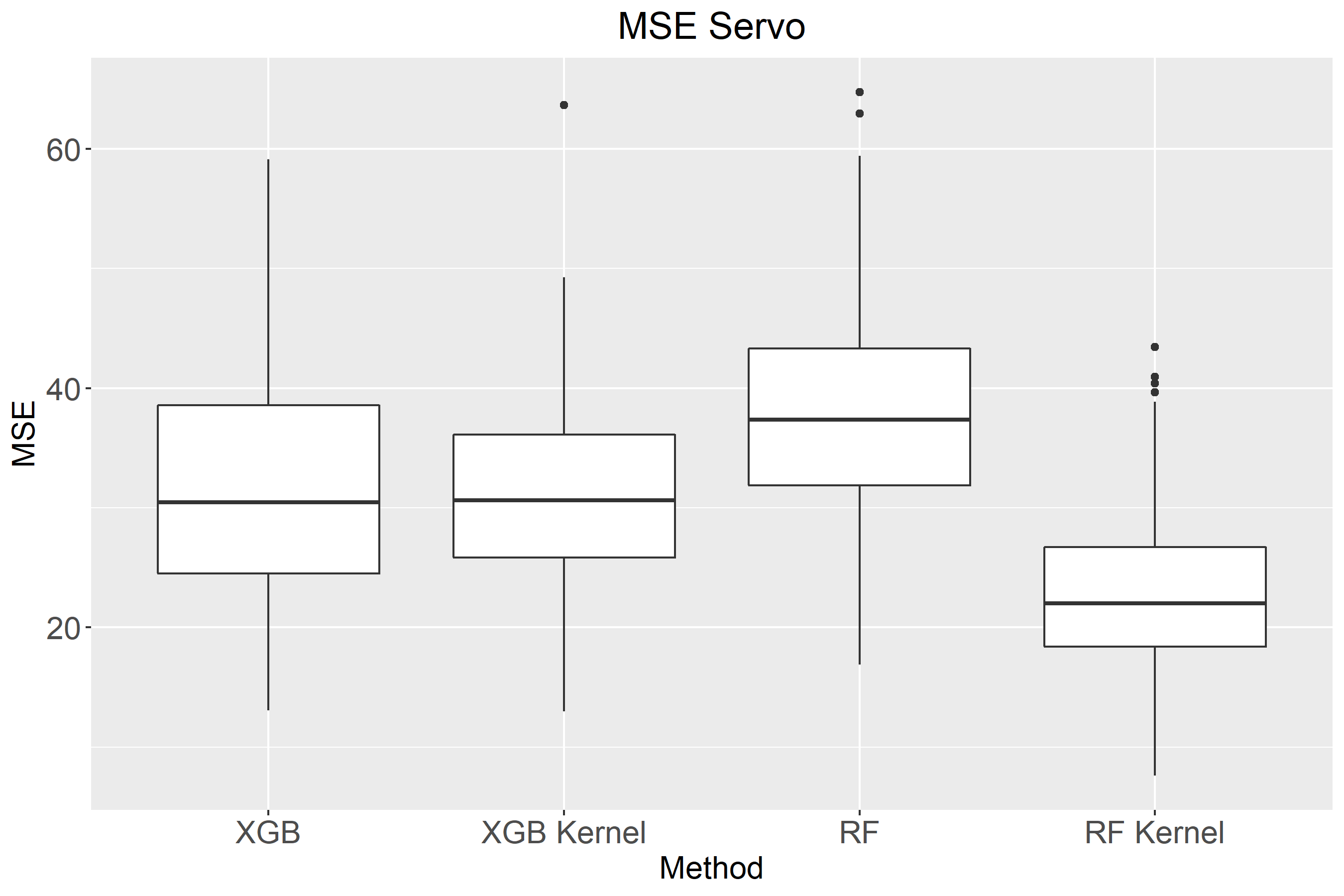}  
     \subcaption{}
  \label{fig:sub-first}
\end{subfigure}
\begin{subfigure}{.45\textwidth}
  \centering
  \includegraphics[height=0.2\textheight]{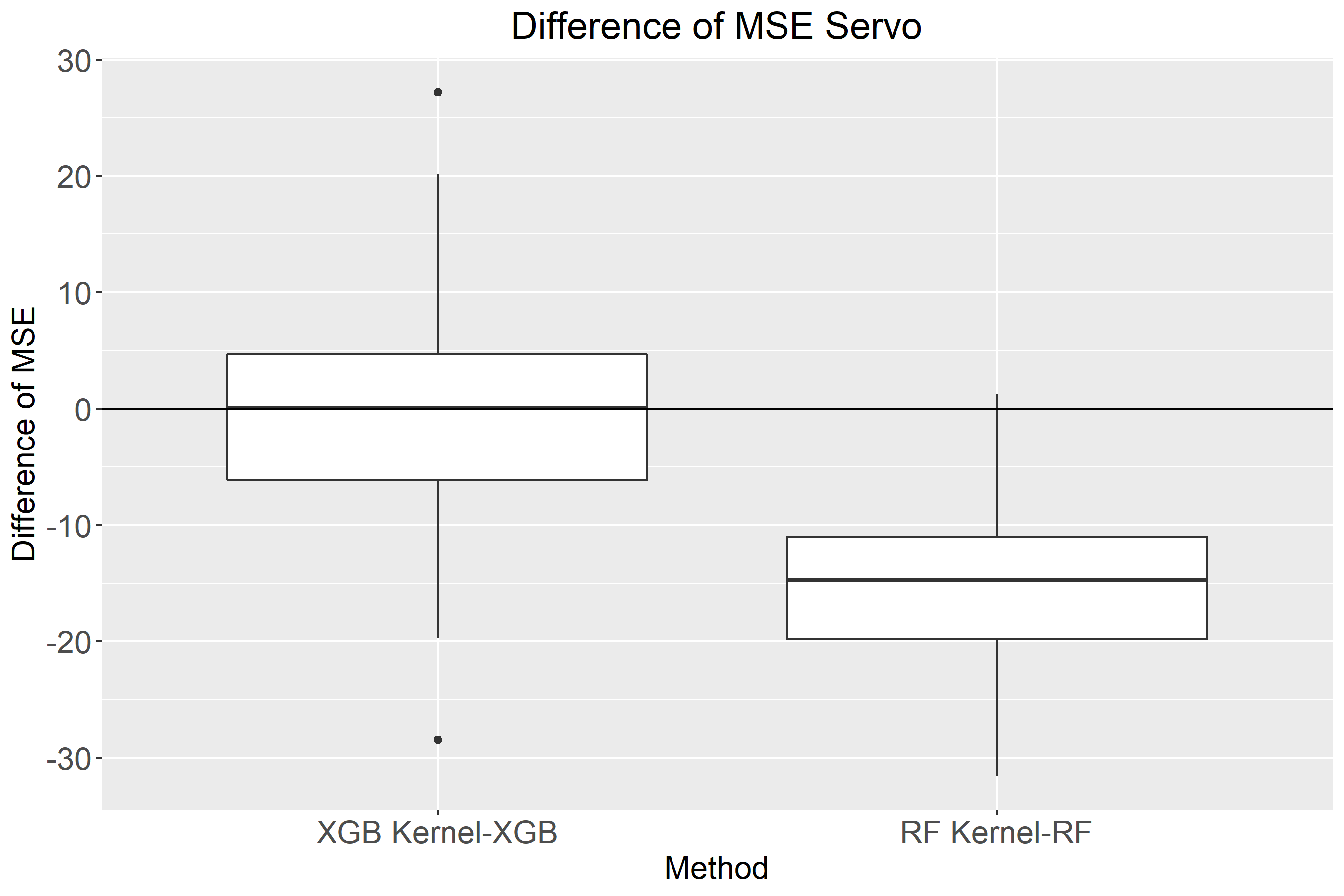}  
     \subcaption{}
  \label{fig:sub-first}
\end{subfigure}\\
\begin{subfigure}{.45\textwidth}
  \centering
  \includegraphics[height=0.2\textheight, width=0.9\textwidth]{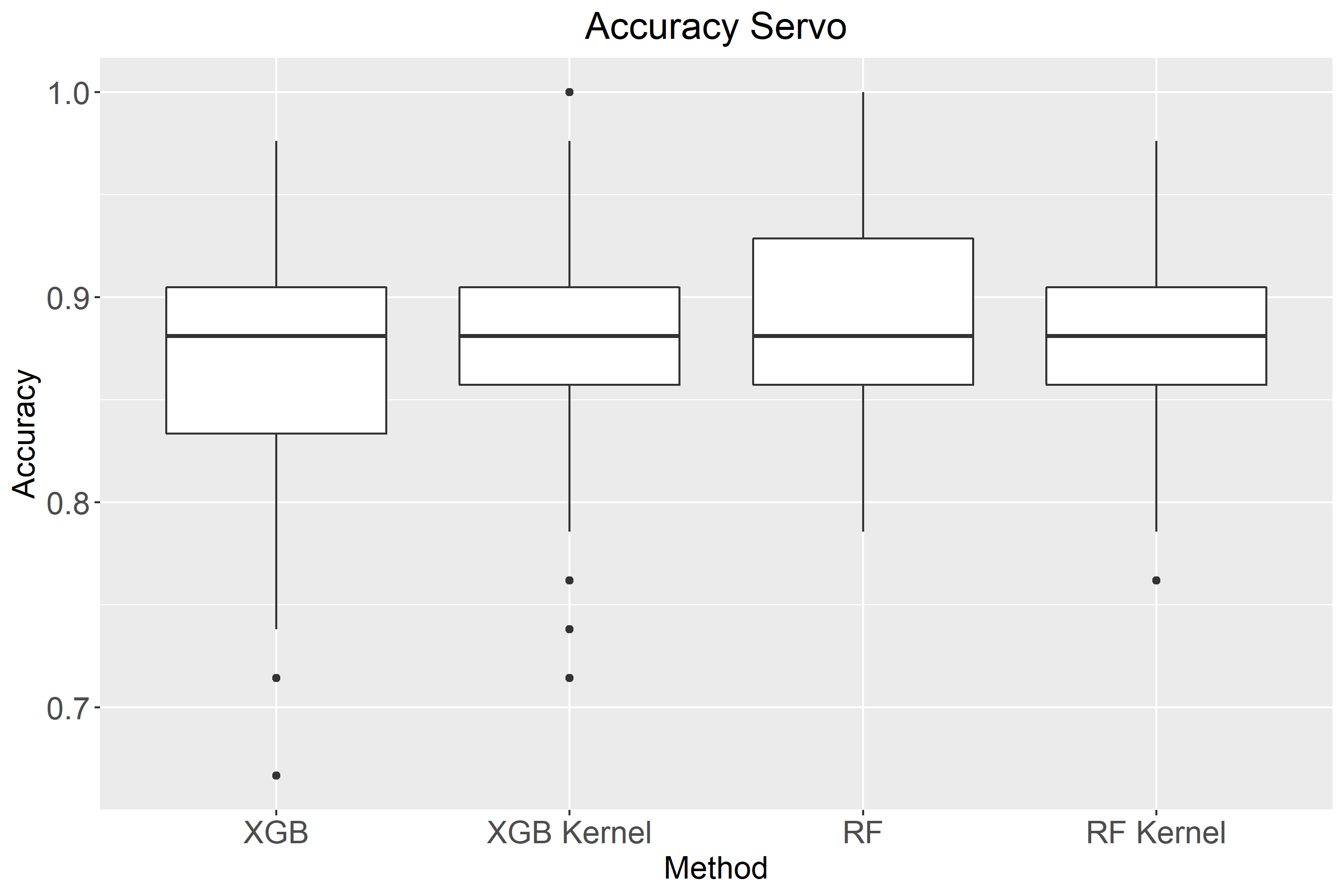}  
     \subcaption{}
  \label{fig:sub-first}
\end{subfigure}
\begin{subfigure}{.45\textwidth}
  \centering
  \includegraphics[height=0.2\textheight]{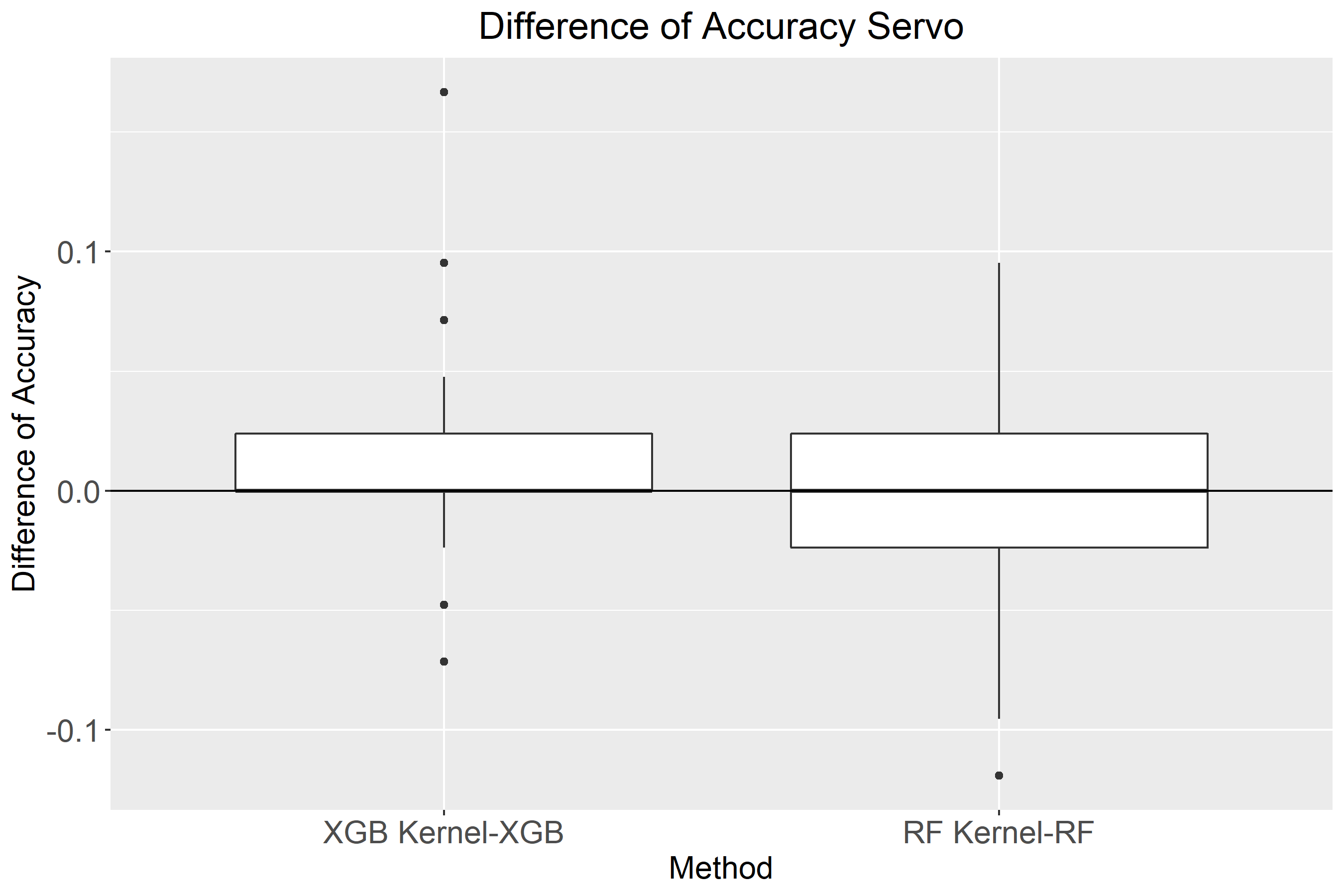}  
     \subcaption{}
  \label{fig:sub-first}
\end{subfigure}
\caption{Comparison of MSE, classification accuracy using RF, RF kernel, XGB, and XGB Kernel for the Servo System data}
\label{fig:servo}
\end{figure}

\begin{figure}[ht]
\begin{subfigure}{.45\textwidth}
  \centering
  \includegraphics[height=0.2\textheight]{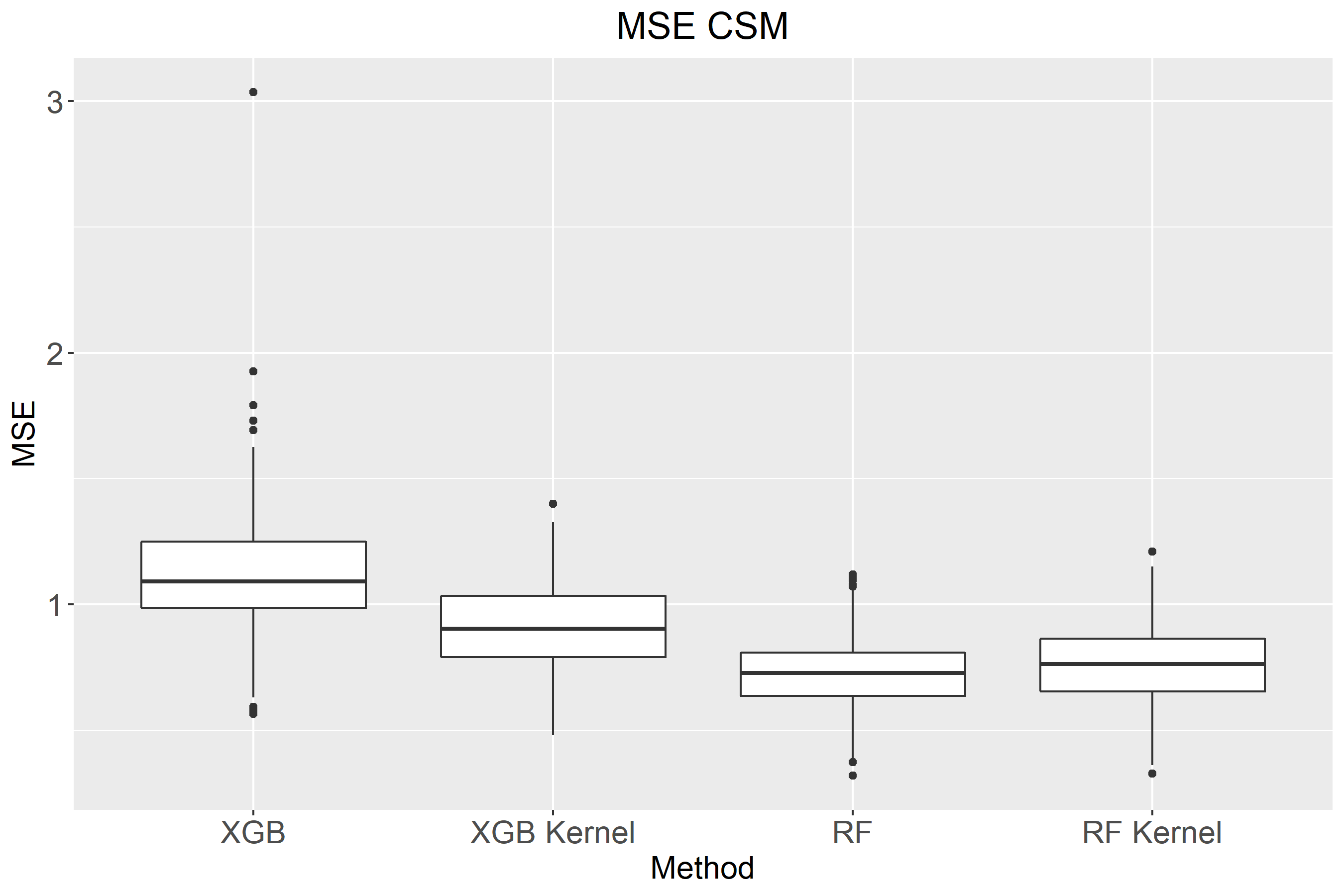}  
     \subcaption{}
  \label{fig:sub-first}
\end{subfigure}
\begin{subfigure}{.45\textwidth}
  \centering
  \includegraphics[height=0.2\textheight]{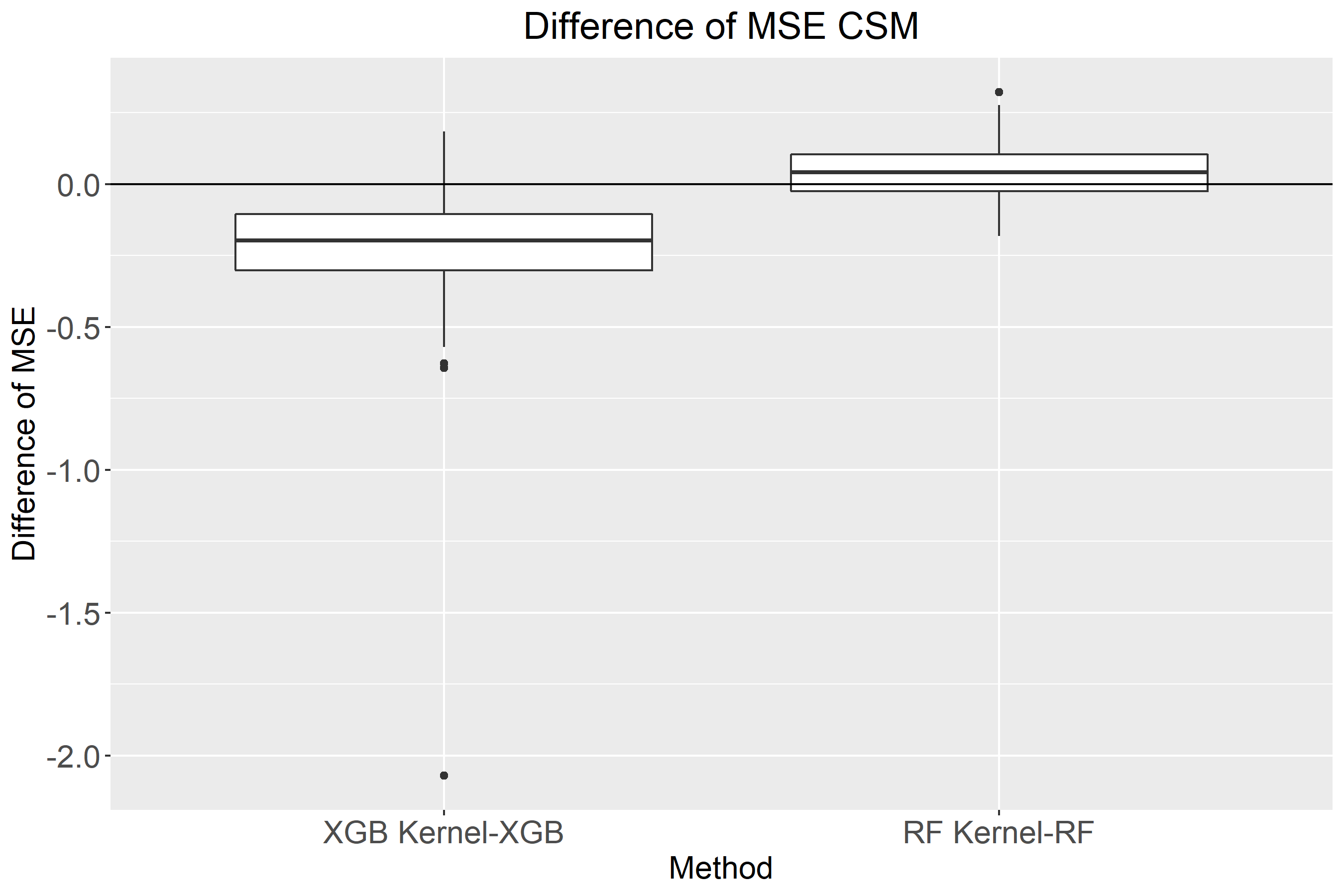}  
     \subcaption{}
  \label{fig:sub-first}
\end{subfigure}\\
\begin{subfigure}{.45\textwidth}
  \centering
  \includegraphics[height=0.2\textheight, width=0.9\textwidth]{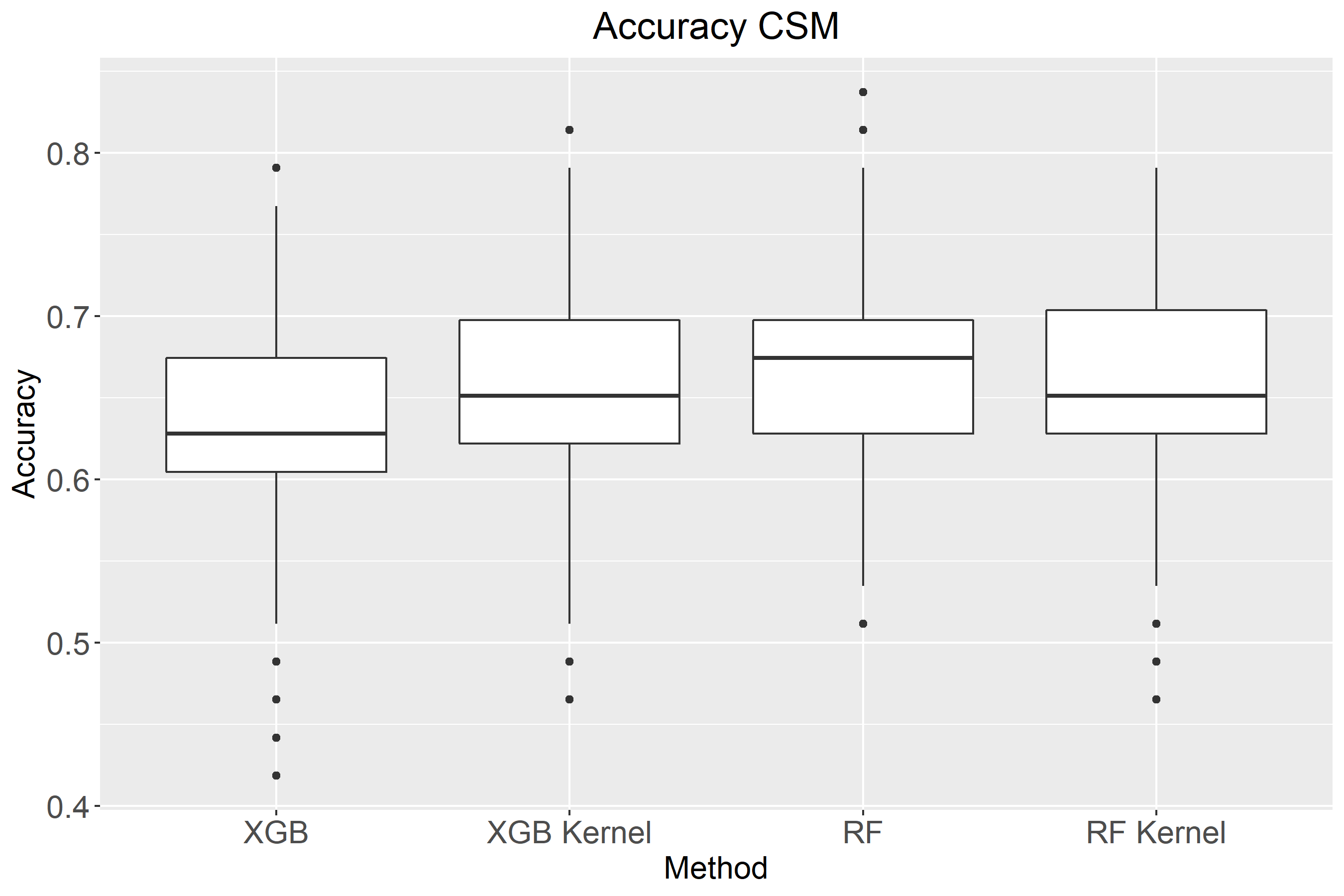}  
     \subcaption{}
  \label{fig:sub-first}
\end{subfigure}
\begin{subfigure}{.45\textwidth}
  \centering
  \includegraphics[height=0.2\textheight]{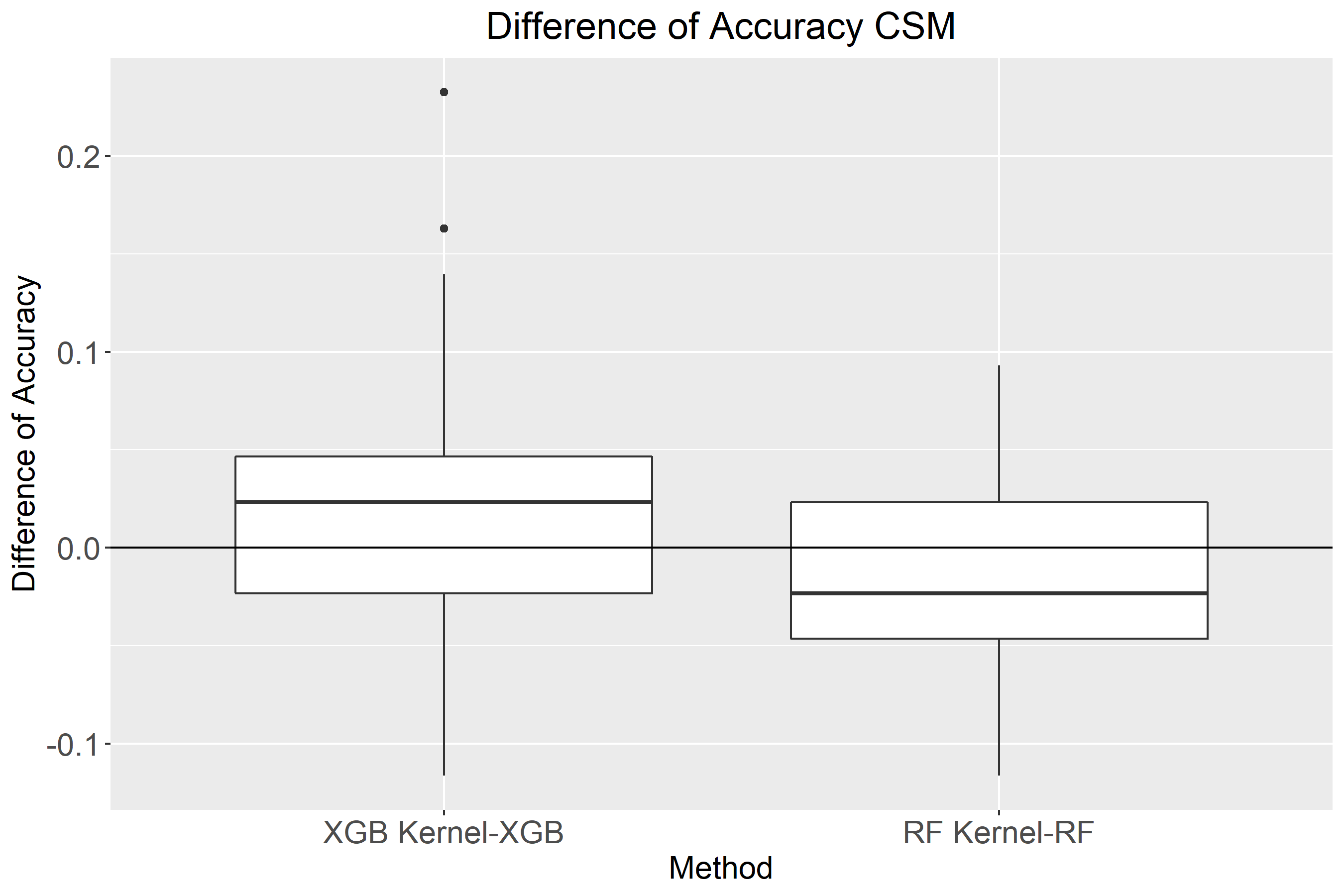}  
     \subcaption{}
  \label{fig:sub-first}
\end{subfigure}
\caption{Comparison of MSE, classification accuracy using RF, RF kernel, XGB, and XGB Kernel for the Conventional and Social Movie data}
\label{fig:CSM}
\end{figure}

\begin{figure}[ht]
\begin{subfigure}{.45\textwidth}
  \centering
  \includegraphics[height=0.2\textheight]{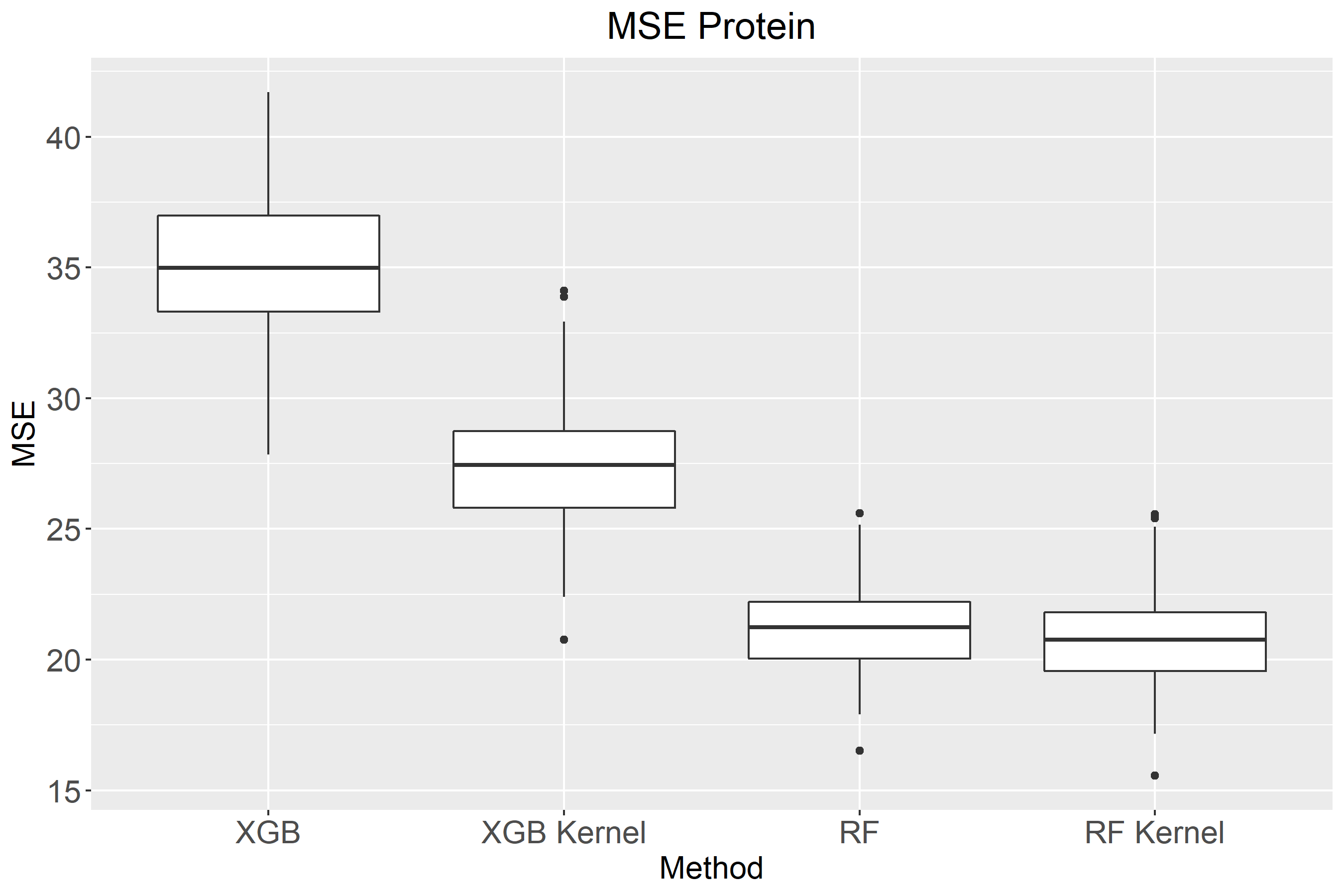}  
     \subcaption{}
  \label{fig:sub-first}
\end{subfigure}
\begin{subfigure}{.45\textwidth}
  \centering
  \includegraphics[height=0.2\textheight]{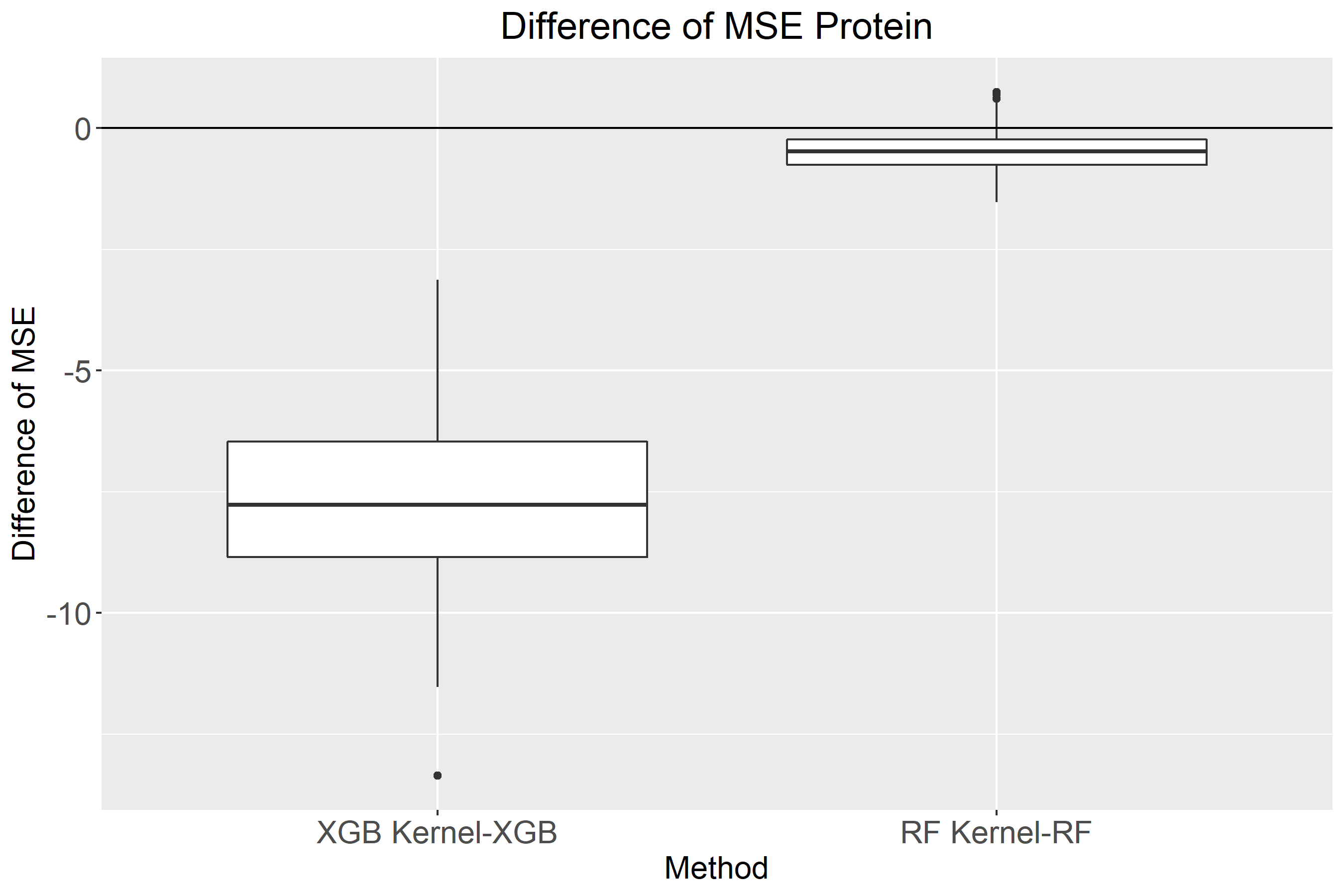}  
     \subcaption{}
  \label{fig:sub-first}
\end{subfigure}\\
\begin{subfigure}{.45\textwidth}
  \centering
  \includegraphics[height=0.2\textheight, width=0.9\textwidth]{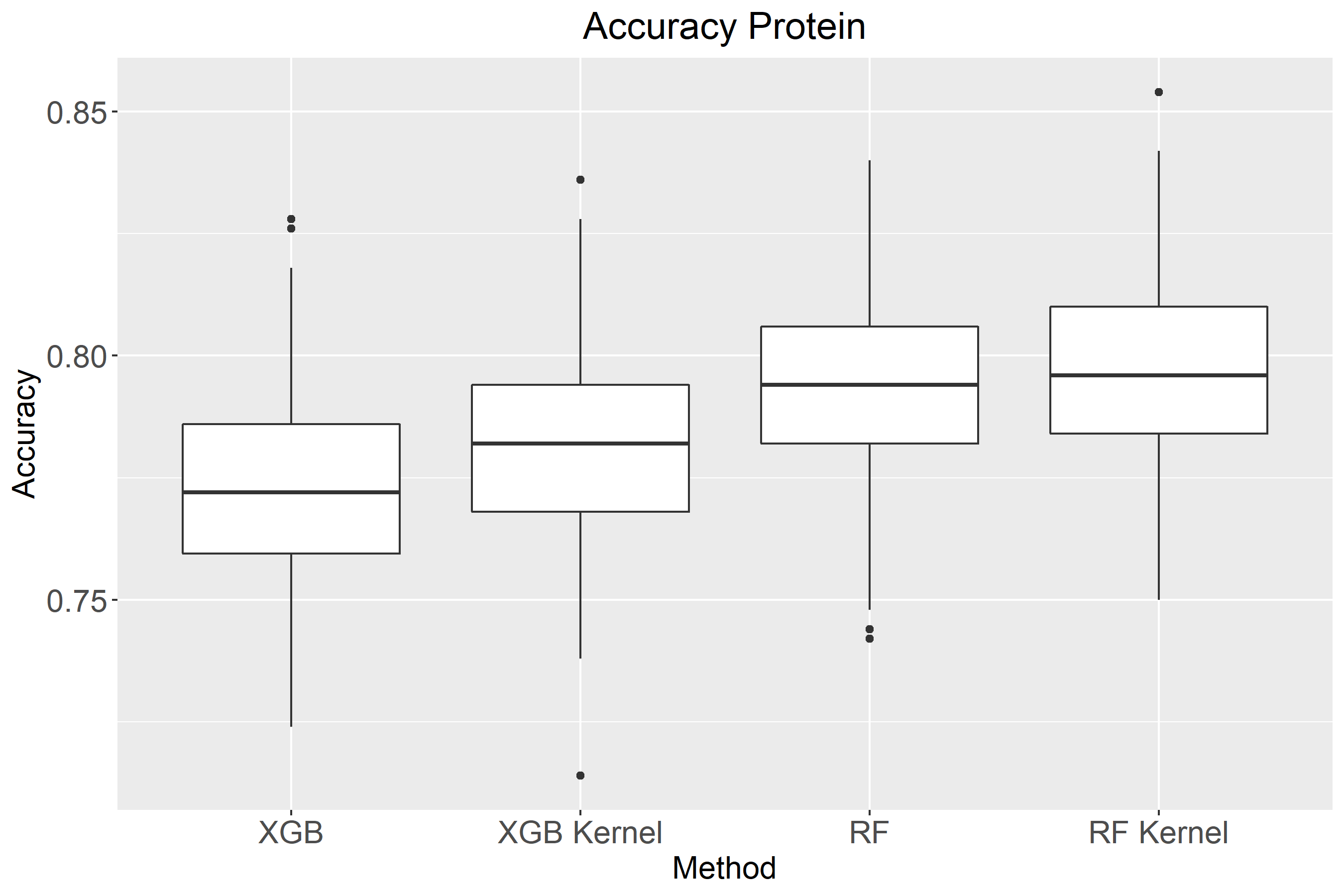}  
     \subcaption{}
  \label{fig:sub-first}
\end{subfigure}
\begin{subfigure}{.45\textwidth}
  \centering
  \includegraphics[height=0.2\textheight]{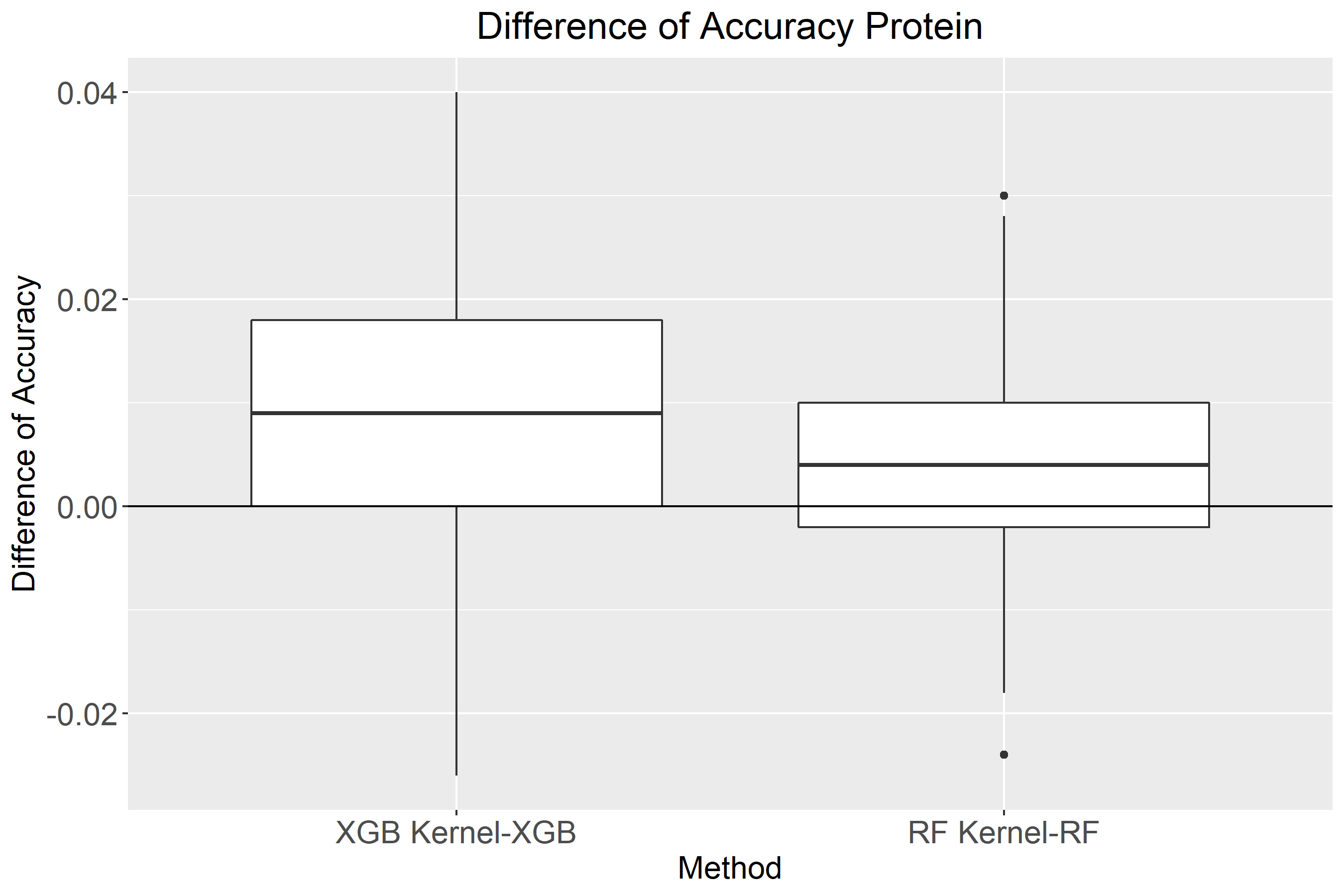}  
     \subcaption{}
  \label{fig:sub-first}
\end{subfigure}
\caption{Comparison of MSE, classification accuracy using RF, RF kernel, XGB, and XGB Kernel for the Protein Tertiary Structure data}
\label{fig:protein}
\end{figure}

\begin{figure}[ht]
\begin{subfigure}{.45\textwidth}
  \centering
  \includegraphics[height=0.2\textheight]{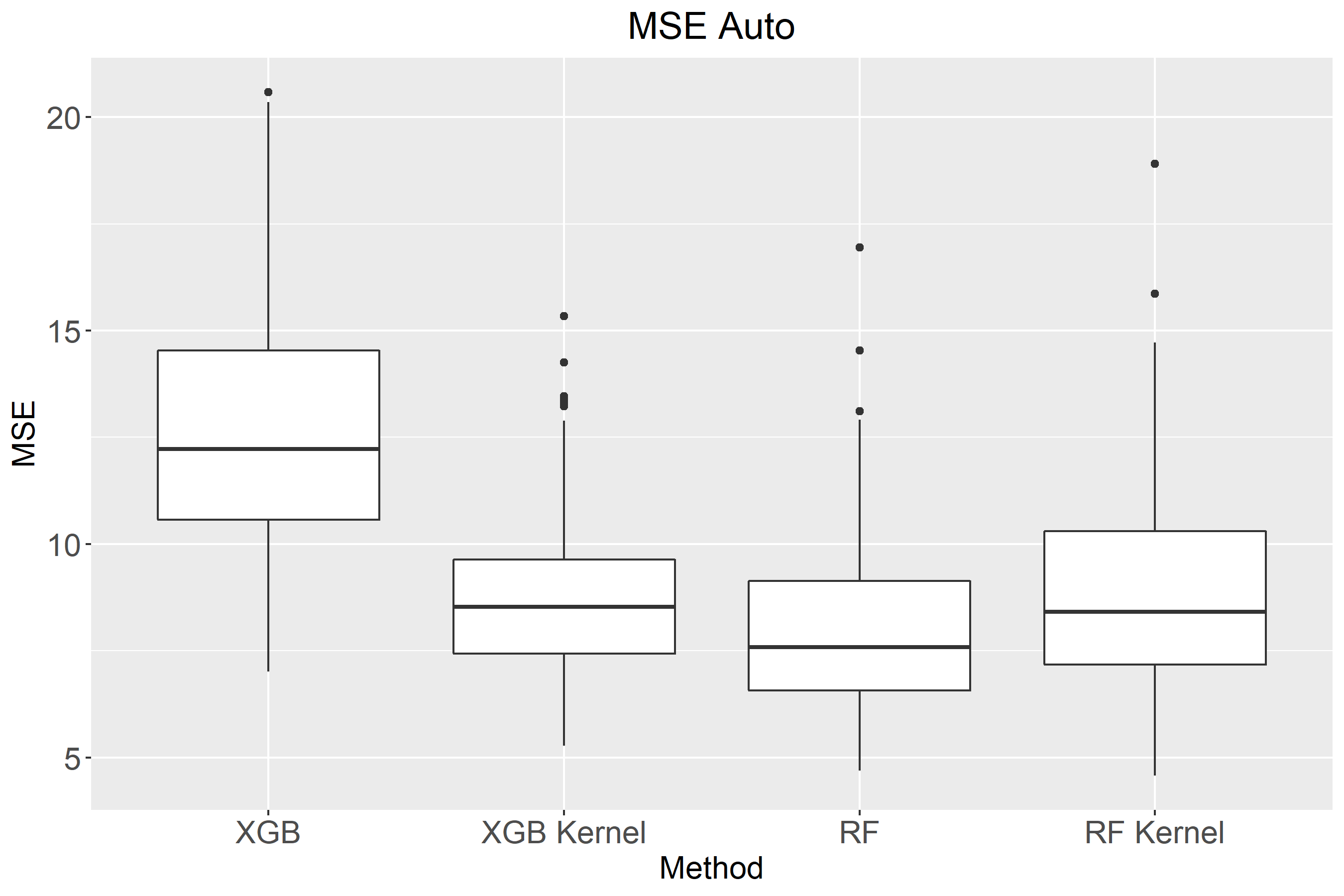}  
     \subcaption{}
  \label{fig:sub-first}
\end{subfigure}
\begin{subfigure}{.45\textwidth}
  \centering
  \includegraphics[height=0.2\textheight]{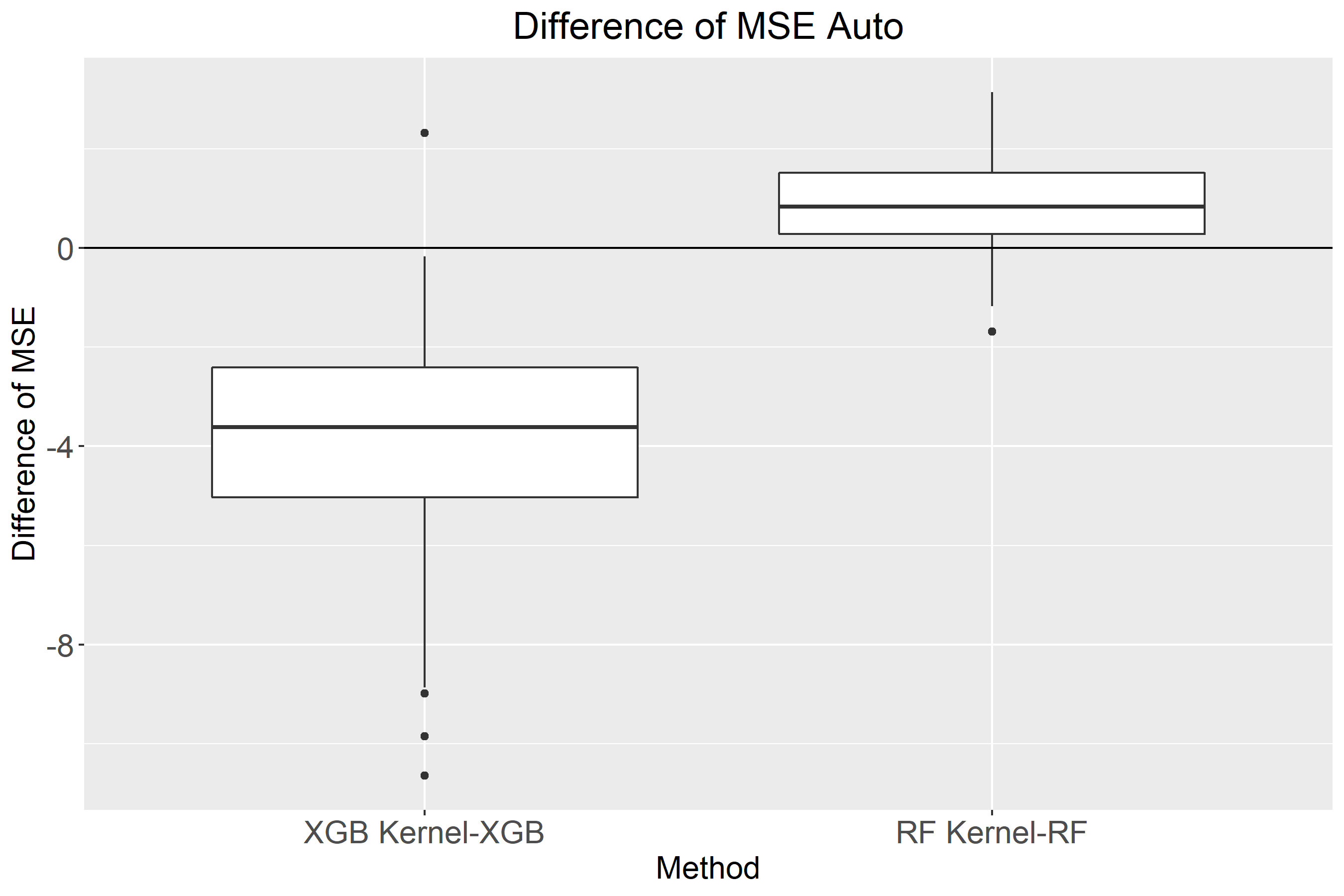}  
     \subcaption{}
  \label{fig:sub-first}
\end{subfigure}\\
\begin{subfigure}{.45\textwidth}
  \centering
  \includegraphics[height=0.2\textheight, width=0.9\textwidth]{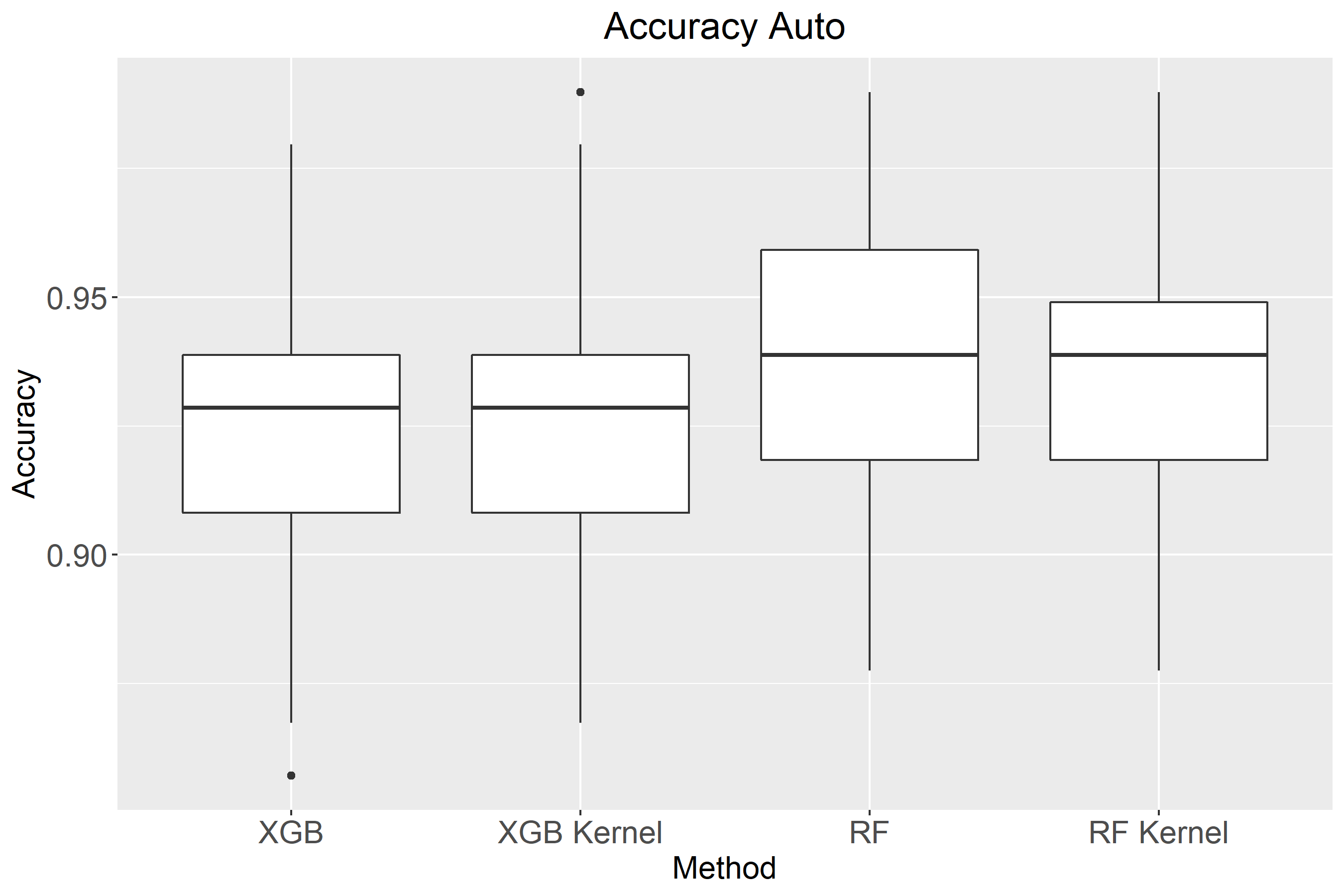}  
     \subcaption{}
  \label{fig:sub-first}
\end{subfigure}
\begin{subfigure}{.45\textwidth}
  \centering
  \includegraphics[height=0.2\textheight]{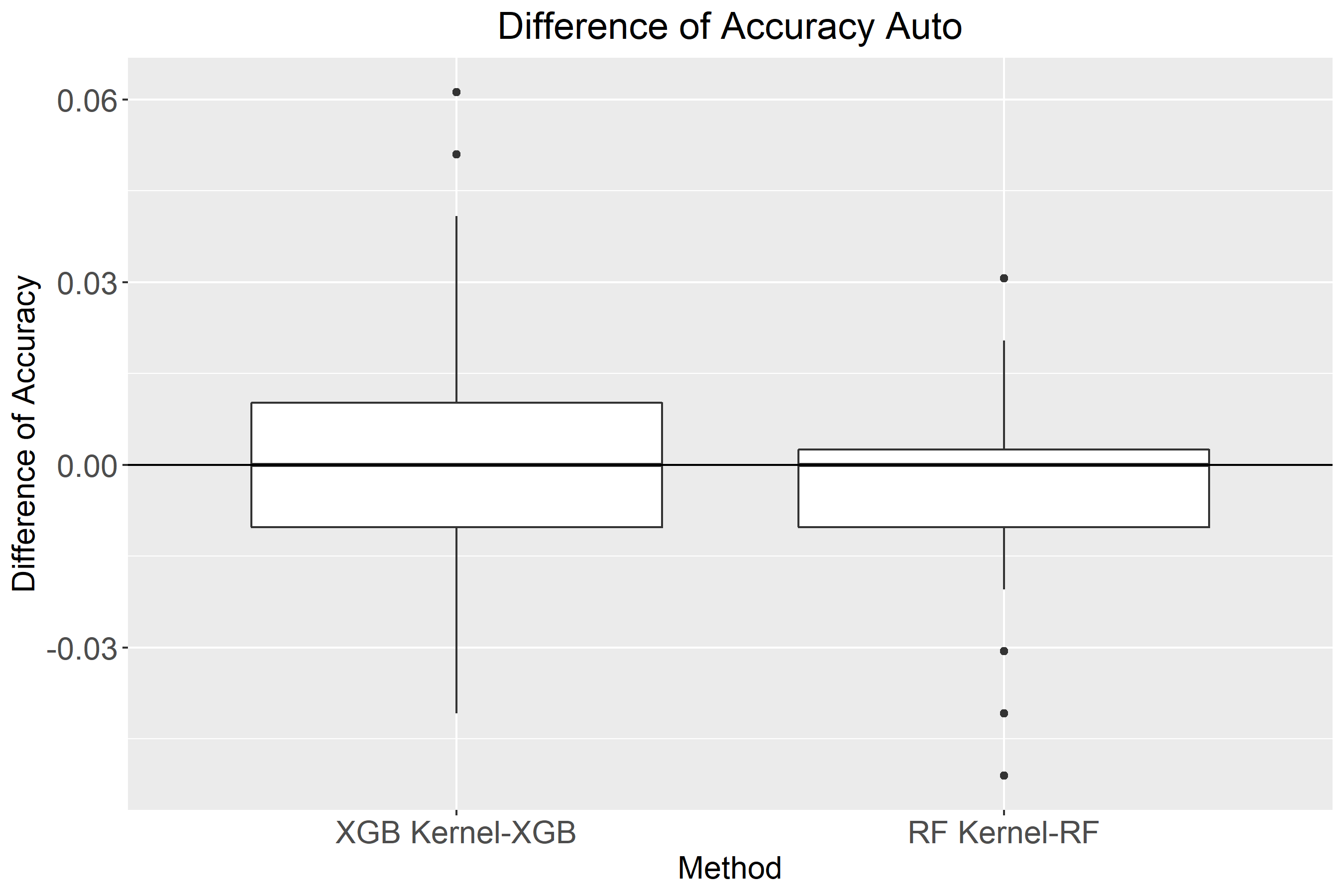}  
     \subcaption{}
  \label{fig:sub-first}
\end{subfigure}
\caption{Comparison of MSE, classification accuracy using RF, RF kernel, XGB, and XGB Kernel for the Auto mpg data}
\label{fig:auto}
\end{figure}

Results from this real life problem mimic those obtained in simulation (Figs. \ref{fig:california} to \ref{fig:auto}). In regression setting, the RF/XGB kernels were generally competitive to the RF/XGB across the data sets investigated. For the Auto mpg, the RF outperformed the RF kernel (see Figs. \ref{fig:auto}(b)). For Servo, the performance of the XGB kernel was comparable to that of the XGB  (see Fig. \ref{fig:servo}(b)) and for CSM the RF kernel was comparable to the RF (see Figs. \ref{fig:CSM}(b)). 

For classification, also in line with the results in simulation, the performance was impacted by dichotomization of the targets. Across the data sets RF/XGB kernels were generally comparable or slightly better than the RF/XGB algorithms. For CSM and Auto data sets, the RF kernel was slightly worse than RF (see Figs.\ref{fig:CSM}(d) and  \ref{fig:auto}(d)).

The results obtained from the real life data sets for regression and classification are summarized in the Tables \ref{tab:tableDataSetsRegression} and \ref{tab:tableDataSetsClassification}, respectively.

\begin{table}

\centering
   \caption{Results from the real life datasets from regression, MSE mean (s.d.) }
   \begin{tabular}{lrrrr}
  \hline
  Dataset & RF & RF kernel & XGB & XGB kernel \\
  \hline
  California housing &  $3.64\times10^9$  & $3.32\times10^9$  & $5.46\times10^9$  & $3.64\times10^9$ \\
   &($3.96\times10^8$) & ($4.0\times10^8$) & ($6.06\times10^8$) & ($3.87\times10^8$) \\
  Boston housing & 12.5 (4.6) & 10.2 (3.75)& 19.17 (5.7)& 13.34 (3.54) \\
  Concrete Compressive Strength & 31.08 (4.88) & 21.20 (4.62) & 38.04 (7.85) & 18.65 (3.91) \\
  Servo System & 38.10 (9.40) &22.97 (6.78) &31.67(9.71) & 31.04 (7.81)\\
  Conventional and Social Movie &  0.72 (0.15) & 0.77 (0.16)& 1.12 (0.27)& 0.90 (0.18)\\
  Protein Tertiary Structure & 21.24 (1.56) & 20.78 (1.69) & 35.12 (2.77) & 27.38 (2.25)\\
  Auto mpg &7.99 (1.92) &8.89 (2.26) &12.56 (2.84) & 8.75 (1.76)\\\hline
   \end{tabular}
\label{tab:tableDataSetsRegression}   
\end{table}

\begin{table}

\centering
   \caption{Results from the real life datasets from classification, accuracy mean (s.d.)}
     \begin{tabular}{lrrrr}
  \hline
  Dataset & RF & RF kernel & XGB & XGB kernel \\
  \hline
  California housing & 0.85 (0.02)  & 0.86 (0.02) & 0.85 (0.01) & 0.86 (0.01) \\
  Boston housing & 0.87 (0.02)& 0.87 (0.02)& 0.84 (0.03)& 0.86 (0.03) \\
  Concrete Compressive Strength & 0.91 (0.02) & 0.92 (0.02) & 0.90 (0.02) & 0.91 (0.02)\\
  Servo System &0.89 (0.05) & 0.88 (0.04) &0.87 (0.05)&0.88 (0.05)\\
  Conventional and Social Movie & 0.67 (0.06) &0.66 (0.06) & 0.63 (0.07)& 0.65 (0.06)\\
  Protein Tertiary Structure & 0.79 (0.02)& 0.80 (0.02) & 0.77 (0.02) & 0.78 (0.02)\\
  Auto mpg &0.94 (0.02) &0.94(0.02) &0.92 (0.03) &0.93 (0.02)\\\hline
   \end{tabular}
\label{tab:tableDataSetsClassification}   
\end{table}


\section{Discussion and Conclusions}\label{sec5}
It has been noticed in \cite{marcus2017}, that the RF kernel has been overlooked and underutilized in the statistical machine learning. RF and more generally, the tree ensemble based kernel matrix is akin to the variable importance \cite{ishwaran2019}, as is obtained across the different prediction targets in the same form, i.e. as an $n\times n$ matrix whose entries range between 0 and 1 and represent the estimates of probability of two points being assigned to the same terminal node. Generalization of the RF kernel for other tree ensembles such as gradient boosting have been considered in \cite{Chen2018} and also discussed in \cite{fan2020}.
In our contribution we systematically evaluated the RF and XGB kernel prediction models in a comprehensive simulation study that included regression and classification targets. Although the RF/XGB kernels are furnished by the RF/XGB, the prediction model is built in a different way than that of RF/XGB. The difference in these two approaches, specifically the way how they use the partitions of the data emanating from the recursive tree partitioning has been also noted in Refs. \cite{balog2016},\cite{ren2015}. In \cite{balog2016} the Mondrian forest and Mondrian kernel in Bayesian framework are contrasted. It has been pointed out that the predictor of the Mondrian forest is obtained by averaging across the single trees, whereas the predictor for the Mondrian kernel is obtained jointly from the kernel by a linear learning method (model). In contrast to the usual RF predictor, the use of RF kernel to obtain the RF kernel predictor (in agreement with \cite{balog2016}) has been referred to as a global refinement of the RF in \cite{ren2015} and \cite{gogic2021} . In our investigation, as RF/XGB underlies both RF/XGB and RF/XGB kernel, where the prediction model for the RF/XGB kernel is a linear model that capitalizes on the RF/XGB kernel. It is expected that RF/XGB and RF/XGB kernels will be working in a similar fashion \cite{balog2016},\cite{ren2015}, nevertheless there still may be differences. In our simulations we showed that for cases with larger number of noisy features the RF/XGB kernel approach may be superior to that of RF/XGB themselves. This beneficial effect was found particularly consistent for the continuous targets, i.e. regression. The simple linear model that follows the RF/XGB kernel construction (akin to the commonly used kernel methods) was found to be less prone to the noisy features in the simulation scenarios investigated. For classification we used the kernel ridge regression with target classes denoted as -1 and 1. Due to dichotomization of the continuous target \cite{fedorov2008}, the results were less pronounced for classification than those for the regression. There are other potential options for linear models that could have been used here. To this end, we also tried the regularized kernel logistic regression as implemented in the package gelnet \cite{sokolov2016} with RF and RF kernel yielding comparable performance to each other and with that of the kernel ridge regression.

The RF based models (RF and the RF kernel) performed as well or slightly outperformed the XGB based methods (XGB and the XGB kernel) in our experiments. However, there is no free lunch for statistical learning and consequently for a universally optimal kernel \cite{wolpert1996}, \cite{davies2014}, \cite{fernandezdelgado2014}. The success of a particular kernel algorithm depends on how well it adapts to the data geometry \cite{olson2018}, i.e. how well it captures the inherent kernel function of a given problem \cite{balcan2008}. The RF/XGB and accordingly the RF/XGB kernel should be competitive in situations when the data generating mechanism is conducive to the recursive partitioning \cite{Chen2018}, e.g. in the presence of feature interactions as frequently found in biomedical applications \cite{Boulesteix2012}. Another recent examples, where the RF kernel has shown promise are studies of the image classification in hyperspectral imaging \cite{zafari2019} and face alignment from imaging data \cite{gogic2021}. Moreover, in a large bench-marking study of general purpose classification algorithms \cite{fernandezdelgado2014}, RF  was found superior to other competitors. Interestingly, kernel methods that used the Gaussian kernel performed also well and were only slightly inferior to the RF. Similarly in a very recent bench-marking study \cite{gonzalez2020}, XGB and RF have been found competitive in classification. These results suggest that across broader spectrum of real life problems RF and XGB classifiers adapt well to the underlying data structure \cite{olson2018} and in many cases perform better than the classifiers based on the conventional kernels such as the Gaussian kernel. 
It would be of interest to conduct more research into how the results from \cite{fernandezdelgado2014} and \cite{gonzalez2020} extend from classification to regression and potentially survival and what implications they have for the tree ensemble based kernels. 

In addition to the simulations, we have also shown that in real life applications RF/XGB kernels are competitive to RF/XGB.
However, the usefulness of tree tree based kernels lies not only in a potential improvement of performance in certain high-dimensional setups. Availability of the tree based kernels for regression and classification explicitly renders the similarity/dissimilarity of the points ($\mbold{X}$-s) induced by the supervised tree based kernels. This can be then straightforwardly leveraged to define prototypical (archetypal) points (observations) with insights into the geometry of a given problem. Usefulness of the prototypes has been shown for the classification in \cite{bien2011} and \cite{brophy2020}. The generality of the tree based kernels extends it also to regression. Tree based kernels can be also used for prototypical or landmarking classification \cite{pekalska2001},\cite{balcan2008},\cite{kar2011}.  Using this approach the similarity/dissimilarity of the points to the points in the reference/landmarking set provides for an embedding that can be used not only to achieve a competitive prediction performance but also for an improved understanding of the intrinsic dimensionality of the problem. Further research in this direction is germane to solving real world prediction problems in classification and regression.

Kernel based methods can be adapted to the survival target \cite{ishwaran2019},\cite{wang2020}, where the potential censoring of the data needs to be addressed. Elucidation of the properties of the survival target attracted interest recently and it has been driven by real life applications, particularly in biomedical area \cite{ishwaran2019}. In a related work Chen investigated survival forest kernel used in conjunction with the Beran's kernel conditional Kaplan-Meier estimator \cite{chen2019} and in a follow-up leveraged the survival forest kernel to warm-start of kernel learning  \cite{chen2020}.
We carried out preliminary experiments with the RF kernel and the survival target with results similar to those presented here for continuous target (regression). An in depth investigation of this topic will be communicated in a separate contribution.

In our work we focused on the tree ensemble based kernels in frequentist framework. Tree based kernels from the Bayesian random forests (e.g. Mondrian forests \cite{balog2016}), Bayesian boosting by BART \cite{linero2017} and other non-parametric Bayesian partitions \cite{fan2020} can be obtained for regression and classification. Therefore further understanding of the performance of the kernels from the Bayesian approaches is another interesting topic for future research. 

Besides point estimates, the RF kernel can be utilized e.g. in Gaussian process or in conformal inference framework to obtain prediction intervals to quantify the uncertainly around prediction estimates, which can be harnessed in subsequent decision making \cite{chen2020}.

\section*{Acknowledgments}
We would like to thank the anonymous reviewers for their insightful comments and suggestions, as they greatly improved the quality of the manuscript. We particularly appreciate the suggestion to extend our work beyond the RF kernel to other tree ensemble based kernels.

\subsection*{Author contributions}
All authors contributed equally to this manuscript.

\subsection*{Financial disclosure}

There is no financial disclosure to report.    

\subsection*{Conflict of interest}

The authors declare no potential conflict of interests.

\clearpage
\section*{Supporting information}
\label{sec:supp}
The following supporting information is available as part of the online article:

\noindent

\section{Figures of the Boxplots of the Performance Metrics Across the Simulation Setups}

\noindent

\noindent
\noindent
\noindent
\clearpage

\section{Tables of Performance Results for Continuous and Binary Targets Across All Setups}
\begin{table}
\centering
   \caption{Summary of simulation results of the mean squared error (MSE) for continuous target}

\begin{tabular}{rlrrrrrr}
  \hline
  & Setup & n & p & RF & RF kernel & XGB & XGB Kernel  \\ 
&       &   &   & mean (sd) & mean (sd) & mean (sd) & mean (sd)  \\
  \hline
1 & Friedman & 800 & 20 & 6.827 (0.669) & 5.263 (0.598) & 10.427 (1.495) & 5.935 (0.739) \\ 
  2 & Friedman & 1600 & 20 & 5.555 (0.409) & 4.325 (0.359) & 8.087 (0.979) & 4.398 (0.481) \\ 
  3 & Friedman & 800 & 40 & 8.939 (0.868) & 6.555 (0.712) & 11.932 (1.779) & 7.838 (0.995) \\ 
  4 & Friedman & 1600 & 40 & 7.419 (0.526) & 5.452 (0.417) & 9.083 (1.1) & 5.92 (0.618) \\ 
  5 & Checkerboard & 800 & 20 & 3.825 (0.847) & 3.268 (0.775) & 6.85 (1.366) & 4.219 (0.758) \\ 
  6 & Checkerboard & 1600 & 20 & 3.226 (0.493) & 2.764 (0.44) & 5.612 (0.736) & 3.445 (0.39) \\ 
  7 & Checkerboard & 800 & 40 & 4.032 (0.862) & 3.479 (0.762) & 7.463 (1.538) & 5.351 (1.027) \\ 
  8 & Checkerboard & 1600 & 40 & 3.495 (0.539) & 2.984 (0.476) & 6.134 (0.829) & 4.377 (0.547) \\ 
  9 & Meier1 & 800 & 20 & 0.499 (0.054) & 0.402 (0.043) & 0.752 (0.086) & 0.583 (0.065) \\ 
  10 & Meier1 & 1600 & 20 & 0.429 (0.033) & 0.36 (0.027) & 0.673 (0.05) & 0.528 (0.039) \\ 
  11 & Meier1 & 800 & 40 & 0.622 (0.065) & 0.468 (0.051) & 0.794 (0.09) & 0.674 (0.079) \\ 
  12 & Meier1 & 1600 & 40 & 0.529 (0.043) & 0.409 (0.033) & 0.716 (0.056) & 0.597 (0.051) \\ 
  13 & Meier2 & 800 & 20 & 1.562 (0.172) & 1.093 (0.122) & 1.537 (0.205) & 1.191 (0.136) \\ 
  14 & Meier2 & 1600 & 20 & 1.259 (0.101) & 0.901 (0.072) & 1.151 (0.141) & 0.916 (0.096) \\ 
  15 & Meier2 & 800 & 40 & 2.132 (0.243) & 1.334 (0.156) & 1.755 (0.257) & 1.44 (0.189) \\ 
  16 & Meier2 & 1600 & 40 & 1.729 (0.137) & 1.106 (0.089) & 1.287 (0.151) & 1.096 (0.12) \\ 
  17 & van der Laan & 800 & 20 & 0.617 (0.062) & 0.571 (0.058) & 0.943 (0.112) & 0.747 (0.09) \\ 
  18 & van der Laan & 1600 & 20 & 0.586 (0.043) & 0.53 (0.04) & 0.849 (0.077) & 0.646 (0.054) \\ 
  19 & van der Laan & 800 & 40 & 0.658 (0.069) & 0.624 (0.065) & 1.03 (0.118) & 0.882 (0.098) \\ 
  20 & van der Laan & 1600 & 40 & 0.628 (0.048) & 0.588 (0.045) & 0.958 (0.093) & 0.797 (0.072) \\ 
   \hline
\end{tabular}
\label{tab:tableContinuous}
\end{table}

\begin{table}
\centering
   \caption{Summary of simulation results of the classification accuracy  for binary target}
\begin{tabular}{rlrrrrrr}
  \hline
 & Setup & n & p & RF & RF kernel & XGB & XGB Kernel  \\ 
&       &   &   & mean (sd) & mean (sd) & mean (sd) & mean (sd)  \\
\hline
  1 & Friedman & 800 & 20 & 0.865 (0.026) & 0.869 (0.025) & 0.83 (0.029) & 0.84 (0.029) \\ 
  2 & Friedman & 1600 & 20 & 0.88 (0.016) & 0.88 (0.017) & 0.844 (0.019) & 0.853 (0.017) \\ 
  3 & Friedman & 800 & 40 & 0.852 (0.027) & 0.859 (0.026) & 0.824 (0.029) & 0.834 (0.026) \\ 
  4 & Friedman & 1600 & 40 & 0.871 (0.017) & 0.875 (0.016) & 0.838 (0.019) & 0.845 (0.018) \\ 
  5 & Checkerboard & 800 & 20 & 0.731 (0.032) & 0.732 (0.033) & 0.653 (0.034) & 0.671 (0.033) \\ 
  6 & Checkerboard & 1600 & 20 & 0.741 (0.022) & 0.739 (0.023) & 0.665 (0.024) & 0.679 (0.024) \\ 
  7 & Checkerboard & 800 & 40 & 0.729 (0.033) & 0.727 (0.033) & 0.65 (0.036) & 0.659 (0.035) \\ 
  8 & Checkerboard & 1600 & 40 & 0.735 (0.022) & 0.736 (0.021) & 0.657 (0.026) & 0.666 (0.025) \\ 
  9 & Meier1 & 800 & 20 & 0.667 (0.031) & 0.653 (0.032) & 0.597 (0.033) & 0.601 (0.033) \\ 
  10 & Meier1 & 1600 & 20 & 0.678 (0.024) & 0.663 (0.025) & 0.599 (0.026) & 0.603 (0.025) \\ 
  11 & Meier1 & 800 & 40 & 0.661 (0.035) & 0.644 (0.036) & 0.593 (0.036) & 0.596 (0.036) \\ 
  12 & Meier1 & 1600 & 40 & 0.678 (0.024) & 0.66 (0.025) & 0.6 (0.026) & 0.6 (0.025) \\ 
  13 & Meier2 & 800 & 20 & 0.766 (0.033) & 0.759 (0.033) & 0.701 (0.036) & 0.709 (0.032) \\ 
  14 & Meier2 & 1600 & 20 & 0.776 (0.021) & 0.771 (0.021) & 0.715 (0.023) & 0.717 (0.023) \\ 
  15 & Meier2 & 800 & 40 & 0.765 (0.029) & 0.759 (0.03) & 0.697 (0.033) & 0.705 (0.032) \\ 
  16 & Meier2 & 1600 & 40 & 0.772 (0.022) & 0.767 (0.021) & 0.708 (0.024) & 0.71 (0.024) \\ 
  17 & van der Laan & 800 & 20 & 0.539 (0.037) & 0.539 (0.038) & 0.516 (0.035) & 0.522 (0.036) \\ 
  18 & van der Laan & 1600 & 20 & 0.55 (0.026) & 0.547 (0.027) & 0.524 (0.026) & 0.526 (0.025) \\ 
  19 & van der Laan & 800 & 40 & 0.523 (0.036) & 0.524 (0.033) & 0.51 (0.036) & 0.511 (0.036) \\ 
  20 & van der Laan & 1600 & 40 & 0.54 (0.025) & 0.537 (0.025) & 0.518 (0.024) & 0.519 (0.025) \\\hline \end{tabular}
\label{tab:tableBinary}
\end{table}

\clearpage
\subsection{Tables of Performance Results for Continuous and Binary Targets Across All Setups for the Sensitivity Analysis}

\begin{table}
\centering
   \caption{Summary of simulation results of the mean squared error (MSE) for continuous target (sensitivity analysis)}
\begin{tabular}{rlrrrrrr}
  \hline
 & Setup & n & p & RF & RF kernel & XGB & XGB Kernel  \\ 
 &       &   &   & mean (sd) & mean (sd) & mean (sd) & mean (sd)\\
 \hline
1 & Friedman & 800 & 20 & 6.956 (0.684) & 4.677 (0.549) & 5.904 (1.257) & 3.552 (0.516) \\ 
  2 & Friedman & 1600 & 20 & 5.644 (0.407) & 3.81 (0.308) & 4.175 (0.556) & 3.219 (0.334) \\ 
  3 & Friedman & 800 & 40 & 9.031 (0.867) & 5.899 (0.636) & 7.037 (1.342) & 4.632 (0.619) \\ 
  4 & Friedman & 1600 & 40 & 7.49 (0.529) & 4.875 (0.371) & 5.04 (0.804) & 3.936 (0.511) \\ 
  5 & Checkerboard & 800 & 20 & 3.95 (0.898) & 3.097 (0.708) & 5.473 (1.191) & 3.591 (0.623) \\ 
  6 & Checkerboard & 1600 & 20 & 3.314 (0.518) & 2.632 (0.405) & 4.253 (0.593) & 3.286 (0.354) \\ 
  7 & Checkerboard & 800 & 40 & 4.129 (0.899) & 3.328 (0.686) & 6.456 (1.325) & 4.593 (0.735) \\ 
  8 & Checkerboard & 1600 & 40 & 3.547 (0.584) & 2.807 (0.46) & 4.748 (0.671) & 3.861 (0.404) \\ 
  9 & Meier1 & 800 & 20 & 0.507 (0.055) & 0.389 (0.04) & 0.607 (0.083) & 0.483 (0.051) \\ 
  10 & Meier1 & 1600 & 20 & 0.434 (0.033) & 0.35 (0.026) & 0.529 (0.05) & 0.494 (0.043) \\ 
  11 & Meier1 & 800 & 40 & 0.627 (0.066) & 0.445 (0.049) & 0.653 (0.105) & 0.521 (0.059) \\ 
  12 & Meier1 & 1600 & 40 & 0.533 (0.043) & 0.393 (0.032) & 0.559 (0.054) & 0.517 (0.045) \\ 
  13 & Meier2 & 800 & 20 & 1.607 (0.177) & 0.999 (0.117) & 0.761 (0.1) & 0.607 (0.069) \\ 
  14 & Meier2 & 1600 & 20 & 1.284 (0.104) & 0.815 (0.063) & 0.61 (0.057) & 0.556 (0.046) \\ 
  15 & Meier2 & 800 & 40 & 2.183 (0.25) & 1.229 (0.14) & 0.878 (0.128) & 0.753 (0.097) \\ 
  16 & Meier2 & 1600 & 40 & 1.759 (0.136) & 1.004 (0.078) & 0.67 (0.061) & 0.615 (0.053) \\ 
  17 & van der Laan & 800 & 20 & 0.62 (0.062) & 0.562 (0.057) & 0.76 (0.09) & 0.623 (0.069) \\ 
  18 & van der Laan & 1600 & 20 & 0.588 (0.043) & 0.519 (0.039) & 0.66 (0.052) & 0.602 (0.047) \\ 
  19 & van der Laan & 800 & 40 & 0.659 (0.069) & 0.618 (0.064) & 0.903 (0.107) & 0.767 (0.089) \\ 
  20 & van der Laan & 1600 & 40 & 0.628 (0.048) & 0.58 (0.044) & 0.779 (0.076) & 0.704 (0.063) \\ 
   \hline
\end{tabular}
\label{tab:tableContinuousNode2}
\end{table}

\begin{table}
\centering
   \caption{Summary of simulation results of the classification accuracy for binary target (sensitivity analysis)}
\begin{tabular}{rlrrrrrr}
  \hline
 & Setup & n & p & RF & RF kernel & XGB & XGB Kernel  \\ 
 &       &   &   & mean (sd) & mean (sd) & mean (sd) & mean (sd)\\
\hline
1 & Friedman & 800 & 20 & 0.865 (0.026) & 0.869 (0.025) & 0.824 (0.031) & 0.828 (0.03) \\ 
  2 & Friedman & 1600 & 20 & 0.88 (0.016) & 0.88 (0.017) & 0.838 (0.021) & 0.84 (0.021) \\ 
  3 & Friedman & 800 & 40 & 0.852 (0.027) & 0.859 (0.026) & 0.817 (0.029) & 0.816 (0.029) \\ 
  4 & Friedman & 1600 & 40 & 0.871 (0.017) & 0.875 (0.016) & 0.827 (0.021) & 0.829 (0.02) \\ 
  5 & Checkerboard & 800 & 20 & 0.731 (0.032) & 0.732 (0.033) & 0.652 (0.036) & 0.67 (0.037) \\ 
  6 & Checkerboard & 1600 & 20 & 0.741 (0.022) & 0.739 (0.023) & 0.659 (0.022) & 0.662 (0.023) \\ 
  7 & Checkerboard & 800 & 40 & 0.729 (0.033) & 0.727 (0.033) & 0.64 (0.037) & 0.652 (0.037) \\ 
  8 & Checkerboard & 1600 & 40 & 0.735 (0.022) & 0.736 (0.021) & 0.653 (0.023) & 0.657 (0.022) \\ 
  9 & Meier1 & 800 & 20 & 0.667 (0.031) & 0.653 (0.032) & 0.596 (0.034) & 0.603 (0.035) \\ 
  10 & Meier1 & 1600 & 20 & 0.678 (0.024) & 0.663 (0.025) & 0.599 (0.024) & 0.597 (0.023) \\ 
  11 & Meier1 & 800 & 40 & 0.661 (0.035) & 0.644 (0.036) & 0.592 (0.033) & 0.597 (0.034) \\ 
  12 & Meier1 & 1600 & 40 & 0.678 (0.024) & 0.66 (0.025) & 0.596 (0.024) & 0.597 (0.023) \\ 
  13 & Meier2 & 800 & 20 & 0.766 (0.033) & 0.759 (0.033) & 0.701 (0.033) & 0.705 (0.03) \\ 
  14 & Meier2 & 1600 & 20 & 0.776 (0.021) & 0.771 (0.021) & 0.711 (0.022) & 0.71 (0.022) \\ 
  15 & Meier2 & 800 & 40 & 0.765 (0.029) & 0.759 (0.03) & 0.7 (0.034) & 0.702 (0.032) \\ 
  16 & Meier2 & 1600 & 40 & 0.772 (0.022) & 0.767 (0.021) & 0.703 (0.024) & 0.702 (0.024) \\ 
  17 & van der Laan & 800 & 20 & 0.539 (0.037) & 0.539 (0.038) & 0.522 (0.033) & 0.528 (0.035) \\ 
  18 & van der Laan & 1600 & 20 & 0.55 (0.026) & 0.547 (0.027) & 0.527 (0.026) & 0.529 (0.026) \\ 
  19 & van der Laan & 800 & 40 & 0.523 (0.036) & 0.524 (0.033) & 0.517 (0.036) & 0.514 (0.035) \\ 
  20 & van der Laan & 1600 & 40 & 0.54 (0.025) & 0.537 (0.025) & 0.525 (0.026) & 0.524 (0.025) \\ 
   \hline
\end{tabular}
\label{tab:tableBinaryNode2}
\end{table}

\clearpage
\appendix

\section{Objective function of the xgboost algorithm\label{app1}}

The objective function of the xgboost algorithm $L_{\text{XGB}}$ was obtained as follows:

\begin{eqnarray}
L_{\text{XGB}} &=& \sum_{i=1}^{n} l(y_i, h_{\text{XGB}}(X_i))+\Omega(h_m) \\
&=& \sum_{i=1}^{n} l(y_i, h_{\text{XGB}}(X_i))+\gamma T+ 0.5 \lambda ||w||^2
\end{eqnarray}
where \\
$l(.)$ is the loss function. The loss function used in xgboost is squared error and logistic loss for regression and classification, respectively  \\
$\Omega$ denotes the regularization penalty \\
$T$, $\gamma$ is the number of tree terminal nodes and a corresponding regularization parameter, respectively\\
$\lambda$ is a regularization parameter controlling for the L2 norm of the individual tree weights $||w||^2$

\bibliography{wileyNJD-AMS}%

\section*{Author Biography}

Dai Feng holds a Ph.D. from the University of Iowa, and is currently employed by AbbVie Inc. Richard Baumgartner holds a Ph.D. from Technische Universit\"{a}t Wien, and is currently employed by 
Merck Sharp \& Dohme Corp., a subsidiary of Merck \& Co., Inc., Kenilworth, NJ, USA.

\end{document}